\PassOptionsToPackage{authoryear}{natbib}

\documentclass[11pt]{article}
\RequirePackage[OT1]{fontenc}
\RequirePackage{amsthm,amsmath}
\usepackage{booktabs}   
\usepackage{tabularx}   
\usepackage{array}      

\newcolumntype{Y}{>{\raggedright\arraybackslash}X}

\usepackage{tikz}
\usetikzlibrary{shapes.geometric, arrows.meta, positioning, shadows}

\RequirePackage{hyperref}
\usepackage{amsfonts}
\usepackage{amssymb}
\usepackage{amstext}
\usepackage{parskip}
\usepackage{fullpage}
\usepackage{graphicx}
\usepackage{subcaption}

\RequirePackage[numbers]{natbib}
\usepackage{amsfonts,amssymb,amsthm,amsmath,enumerate}

\newcommand{\prob}[1]{\ensuremath{\mathbb P}\left(#1\right)}
\newcommand{\prbb}[1]{\ensuremath{\mathbb P}\big(#1\big)}
\newcommand{\R}{\ensuremath{\mathbb R}}

\newcommand{\Z}{\ensuremath{\mathbb Z}}

\newcommand{\size}[1]{\ensuremath{\left|#1\right|}}

\newcommand{\argmin}{\operatorname{argmin}}

\newcommand{\e}{\epsilon}
\newcommand{\ve}{\varepsilon}

\newcommand{\half}{\ensuremath{\frac{1}{2}}}
\newcommand{\silent}[1]{}

\newcommand{\D}{{\mathbb D}}
\newcommand{\Ball}{{\mathbf B}}

\newcommand{\MC}{{\mathcal C}}

\newcommand{\M}{{\mathcal M}}

\newcommand{\cov}{\textsf{Cov}}

\newcommand{\gap}{\textsf{gap}}

\newcommand{\inv}[1]{\frac{1}{#1}}
\newcommand{\abs}[1]{\left\lvert#1\right\rvert}
\newcommand{\twonorm}[1]{\left\lVert#1\right\rVert_2}
\newcommand{\norm}[1]{\left\lVert#1\right\rVert}

\def\cp{{\mathcal P}}
\def\cpk{{\mathcal P}_k}
\def\Z{{\mathbb Z}}
\def\E{{\mathbb E}}

\def\M{{\mathcal M}}

\def\supp{\mathop{\text{supp}\kern.2ex}}
\def\argmin{\mathop{\text{arg\,min}\kern.2ex}}
\let\hat\widehat
\let\tilde\widetilde

\def\qed{\hskip1pt $\;\;\scriptstyle\Box$}

\newcommand{\vecone}{{\bf 1}}
\newcommand{\vc}{{\bf c}}

\newcommand{\BW}{\ensuremath{\mathbb{W}}}

\def\argmax{\mathop{\text{\rm argmax}\kern.2ex}}

\def\fatnorm#1{|\kern-.2ex|\kern-.2ex| #1 |\kern-.2ex|\kern-.2ex|}

\newcommand{\shnorm}[1]{\lVert#1\rVert}
\newcommand{\fnorm}[1]{\left\lVert#1\right\rVert_F}
\newcommand{\infonenorm}[1]{\left\lVert#1\right\rVert_{\infty \to 1}}
\newcommand{\onenorm}[1]{\left\lVert#1\right\rVert_{1}}

\def\SDP{\mathop{\text{SDP}\kern.2ex}}
\newcommand{\func}[1]{\ensuremath{\mathrm{#1}}}
\newcommand{\diag}{\func{diag}}

\newcommand{\offd}{\func{offd}}

\newcommand{\Sp}{\mathbb{S}}
\newcommand{\ip}[1]{\langle{#1}\rangle}

\newcommand{\beq}{\begin{equation}}
\newcommand{\eeq}{\end{equation}}
\newcommand{\ben}{\begin{eqnarray}}
\newcommand{\een}{\end{eqnarray}}
\newcommand{\bnum}{\begin{enumerate}}
\newcommand{\enum}{\end{enumerate}}
\newcommand{\bit}{\begin{itemize}}
\newcommand{\eit}{\end{itemize}}
\newcommand{\bens}{\begin{eqnarray*}}
\newcommand{\eens}{\end{eqnarray*}}
\newcommand{\MG}{{\mathcal G}}

\def \D {\mathbb{D}}

\DeclareMathOperator*{\Span}{span}

\newcommand{\tr}{{\rm tr}}

\newcommand{\mvec}[1]{\rm{vec}\left\{#1\right\}}
\newcommand{\opt}{\ensuremath{\text{opt}}}
\newcommand{\CC}{{\mathcal C}}
\newcommand{\Net}{{\mathcal N}}

\newtheorem{theorem}{Theorem}

\newtheorem{lemma}[theorem]{Lemma}
\newtheorem{proposition}[theorem]{Proposition}
\newtheorem{conjecture}[theorem]{Conjecture}
\newtheorem{definition}[theorem]{Definition}

\newtheorem{remark}[theorem]{Remark}
\newtheorem{corollary}[theorem]{Corollary}

{\end{eqnarray}\end{subequations}\hskip-4.0pt}%

\newenvironment{proofof}[1]{\hspace*{20pt}{\it Proof}{ of #1}.\hskip10pt}{\qed\vskip5pt}
\newenvironment{proofof2}{}{\qed\vskip5pt}

\makeatletter

\newif\ifisarticle
\isarticlefalse

\def\barticle[#1]{%
  \item\parindent=0pt\hangindent=2em\hangafter=1\vspace{-2pt}%
  \bgroup
  \isarticletrue%
  \bauthor@checknumber#1,number=\relax,
}
\def\endbarticle{\egroup}

\def\bauthor@checknumber#1,number=#2,#3{%
  \ifx\relax#2\else
    \bauthor@extractnum#2\bauthor@end
  \fi%
}
\def\bauthor@extractnum#1 #2\bauthor@end{\def\tempissue{(#1)}}

\def\bbook[#1]{\item\parindent=0pt\hangindent=2em\hangafter=1}

\def\bproceedings[#1]{\item\parindent=0pt\hangindent=2em\hangafter=1}

\def\bcollection[#1]{\item\parindent=0pt\hangindent=2em\hangafter=1}

\def\breport[#1]{\item\parindent=0pt\hangindent=2em\hangafter=1}

\def\binproceedings[#1]{\item\parindent=0pt\hangindent=2em\hangafter=1}

\def\bincollection[#1]{\item\parindent=0pt\hangindent=2em\hangafter=1}

\def\binbook[#1]{\item\parindent=0pt\hangindent=2em\hangafter=1}

\let\bphdthesis\undefined

\def\bphdthesis[#1]{\item\parindent=0pt\hangindent=2em\hangafter=1\vspace{-2pt}}

\newcommand{\bauthor}[1]{#1}

\newcommand{\beditor}[1]{#1}

\renewcommand{\bphdthesis}[1]{#1}

\def\bpages#1{%
  \ifisarticle%
    \unskip
    \ifdefined\tempvol\tempvol\else\relax\fi
    \ifdefined\tempissue\tempissue\else\relax\fi
    :#1
    \let\tempvol\undefined
    \let\tempissue\undefined%
  \else%
    \unskip, #1
  \fi%
}

\def\bsnm#1{#1}     
\def\bfnm#1{#1}     
\def\binits#1{}

\def\bauthor{\@ifnextchar[{\bauthor@bracket}{\bauthor@nobracket}}
\def\bauthor@bracket[#1]#2{\bauthor@nobracket{#2}}
\def\bauthor@nobracket#1{#1}

\renewcommand{\beditor}[1]{%
  \bgroup
  \def\bsnm##1{##1}
  \def\bfnm##1{##1}
  \def\gobblecomma##1{}
  #1
  \egroup
}

\makeatother

\begin{document}

\title{Semidefinite programming relaxations and debiasing for MAXCUT-based clustering}

\author{Shuheng Zhou\\
    University of California, Riverside, CA 92521}

\date{}

\maketitle

\begin{abstract}
In this paper, we consider the problem of partitioning a small data sample of 
size $n$ drawn from a mixture of $2$ sub-gaussian distributions in
$\R^p$. We consider semidefinite programming relaxations of an
integer quadratic program that is formulated essentially as finding
the maximum cut on a graph, where edge weights in the cut represent
dissimilarity scores  between two nodes based on their $p$ features.
We define the signal-to-noise ratio (SNR) as
$s^2 := \min\{n p \gamma^2, 
\Delta^2\}$, where $\Delta^2 := p \gamma$ denotes the $\ell_2^2$
distance between the two cluster centers. 
For both balanced and unbalanced cases, we allow each population to 
have distinct covariance structures $\Sigma_1,  \Sigma_2$ with diagonal 
matrices as special cases.
Our contributions are twofold. First, we provide a unified framework
for analyzing three computationally efficient algorithms: SDP1,
BalancedSDP, and Spectral clustering, yielding universal
polynomial-rate misclassification guarantees for all three 
algorithms;
Moreover, our framework accommodates a full range of tradeoffs between $n, p,$ and $\gamma$, allowing for 
partial recovery (success  rate $< 100\%$)  when $s^2$ is lower
bounded by a constant.
Second, we prove that the misclassification errors for SDP1 and {\bf BalancedSDP} decay exponentially with respect to
the SNR $s^2$ and the {\bf BalancedSDP} requires  no explicit
debiasing when the two clusters have equal size.
To our knowledge, this is the first time such results are 
obtained for semidefinite relaxations of MAX CUT in population clustering.
We provide simulation evidence illuminating the theoretical 
predictions. 

\end{abstract}

\section{Introduction}
\label{sec:intro}
We explore a type of classification problem that arises in the context
of computational biology.
The biological context for this problem is that we are
given DNA information from $n$ individuals from $k$ populations 
of origin and we wish to classify each individual into the correct population.
DNA contains a series of markers called SNPs  (Single Nucleotide
Polymorphisms), each of which has two variants (alleles). We use bit
$1$ and bit $0$ to denote them. The problem is that we are given a small
sample of size $n$, e.g., DNA of $n$ individuals (think of $n$ in the hundreds or thousands),
each described by the values of $p$ {\em features} or {\em markers},
e.g., SNPs (think of $p$ as an order of magnitude larger than $n$).
Features have slightly different probabilities 
depending on which population the individual belongs to.
For the rest of this section, we focus on $k=2$.
We defer the discussion of the more general model with $k > 2$ in 
Sections~\ref{sec::background} and \ref{sec::kmeansintro}.
Denote by $\Delta^2$ the $\ell_2^2$ distance between two  population 
centers (mean vectors), namely, $\mu^{(1)}, \mu^{(2)} \in \R^p$, where
$\mu^{(i)} = (\mu^{(i)}_1, \ldots, \mu^{(i)}_p), \; i =1, 2$, 
\ben
\label{eq::Delta}
\Delta^2 := \twonorm{\mu^{(1)} -\mu^{(2)}}^2 =   \sum_{k=1}^p (\mu^{(1)}_k - \mu^{(2)}_k)^2 \quad
\text{ and } \quad \gamma := \Delta^2/p.
\een 
Our goal is to use these features to classify the individuals according to 
their population of origin.
The objective we consider is to minimize the total data size $D=n p$
needed to correctly classify the individuals in the sample as a 
function of the ``average quality'' $\gamma := \Delta^2/p$ of the 
features; cf. \eqref{eq::Delta}.
Let $A =(a_{ij})$ be an $n \times n$ symmetric random matrix, 
where for $1 \le i, j \le n$, $a_{ij} = a_{ji} >0$ denotes the edge 
weight, also known as {\it the dissimilarity score}, between nodes $i$
and $j$, computed from the individuals’ bit vectors.

For ease of discussion,
let the edge weight equal the Hamming distance between the bit vectors of the two 
endpoints. Recall the weight of a cut is the sum of non-negative weights across all 
edges in the cut.
It was previously shown that, in expectation, among all balanced cuts 
in the complete graph formed among $n$ vertices (sample points), the 
cut of maximum weight corresponds to the correct partition of the $n$
points according to their distributions in the balanced case with two populations ($n_1 =
n_2  = n/2$); See, for example~\cite{Zhou06}.
Under suitable conditions, the statement above also holds with high 
probability (w.h.p.)~\citep{CHRZ07,Zhou06}. 

In other words, in the context of population clustering,
~\cite{CHRZ07} and~\cite{Zhou06} show that
one can use a random instance of the integer quadratic program:
\ben
\label{eq::graphcut}
(Q) && 
\text{maximize} \quad x^T A x \quad  \text{ subject to} \quad x \in  \{-1, 1\}^n
\een
to identify the correct partition of nodes according to their population
of origin w.h.p. in the balanced case, as long as the data size $D=np$ is sufficiently large
and the separation metric satisfies $\Delta^2 = \Omega(\log 
n)$. The integer quadratic program~\eqref{eq::graphcut} is NP-hard~\citep{Karp72}.
In a groundbreaking paper,
~\cite{GW95} show that one can use a semidefinite program (SDP) as a
relaxation to solve (Q) approximately.

\subsection{Estimators and convex relaxation}
\label{sec::estimators}
In the present setting, we study three semidefinite programs related
to the graph cut problem~\eqref{eq::graphcut} and provide a unified analysis;
cf. Theorems~\ref{thm::SDPmain},~\ref{thm::exprate},~\ref{thm::YYaniso}
and~\ref{thm::SVD}.
Formally, suppose we are given a data matrix
$X \in \R^{n \times p}$ with samples from two populations $\MC_1,
\MC_2$, where
\ben 
\label{eq::Xmean}
\forall i \in  \MC_g,  \quad \E (X_{i j}) = \mu^{(g)}_{j},  \; j \in
[p], \;
\text{ where }  \; \mu^{(g)} = (\mu^{(g)}_1, \ldots, \mu^{(g)}_p) \in \R^p, \forall g \in \{1, 2\},
\een
and the sizes of clusters $\abs{\MC_{\ell}} =: n_{\ell}, \forall \ell$ may not be 
the same.  Here we denote by $[p]$ the set of integers $\{1, \ldots,
p\}$.
Our goal is to estimate the group membership
vector 
\ben
\label{eq::u2}
u_2 \in \{-1,1\}^n \; \text{ such that} \;
{u}_{2,j} = 1 \; \text{ for } \; j \in \MC_1 \;  \text{ and } \;
{u}_{2,j} =- 1 \; \text{ for } \; j \in \MC_2.
\een
\noindent{\bf Semidefinite relaxations.}
To estimate the group membership vector ${u}_2 \in \{-1, 1\}^n$, 
we consider the following semidefinite optimization problem: for a
symmetric matrix $A$,
\ben 
\label{eq::sdpmain}
\text{(SDP) } && \text{maximize}
\; \;  \ip{A, Z} \quad  \text{subject to} \quad Z \succeq 0, 
\; I_n \succeq \diag(Z),
\een
where $I_n$ denotes the  identity matrix and $A \in \R^{n \times n}$
will be specified in the sequel.
Here $A \succeq B$ means that $A - B \succeq 0$ and the inner product
of matrices $\ip{A, B} = \tr(A^T B)$.
Hence $Z \succeq 0$ indicates that the matrix $Z$ is constrained to be 
positive semidefinite.
Let $\vecone_n^T = [1, \ldots, 1] \in \R^n$ denote a vector of all $1$s.
Let  $E_n  := \vecone_n \vecone_n^T$.
As in many statistical problems,  one simple but critical step is 
to first obtain the centered data $Y$. 
Loosely speaking, this procedure is called ``global 
centering'' in the statistical literature; See for example~\cite{horns19} and references therein.
In the present work, to find a convex relaxation, we will first
construct a matrix $Y$ by subtracting the sample mean
$\hat{\mu}_n := \inv{n} \sum_{i=1}^n  X_i \in \R^{p}$ from each row
vector $X_i$ of the data matrix $X$:
\ben 
\label{eq::defineY}
Y & = & X - P_1 X, \quad \text{ and } \quad P_1 :=  \vecone_n 
\vecone_n^T/n = E_n/n \text{ is a projection matrix}. 
\een 
Given matrix $Y$, with row vectors $Y_1, \ldots, Y_n \in \R^{p}$,
where $Y_i = X_i -\hat{\mu}_n$,  we construct: 
\ben 
\label{eq::defineAintro}
A & := & Y Y^T -\lambda (E_n - I_n), \;\; \text{ where} \; \lambda  =
\frac{2}{n(n-1)}\sum_{i<j} \ip{Y_i, Y_j}.
\een
More explicitly, (SDP)~\eqref{eq::sdpmain} with $A$ as
in~\eqref{eq::defineAintro} is equivalent to the optimization problem (SDP1),  
\ben 
\label{eq::hatZintro}
\text{(SDP1)} && \text{maximize}
\; \;  \ip{YY^T - \lambda E_n,  Z}  \quad  \text{subject to} \quad 
Z \succeq 0,  \diag(Z) = I_n.
\een
See Proposition~\ref{prop::optsol} for the equivalence of (SDP) and (SDP1). 
Since all possible solutions of (Q) are feasible for (SDP), 
the optimal value of (SDP) is an upper bound on the optimal value of 
(Q). We refer to~\cite{GW95} for earlier works in 
the design and analysis of approximation algorithms for MAX CUT.
As we will show, centering the data matrix $X$ 
plays a crucial role not only in the design and analysis of 
our convex optimization algorithms, but also in understanding the 
sample size lower bounds for partial recovery of the clusters for each algorithm. 

\noindent{\bf Variations for balanced partitions.}
For balanced partitions, we propose the following new estimator {\bf BalancedSDP},
which is a variation of (SDP1):
\ben 
\label{eq::sdpball}
\quad \quad \text{(BalancedSDP)}: && \text{maximize} \; \;  \ip{YY^T, Z} \\
\nonumber
&& \text{subject to} \; \; Z \succeq 0, \; \; I_n = \diag(Z),\; \text{ and } \ip{E_n, Z} =0.
\een
Here we add one more constraint that $\tr(E_nZ) =\vecone_n^T Z
\vecone_n =0$ and hence $\vecone_n$ is the eigenvector corresponding
to the smallest eigenvalue $\lambda_{\min}(Z) =0$ of $Z \succeq 0$.

\subsection{Contributions}
\label{sec::contrib}
Our work fills an important gap in the formulation and analysis of 
the MAXCUT-based SDP~\eqref{eq::sdpmain} as well as 
in understanding its connection to spectral methods~\citep{BCFZ09} and
the $k$-means criterion in clustering.
For two numbers $a, b$, $a \wedge b := \min(a, b)$ and 
$a \vee b := \max(a, b)$.
In this paper,  we propose using the matrix $\tilde{A}:= YY^T -
\lambda E_n$~\eqref{eq::hatZintro} (resp. the Gram matrix $YY^T$)
instead of $X X^T$ as the  input to (SDP1) (resp. BalancedSDP) as
convex relaxations of MAX CUT. We make the following contributions:
(a) The global centering of the data matrix $X$ enables sharp
concentration of measure bounds to be derived in
Theorems~\ref{thm::YYnorm} and~\ref{thm::YYaniso},
directly leading to the polynomial-rate misclassification guarantees
for (SDP1) and spectral clustering in Theorems~\ref{thm::SDPmain} and~\ref{thm::SVD};
(b) This innovative approach enables a simultaneous, transparent, and
unified global and local analysis framework, resulting in an exponentially 
decaying error bound in $s^2$ for both aforementioned semidefinite
programs in Theorems~\ref{thm::exprate} and~\ref{thm::expratenew},
respectively, where $s^2 \asymp (p \gamma) \wedge (n p \gamma^2)$.
This exponentially decaying error bound implies that when $s^2 = \Omega(\log n)$, perfect recovery of the cluster structure is achieved;
(c)
We demonstrate statistical convergence and computational feasibility  
even in the low-SNR (signal-to-noise ratio) regime when $s^2 = o(\log
n)$;
(d) The design and  analysis of the semidefinite programs, namely, (SDP1)
and (BalancedSDP), also shed light on the important issue of debiasing.
In particular, the remarkably good solution to balanced MAX CUT as 
shown in Theorem~\ref{thm::expratenew} is obtained without any
explicit debiasing.

To the best of our knowledge, this is the first time such results are 
obtained for semidefinite relaxations of MAX CUT in population
clustering. 
Previously, the spectral algorithms in~\cite{BCFZ09} partitioned
sample points based on the top $k$ eigenvectors of the Gram matrix
$XX^T$, following an idea that goes back at least to~\cite{Fie1973}.
In contrast, the spectral analysis in this paper uses the Gram matrix 
$YY^T$ based on centered data, and directly improves upon the results 
in~\cite{BCFZ09} by removing the lower bound on $n$.
Importantly,
through this design choice of global centering of the data matrix, we
not only overcome technical obstacles but are also able to make
meaningful connections to the semidefinite relaxations for the
$k$-means criterion by~\cite{PW07}, as we will show in
Section~\ref{sec::kmeansintro}.

\noindent{\bf Organization.}
The rest of the paper is organized as follows.
Section~\ref{sec::background}
provides background, describes our global and local theory, and
presents preliminary results.
In Section~\ref{sec::theory},  we present
our main theoretical results in Theorems~\ref{thm::SDPmain},~\ref{thm::exprate},
~\ref{thm::expratenew},~\ref{thm::reading}, and~\ref{thm::SVD},  respectively.
We summarize the relevant related work  in Section~\ref{sec::related}.
In Section~\ref{sec::exprate}, we present a proof sketch for
Theorems~\ref{thm::exprate} and~\ref{thm::expratenew}.
In Section~\ref{sec::reduction}, we state a reduction principle for 
proving the new concentration bounds on $\norm{YY^T -\E  YY^T}_2$. 
In Section~\ref{sec::kmeansintro}, we highlight the connections and key 
differences between our approach and convex relaxations based on 
the $k$-means criterion. 
In Section~\ref{sec::experiments}, we present numerical examples, where
we compare three algorithms based on (SDP1), BalancedSDP, 
and spectral clustering, and show they have similar trends as predicted by the SNR. 
We prove Theorems~\ref{thm::SDPmain},~\ref{thm::reading},~\ref{thm::exprate},~\ref{thm::expratenew},~\ref{thm::SVD},
and their corollaries in
Sections~\ref{sec::proofSDPmain},~\ref{sec::proofJR}, and~\ref{sec::proofexprate} to~\ref{sec::biasredo},  respectively.
We conclude in Section~\ref{sec::conclude}.
We place additional technical proofs  in the Appendix.
This paper combines two arXiv preprints ~\cite{Zhou23a} and ~\cite{Zhou24a}.

\noindent{\bf Notation.}
Let $\Ball_2^n$ and $\Sp^{n-1}$ be the unit Euclidean ball and the 
unit sphere in $\R^n$, respectively.
Let $e_1, \ldots, e_n$ be the canonical basis of $\R^n$.
For a set $J \subset [n]$, denote
$E_J = \Span\{e_j: j \in J\}$.
For a vector $v \in \R^n$, we use $v_{J}$ to denote the subvector
$(v_j)_{j \in J}$.
Let $\onenorm{v} := \sum_{j} \abs{v_j}$ and  $\twonorm{v} := (\sum_{j} v^2_j)^{1/2}$. 
For a symmetric matrix $M$, let $\lambda_{\max}(M)$ and $\lambda_{\min}(M)$ be the largest and smallest eigenvalues of $M$,
respectively. The operator norm is defined as $\twonorm{M}=\sqrt{\lambda_{\max}(M^T M)}$.
For a matrix $B= (b_{ij})$ of size $n \times n$, let $\mvec{B}$ be
formed by concatenating the columns of $B$ into a vector of size
$n^2$.
Denote by $B_1^{n \times n} = \{(b_{ij}) \in \R^{n \times n} : \sum_{i,j} \abs{b_{ij}} \le 1\}$ the unit $\ell_1$-ball over $\mvec{B}$.
Let $\onenorm{B} =  \sum_{i,j=1}^n |b_{ij}|$
and $\fnorm{B} = (\sum_{i, j} b_{ij}^2)^{1/2}$.
Let $\norm{A}_{\max} = \max_{i,j} |a_{ij}|$.
Let $\diag(A)$ and $\offd(A)$ be the diagonal and off-diagonal parts of matrix $A$, respectively. 
We write $a \asymp b$ if $ca \le b \le Ca$ for some positive absolute
constants $c, C$ that  are independent of $n, p$, and $\gamma$.
We write $f = O(h)$ or $f \ll h$ if $\abs{f} \le C h$ for some absolute constant
$C< \infty$, and $f=\Omega(h)$ or $f \gg h$ if $h=O(f)$.
We write $f = o(h)$ if $f/h \to 0$ as $n \to \infty$.
In this paper, $C, C_1, C_2, C_3, C_4, c, c', c_1, \ldots$ denote various
absolute positive constants that may
change line by line.

\section{Approach and background}
\label{sec::background}
Our work is motivated by two threads of research in combinatorial 
optimization and in community detection to reformulate the MAX CUT 
problem~\eqref{eq::graphcut} and its convex relaxations (SDP1 and
BalancedSDP) in clustering.
The present work is inspired by 
the original combinatorial analysis in~\cite{Zhou06}
and~\cite{CHRZ07}, as well as the novel ideas in~\cite{GV15} to revisit the
MAXCUT-based graph partition problem~\eqref{eq::graphcut}.
While~\cite{CHRZ07} and~\cite{Zhou06} originally proposed the integer 
quadratic program~\eqref{eq::graphcut} in the context of population clustering, 
their focus was on the formulation of (Q) with various choices of $A$
rather than using computationally efficient algorithms such as
the SDP relaxation~\eqref{eq::sdpmain} to solve~\eqref{eq::graphcut}.
The global analysis framework for the semidefinite
relaxation~\citep{GV15} was set in the context of community detection
in sparse networks, where $A$ represents the adjacency matrix
of a random graph.

\citet{GV15} study the semidefinite
relaxation of the integer program~\eqref{eq::graphcut}, where an $n \times n$ 
random matrix $A$ (observed) is used to replace the hidden
static $R$ in the original integer quadratic
problem~\eqref{eq::quadratic} such that $\E A = R$:
\ben 
\label{eq::quadratic}
\text{maximize} \; \; x^T R x \quad \text{subject to} \; \; x 
\in  \{-1, 1\}^n. 
\een
The proof strategy of~\cite{GV15} is to apply Grothendieck's
inequality to the random error $A - \E A$ rather than to the original
matrix $A$ as considered in earlier literature:
bounding the cut norm of $A - \E A$ plays an important role in
deriving an error bound for $\hat{Z} -u_2 u_2^T$ in the Frobenius
norm, where $\hat{Z}$ is an optimal solution to (SDP)
\eqref{eq::sdpmain}; cf. \eqref{eq::u2}.
We refer to this approach as the {\it global analysis}, following~\cite{CCLN21}.

\subsection{Roadmap}
In the present work, we focus on the regime where partial recovery (rather than exact
recovery) is possible. We first develop a global theory by centering
the data and constructing an ideal reference matrix, which via Grothendieck’s inequality and
Davis–Kahan bounds yields universal polynomial-rate misclassification
guarantees for three algorithms (SDP1, BalancedSDP, and spectral
clustering) under minimal signal conditions.

The important distinction between the present work and that considered 
by~\cite{GV15} is that to estimate the group membership  vector ${u}_2$~\eqref{eq::u2}, we replace the random adjacency matrix  arising from stochastic block 
models with an instance of a symmetric matrix 
$A$~\eqref{eq::defineAintro} computed from the centered data matrix $Y$ as in~\eqref{eq::defineY}. 
However, unlike the setting in~\cite{GV15},
$\E A \neq R$, resulting in a bias that needs to be corrected.
Moreover, with the new concentration of measure bound on
$\norm{YY^T -\E YY^T}_2$ in Theorem~\ref{thm::YYaniso}, we can
simultaneously analyze (SDP1) in Theorem~\ref{thm::SDPmain} as well as
the spectral algorithm in Theorem~\ref{thm::SVD}.
We are also able to explain the theoretical underpinning of the observed 
concentration of measure phenomena in high
dimensions~\citep{Zhou06,BCFZ09}, simultaneously for the spectral
method and the semidefinite optimization goals, using global and local analyses, where the input is now based on the Gram matrix $YY^T$ computed 
from centered data.
With local analyses, we prove an exponentially decaying error in Frobenius norm for $\hat{Z} -u_2 u_2^T$, where $\hat{Z}$ is an optimal solution to (SDP1) 
(resp. BalancedSDP)  with respect to the signal-to-noise ratio (SNR)
in Theorem~\ref{thm::exprate} (resp.~\ref{thm::expratenew}).
We first present the technical ingredients and important ideas
for proving these results, while highlighting connections to and key 
distinctions from existing work.
We will briefly introduce the $k$-means criterion, 
where $k \ge 2$, in Section~\ref{sec::kmeans}. 
We defer further discussion to Sections~\ref{sec::theory}
to~\ref{sec::kmeansintro}.

\subsection{Technical ingredients}
We now elaborate on the construction of a reference matrix and the 
tools for deriving the large deviation bound on $\norm{YY^T -\E
  YY^T}_2$, while deferring the discussion of bias and local analyses to Sections 
\ref{sec::LLD},~\ref{sec::oracle},~\ref{sec::proofJR},~\ref{sec::exprate}, 
and~\ref{sec::kmeansintro}.
Denote by $w_{\min} := \min_{j =1, 2} w_j$ and $w_{\max} := \max_{j =1, 2} w_j$. 
We use $n_{\min}  := n w_{\min}$ and $n_{\max} := n w_{\max}$ to 
represent the size of the smallest and the largest cluster, respectively. We consider  \emph{$w_{\min}$ as a constant} throughout 
this work.

Let $\M^{+}_{G}=  \{Z: Z \succeq 0,  I_n \succeq \diag(Z)\}$ and
\ben 
\label{eq::moptintro}
\M_{\opt}
& = &
\{Z: Z \succeq 0,  \diag(Z) = I_n\} \subset \M^{+}_{G}  \subset [-1, 1]^{n \times n}.
\een
\noindent{\bf Construction of a reference matrix.}
To take this global approach, upon centering, we crucially
construct a static reference matrix $R:= \E(Y) \E(Y)^T$ with the desired properties.
Then, $Z^* =u_2 u_2^T \in \M_{\opt}$, for the convex set
$\M_{\opt}$ as in~\eqref{eq::moptintro}
will maximize the reference objective function
$\ip{R,  Z}$
among all $Z \in [-1,1]^{n   \times n}$, and naturally among all $Z
\in \M_{\opt}$; cf.  Lemma~\ref{lemma::ZRnormintro}.

Recall the overall goal of convex relaxation is to:
(a) estimate the solution of the  integer quadratic
problem~\eqref{eq::quadratic} with an appropriately chosen reference
matrix $R$ such that solving~\eqref{eq::quadratic} will recover the
cluster {\it exactly}; and (b) choose the convex set $\M^{+}_{G}$
(resp. $\M_{\opt}$~\eqref{eq::moptintro})
so that the semidefinite relaxation of the static
problem~\eqref{eq::quadratic} is tight.
This means that when we replace $A$  (resp. $\tilde{A} := YY^T -
\lambda E_n$) with $R$ in  (SDP)~\eqref{eq::sdpmain} (resp. SDP1~\eqref{eq::hatZintro}),
we obtain a solution $Z^{*}= u_2 u_2^T$ for the group membership
vector $u_2 \in \{-1, 1\}^n$ in~\eqref{eq::u2},  which can then be used to recover the
clusters exactly.
\begin{lemma}[\bf The reference matrix]
    \label{lemma::ZRnormintro}
W.l.o.g., suppose the first $n_1$ rows of 
$X$ belong to $\MC_1$ and the rest belong to $\MC_2$. 
Let $w_1 := n_1 /n$ and $w_2 := n_2 /n$.
We construct
\ben
\label{eq::Rtilt}
 R  &:= &
 \E(Y) \E(Y)^T 
 =p \gamma \left[\begin{array}{cc}
  w_2^2 \vecone_{n_1} \otimes \vecone_{n_1} &- w_1 w_2 \vecone_{n_1}
\otimes \vecone_{n_2 } \\
 - w_1 w_2 \vecone_{n_2} \otimes \vecone_{n_1}  & w_1^2 \vecone_{n_2}
                                                  \otimes \vecone_{n_2}\end{array}
\right],
\een
for $Y$ as defined in~\eqref{eq::defineY}.
Then $Z^{*} = \argmax_{Z \in \M_{\opt}}\ip{R,  Z} = u_2 u_2^T$ for
$\M_{\opt}$ as in~\eqref{eq::moptintro}. 
\end{lemma}

\noindent{\bf Anisotropic subgaussian random vectors as noise.}
First, recall for a random
variable $X$, the $\psi_2$-norm of $X$, denoted by
$\norm{X}_{\psi_2}$, is $\norm{X}_{\psi_2} = \inf\{t >  0\; : \; \E
\exp(X^2/t^2) \le 2 \}$.
\begin{definition}
  \label{def::isotropic}
A random vector $W \in \R^m$ is called subgaussian if the one-dimensional marginals $\ip{W, h}$ are subgaussian random 
variables for all $h \in \R^{m}$:
(1) $W$ is called isotropic if for every $h \in \Sp^{m-1}$,
$\E \abs{\ip{W, h}}^2= 1$;
(2)  $W$ is $\psi_2$ with a constant $\alpha$ if for every $h \in
\Sp^{m-1}$, $\norm{\ip{W, h}}_{\psi_2} \le \alpha$.
The subgaussian norm of $W \in \R^m$ is denoted by
  \ben
  \label{eq::Wpsi}
  \norm{W}_{\psi_2}  := \sup_{h \in \Sp^{m-1}} \norm{\ip{W, h}}_{\psi_2}.
  \een
\end{definition}
Definition~\ref{def::isotropic} is standard;
See for example Definition 1.1~\cite{RZ17} and Definition 
3.4.1~\cite{Vers26}. 
Denote by $\Z = X - \E X = [\Z_1, \ldots, \Z_n]^T $ the noise matrix 
with  independent, mean-zero, subgaussian random vectors
$\Z_1, \ldots, \Z_n \in \R^{p}$.
We now state the data generative process for $\Z \in \R^{n \times p}$
in Definition~\ref{def::WH} that we use throughout this paper.
Crucially, we prove the concentration bound in Theorem~\ref{thm::ZHW},
which corresponds to the Hanson-Wright inequalities~\citep{HW71,RV13},
but for anisotropic subgaussian vectors.
\begin{definition}[\bf Data generative process]
\label{def::WH}
   Let $H_1, \ldots, H_n$ be deterministic $p \times m$ matrices, 
   where we assume that $m \ge p$ for simplicity; Let $\sigma_{\max}^2 = \max_{i} \shnorm{H_i}_2^2$. 
   Suppose random matrix 
   $\BW =[W_1, \ldots, W_n]^T = (w_{jk})\in \R^{n \times m}$, where
   $W_1, \ldots, W_n \in   \R^{m}$ are independent, isotropic, subgaussian random vectors, each with independent, mean-zero coordinates such that
   $\cov(W_j) :=\E  W_j W_j^T = I_m, \forall j \in [n]$  and $\E[w_{jk}] = 0, \forall j, k$. Denote by
\ben 
\label{eq::Wpsi2}
K := \max_{j,k} \norm{w_{jk}}_{\psi_2} \quad \text{ and } \quad
C_0 := \max_{j} \norm{W_j}_{\psi_2}, \quad \text{ where } \quad
K \le C_0 \le C K
\een
for an absolute constant $C$; See Lemma 3.4.2~\citep{Vers26}. 
Suppose 
\bens
\forall j \in [n], \quad \Z_j =  H_j W_j, \; \text{ where } \; H_j \in \R^{p \times 
  m} \; \text{ and} \quad  0< \twonorm{H_j} \le \sigma_{\max} < \infty.
\eens
\end{definition}

\begin{theorem}[\bf Hanson-Wright inequality for anisotropic subgaussian vectors]
  \label{thm::ZHW}
 Let $H_1, \ldots, H_n$ be deterministic $p \times m$ matrices, 
 where $m \ge p$. 
 Let $\Z^T_1, \ldots, \Z^T_n \in \R^{p}$ be row vectors of $\Z$.
 We generate $\Z = [\Z_1, \ldots, \Z_n]^T \in \R^{n \times p}$
 according to Definition~\ref{def::WH}. Let $K$ and $C_0$ be as defined
 in~\eqref{eq::Wpsi2}. Then we have for any $A = (a_{ij}) \in \R^{n \times n}$ and $t > 0$,
\ben
\nonumber
\prbb{|\sum_{i=1}^n  \sum_{j \neq i}^n \ip{\Z_{i}, \Z_{j}} a_{ij}| > t}
& \le & 2 \exp \big(- c\min\big(\frac{t^2}{p
    (K\sigma_{\max})^4  \shnorm{A}_F^2}, \frac{t}{(K\sigma_{\max})^2
    \shnorm{A}_2} \big) \big), \\
\label{eq::Zpsi0}
 \text{ where} \; \forall j \in [n],\; 
\norm{\Z_j}_{\psi_2} & := & \sup_{v \in \Sp^{p-1}} \norm{\ip{\Z_j, v}}_{\psi_2}
\le  \norm{W_j}_{\psi_2} \twonorm{H_j} \le C_0 \sigma_{\max}. 
\een
\end{theorem}
Theorem~\ref{thm::ZHW} and its proof may be of independent interest; 
cf. Section~\ref{sec::opcorrelated}. See Exercise 6.8 by~\citet{Vers26} for a related result.
In the context of Theorem~\ref{thm::ZHW}, $H_j$'s are allowed to
repeat, e.g., across rows from the same cluster $\MC_i$ for $i=1, 2$. 
Lemma~\ref{lemma::twogroup} characterizes the two-group design matrix
covariance structures considered in
Theorems~\ref{thm::YYaniso},~\ref{thm::SDPmain},~\ref{thm::exprate},~\ref{thm::expratenew}
and~\ref{thm::SVD}, using the data generative process in 
Definition~\ref{def::WH}.
We prove Lemma~\ref{lemma::twogroup} in Section~\ref{sec::prooftwogroup}.
\begin{lemma}[\bf Two-group subgaussian mixture model]
\label{lemma::twogroup}
  Denote by $X$ the two-group design matrix considered
  in~\eqref{eq::Xmean}.
Let random matrix $\BW =[W_1, \ldots, W_n]^T = (w_{jk})\in \R^{n
  \times m}$ and $C_0$ be the same as in Definition~\ref{def::WH} with
$C K \ge C_0 \ge K$ where $K := \max_{j,k} \norm{w_{jk}}_{\psi_2}$ and
$C$ is an absolute constant.
Let $H_i \in \R^{p \times m}, i =1, 2$ be deterministic matrices with
$0< \shnorm{H_i}_2 \le \sigma_{\max} < \infty, \forall i$.
Suppose for $j \in \MC_i \subset [n]$, $\E X_j = \mu^{(i)}$
and $\Z_j := X_j - \E X_j = H_i W_j \in \R^{p}$, where $i =1, 2$.
Then for both $i \in \{1, 2\}$, 
\bens 
\label{eq::rowcov}
\forall j \in \MC_i,\quad 
\cov(\Z_j) := \E \Z_j \Z_j^T =  H_i H_i^T =: 
\Sigma_i
\quad \text{ where } \quad \tr(\Sigma_i) =: V_i =\fnorm{H_i}^2.
\eens
Moreover, $\Z_1, \ldots, \Z_n \in \R^{p}$ are independent, mean zero, 
subgaussian random vectors with covariance $\cov(\Z_j)$ satisfying
\eqref{eq::covZ1} and \eqref{eq::covZ2}: 
\ben
\label{eq::covZ1}
\forall j \in [n], \;
\norm{\ip{\Z_j,   x}}_{\psi_2} & \le & C_0 \norm{\ip{\Z_j, x}}_{L_2}
\; \text{for any } \; x \in  \R^{p},\;\text{ where }\\
\label{eq::covZ2}
\norm{\ip{\Z_j, x}}^2_{L_2}  & :=& \E \ip{Z_j, x}^2 = x^T (\E \Z_j 
\Z_j^T) x = x^T \cov(\Z_j) x.
\een
\end{lemma}

\subsection{Concentration of measure,  local analysis and  debiasing}
\label{sec::LLD}
Theorem~\ref{thm::YYaniso} is an important result in this paper.
The concentration bounds on the operator norm of $YY^T - \E 
  YY^T$ imply that, up to a constant factor,  the same bounds also 
  hold for $A - \E A$ (and $B - \E B$, cf.~\eqref{eq::defineBintro}). 
As a result, the error bound in Theorem~\ref{thm::YYaniso}
is crucial in the global analysis leading to 
Theorem~\ref{thm::SDPmain}, as well as our analysis of a 
spectral clustering algorithm in Theorem~\ref{thm::SVD}. 
\begin{theorem}[\bf Concentration for anisotropic design]
  \label{thm::YYaniso}
  Let $X$ and $C_0$ be as in Lemma~\ref{lemma::twogroup} and 
  $Y$ be as in \eqref{eq::defineY}.
  Let $\Delta^2 = \twonorm{\mu^{(1)}-\mu^{(2)}}^2 = p \gamma$.
Let
  \ben 
\label{eq::definemu}
\mu :={\mu^{(1)}-\mu^{(2)}/}{\sqrt{p \gamma}} \in \Sp^{p-1}.
\een
Suppose  for some parameter $\xi$ satisfying $w_{\min}^2/16 > \xi =
\Omega(1/n_{\min})$ and $\sigma^2_{\max}  := \max_{i} \twonorm{H_i}^2$,
\ben
\label{eq::NKlower}
p \gamma & \ge &  {C_2 C_0^2 \sigma_{\max}^2 }/{\xi^2}
\quad \text{ and } \quad np  \ge  {C_{3} C_0^4 \sigma_{\max}^4}/{(\gamma^2 \xi^2)}.
\een
Then with probability at least $1 - 2\exp(-c_8 n)$, we have
\ben
\label{eq::definemuYY}
\twonorm{YY^T - \E Y Y^T} & \le &
C_4 C_0 (\max_{i} \twonorm{H_i^T \mu}) n \sqrt{p \gamma} +
  C_5 C_0^2 \sigma_{\max}^2 (\sqrt{p n} \vee n)\\
\nonumber
& \le & \xi n p \gamma/6.
\een
\end{theorem}

We mention in passing that a natural and common choice for $A$ in
(SDP) in the literature is the Gram matrix $XX^T$; however, the large deviation 
bounds on $\norm{XX^T -\E XX^T}_2$ will not be sufficient to apply the
global analysis framework as in~\cite{GV15}.
We give a proof sketch of Theorem~\ref{thm::YYaniso} in Section~\ref{sec::reduction}, with full proof in Section~\ref{sec::proofYYconc}.

\noindent{\bf Local analysis.}
In the context of Lemma~\ref{lemma::twogroup}, the notion of SNR
introduced in Section~\ref{sec::contrib} can now be properly adjusted
in view of~\eqref{eq::covZ1} and~\eqref{eq::covZ2}:
\ben
\label{eq::SNR2}
\text{\bf SNR anisotropic: }   \; \; s^2 
&=& \frac{\Delta^2}{C_0^2 \sigma_{\max}^2} \wedge \frac{n p \gamma^2}
{C_0^4 \sigma_{\max}^4}, \; \text{where}  \; \sigma_{\max}^2 := \max_{j}\shnorm{\cov(\Z_j)}_2.
\een
Crucially, we also obtain a uniform control for
$\ip{Y Y^T - \E YY^T,  \hat{Z} - Z^{*} }$ over all $\hat{Z} \in \M_{\opt}$ in
the local neighborhood of $Z^{*} = u_2 u_2^T$ with high probability;
cf.  Lemma~\ref{lemma::YYlocal}.
As a result, the misclassification errors are shown to decay exponentially in the
SNR parameter $s^2$~\eqref{eq::SNR2} in the settings of
Theorems~\ref{thm::exprate},~\ref{thm::expratenew}, and
Corollary~\ref{coro::misexp} for the subgaussian design with
dependent features considered in Lemma~\ref{lemma::twogroup},
as long as $s^2$ is bounded below by a constant as in
\eqref{eq::NKlower}.
This gives rise to the notion of a \emph{local analysis}; 
we refer to~\cite{CCLN21} and
Sections~\ref{sec::exprate},~\ref{sec::proofexprate},~\ref{sec::keylemmas},
and~\ref{sec::biasredo} for further references.

\noindent{\bf Debiasing.}
Consider the model in Lemma~\ref{lemma::twogroup}.
For both balanced and unbalanced cases, we allow each population to  
have distinct covariance structures with diagonal  matrices as special 
cases.
Let $\abs{\MC_1} =n_1$ and $\abs{\MC_2} =n_2$. 
Suppose $k=2$. Let $\Z_j = X_j - \E X_j \in \R^p$, $j \in [n]$.
Let $V_i$ denote the trace of the covariance
$\cov(\Z_j) = \E \Z_j \Z_j^T$ for all $j \in \MC_i$:
\ben
\label{eq::V2intro}
V_1 = \E \ip{\Z_j, \Z_j}  \quad \forall j \in \MC_1 \;\;
\text{and}\;\;
V_2   =   \E \ip{\Z_j, \Z_j} \quad \forall j \in \MC_2.
\een
Such a discrepancy in their traces is known to cause a 
bias, which needs to be explicitly addressed;
cf.~\eqref{eq::relaxadjust}. See, for example~\cite{Royer17}.
Through the refined \emph{local analysis} in the present work, we conjecture 
that for the balanced case  ($n_1 = n_2  = n/2$), the bias 
may have already been {\em substantially reduced} due to our centering and 
adjustment steps in the construction of $A$~\eqref{eq::defineAintro}
for (SDP) and (SDP1).
This insight leads to the design of a new estimator, namely, BalancedSDP~\eqref{eq::sdpball}.
As such, we \emph{eliminate} the bounded trace-difference assumption
{\bf (A2)} in Theorem~\ref{thm::expratenew} despite not making any
explicit adjustment to the Gram matrix $YY^T$.
In this sense,
the results obtained in Theorem~\ref{thm::expratenew} for balanced partitions are stronger and more robust than those in~\cite{GV19} for $k=2$. 
See the discussion following Theorem~\ref{thm::expratenew}.
See Sections~\ref{sec::kmeansintro},~\ref{sec::keylemmas},
and~\ref{sec::biasredo} for full exposition.
We defer the discussion of SNR to Section~\ref{sec::theoryaniso} after stating 
Theorem~\ref{thm::exprate}.

\subsection{Generalization to  the $k$-means criterion}
\label{sec::kmeans}
More generally, one may consider semidefinite relaxations
for the following subgaussian mixture model with $k$ centers,
where we have $n$ observations $X_1, X_2, \ldots,  X_n \in \R^{p}$:
\ben
\label{eq::model}
X_i = \mu^{(\psi_i)} + \Z_i, \; \; i \in [n],
\een
where $\Z_1, \ldots, \Z_n \in \R^{p}$ are independent, subgaussian,
mean-zero random vectors, and $\psi_i \in \{1, \ldots, k\}$ assigns node $i$ to a group $\MC_j$ with mean $\mu^{(j)} \in \R^{p}$ for some $j \in [k]$.

\citet{GV19} establish bounds for a semidefinite program based on the
$k$-means criterion, following the proposal by~\cite{PW07}, where the
positive semidefinite matrix $Z \in \R^{n \times n}$ is constrained to
have non-negative entries. Specifically, they consider $\hat{Z} =
\argmax_{Z \in \M_k} \ip{XX^T, Z}$,
where $\M_k = \{Z \in  \R^{n \times n}: Z \succeq 0, \tr(Z) = k, Z
\vecone_n = \vecone_n, Z_{ij} \ge 0, \forall i, j\}$.
Enforcing these non-negativity constraints ($Z_{ij} \ge 0, \forall i, j$)
in $\M_k$ regularizes the convex problem to  resemble the $k$-means criterion, but it nonetheless incurs an additional computational cost.
In Section~\ref{sec::kmeansintro}, we elaborate on $k$-means 
relaxations, including SVD-based algorithms that allow for reduced 
computational complexity by eliminating both the non-negativity and 
row-sum ($Z \vecone_n = \vecone_n$) constraints. We analyze this
algorithm for $k=2$ in Theorem~\ref{thm::SVD} in  the present work.

While the theoretical results in Theorem~\ref{thm::exprate} are 
developed in the same spirit as those of~\cite{GV19} for $k=2$,
we prove them for the much simpler MAXCUT-based
SDP~\eqref{eq::sdpmain} with the constraint set $\{Z_{ii} \le 1, \forall i
\in [n]\}$, and for (SDP1) with the minimal constraint set $\{Z_{ii} =
1, \forall i \in [n]\}$, both under the condition $Z \succeq 0$.
Consequently, we reduce the problem from $O(n^2)$
linear constraints in $\M_k$ to exactly $n$ linear constraints in
$\M^{+}_{G}$ or $\M_{\opt}$, as defined in~\eqref{eq::moptintro}.
The objective of dropping the $O(n^2)$ non-negativity constraints on
the elements of $Z$ in $\M_k$ introduced by~\cite{PW07} is to reduce
the computational complexity \textbf{without compromising statistical
  convergence}.
Our theory confirms that the MAXCUT-based semidefinite relaxations,
such as (SDP1)~\eqref{eq::hatZintro} and
BalancedSDP~\eqref{eq::sdpball},
as well as the spectral clustering algorithm, provide a mathematically
rigorous framework with {\it substantially reduced constraint
  complexity}, while ensuring statistical convergence.

We emphasize that the reduction techniques used for the concentration
of measure bound in Theorem~\ref{thm::YYaniso}, as well as the
probabilistic bounds in Theorem~\ref{thm::ZHW} (and
Lemmas~\ref{lemma::YYcovcorr} and~\ref{lemma::anisoproj}), are derived
for the general $k$-means clustering problem where $\Z_1, \ldots,
\Z_n$ are independent but not identically distributed.
In particular, the spectral clustering algorithm based on the top $k-1$ eigenvectors of $YY^T$ is directly related to the relaxations for the $k$-means criterion proposed by~\cite{PW07}.
Consequently, these concentration bounds are applicable to the general
$k$-means model \eqref{eq::model}, allowing for immediate extensions
of our framework.

\section{Theory}
\label{sec::theory}
To set up the context for our analysis, we first state in
Section~\ref{sec::theoryaniso}
the main Theorem~\ref{thm::SDPmain},
under assumptions (A1) and (A2), motivating the consideration
of~\eqref{eq::SNR} as our baseline scenario.
We then state Theorems~\ref{thm::exprate} and~\ref{thm::expratenew},
followed by a discussion.
We construct an oracle SDP to enable the global and 
local analyses in Section~\ref{sec::oracle}.
We prove Theorem~\ref{thm::SDPmain} in Section~\ref{sec::proofSDPmain}.
In Section~\ref{sec::SVDthm}, we state Theorem~\ref{thm::SVD} for the
SVD-based algorithm, along with its corollary. In
Section~\ref{sec::proofJR}, we provide a proof sketch for
Theorem~\ref{thm::reading}, which is an important intermediate result
unifying Theorems~\ref{thm::SDPmain} and~\ref{thm::SVD}.
We discuss the relevant related work in Section~\ref{sec::related}.
Let $C, C', C_{1}, C_2, c_0, c, c', c_1, \ldots$ be absolute constants.

\subsection{Main theorems from the global and local analyses}
\label{sec::theoryaniso}
We first state assumptions (A1) and (A2).

\noindent{\bf (A1)}
Let $\Z = X - \E X$. Let $\Z_i = X_i - \E X_i, i \in [n]$ be 
  independent, mean-zero, subgaussian random vectors with independent 
  coordinates such that for all $i,j$,
  $\norm{X_{ij} - \E   X_{ij}}_{\psi_2} \le C_0.$ \\
\noindent{\bf (A2)}
The two distributions have variance profile discrepancy bounded:
\ben
\label{eq::Varprofilepreview}
 \abs{V_1 - V_2}  \le \xi 
n p \gamma /3 \; \;  \text{for $V_i$ as in~\eqref{eq::V2intro}},
\;\; \text{ where } \; \; 1/2> \xi = \Omega(1/n_{\min}).
\een
As mentioned, Theorem~\ref{thm::YYnorm} is an important 
intermediate result for proving Theorem~\ref{thm::SDPmain}. 
The anisotropic case is stated in Theorem \ref{thm::YYaniso}. 
\begin{theorem}
\label{thm::SDPmain}
 Suppose for $j \in \MC_i$, $\E X_j = \mu^{(i)}$, where $i =1, 2$.
 Let $\hat{Z}$ be a solution of (SDP1).
 Let $K_G \le \frac{\pi}{2 \ln(1+\sqrt{2})} \le 1.783$.
 Suppose (A1) and (A2) hold, and 
 \ben 
  \label{eq::kilo}
 p \gamma = \Delta^2 \ge {C' C_0^2}/{\xi^2}  \quad  \text{ and } \quad
 n p \ge {C C_0^4}/{(\xi^2 \gamma^2)},
 \een
for some  absolute constants  $C, C'$, where $\xi$ is the  same as in~\eqref{eq::Varprofilepreview}.
Then with probability at least $1-2 \exp(-c n)$, for ${u}_2$
as in \eqref{eq::u2},
\ben
\label{eq::hatZFnorm}
\shnorm{\hat{Z} - u_2 u_2^T}^2_F \le 2 \shnorm{\hat{Z} - u_2 u_2^T}_1
\le 2  \delta n^2 \; \text{ where } \delta =  {2 K_G  \xi}/{w_{\min}^2}.
\een
\end{theorem}

\begin{theorem}[\bf Design with independent entries~\cite{Zhou23a}]
 \label{thm::YYnorm}
Suppose that $(A1)$ and~\eqref{eq::kilo} hold with 
$\max_{j, k} \norm{X_{jk} - \E X_{jk}}_{\psi_2} \le C_0$.
Then, with probability at least $1 - 2\exp(-c_7 n)$,
  \bens
\twonorm{YY^T - \E Y Y^T}
& \le &
C_2 C_0^2 (\sqrt{p n} \vee n) + C_3 C_0 n \sqrt{p \gamma}
\le \inv{6}\xi n p \gamma. 
\eens
\end{theorem}

\noindent{\bf Remarks on the misclassification error rate.}
The parameter $\xi$ in (A2)  is understood to be set as a fraction of
the misclassification error rate $\delta$ in Theorem~\ref{thm::SDPmain}, 
which in turn is assumed to be lower bounded: $1> \delta =
\Omega(1/n)$, since we focus on the partial recovery of the group
membership vector $u_2$.
To control the misclassification error,
the parameters $(\delta, \xi)$ in Theorem~\ref{thm::SDPmain} can be chosen to
satisfy the following relations:
\ben
\label{eq::trend}
\delta & = &  {2 K_G \xi}/{w_{\min}^2} \asymp {2 K_G}/{(w_{\min}^2 \sqrt{s^2})} 
\;\; \text{ and } \; \xi^2 \asymp  {1}/{s^2},  \; \text{ where} \; \\ 
\label{eq::SNR}
s^2 & := & ({p \gamma}/{C_0^2}) 
\wedge ({n p \gamma^2}/{C_0^4}) \quad (\text{\bf SNR isotropic } )
\een
in view of~\eqref{eq::kilo} and~\eqref{eq::hatZFnorm}.
Roughly speaking, $\xi^2$ in~\eqref{eq::trend} is chosen to be inversely 
proportional to $s^2$, resulting in the misclassification error rate 
$\delta \asymp \xi/w_{\min}^2$ being inversely proportional to the
square root of $s^2$, as we show in  Corollary~\ref{coro::misclass}.
\begin{corollary}[$O(\xi n/w_{\min}^2)$ misclassified vertices]
  \label{coro::misclass}
  Let $\hat{x}$ denote the eigenvector of $\hat{Z}$ corresponding
  to the largest eigenvalue, with $\twonorm{\hat{x}}  = \sqrt{n}$.
  Then in both settings of Theorem~\ref{thm::SDPmain}, we
  have with probability at least $1-2 \exp(-c n)$,
  $\min_{\alpha = \pm 1} \twonorm{\alpha \hat{x} - u_2}^2 \le 
  \delta' n  = \delta'  \twonorm{u_2}^2$, where  $\delta' \le {32  K_G \xi}/{w_{\min}^2}$.
  Moreover, the signs of the coefficients of $\hat{x}$ correctly
  estimate the partition of the vertices into two clusters, 
  up to at most $\delta' n$ misclassified vertices.
\end{corollary}

While (A1) assumes that the random matrix $\Z$ has independent 
subgaussian entries, matching the separation
condition~\eqref{eq::kilo} and SNR~\eqref{eq::SNR},
the conclusions stated in Theorems~\ref{thm::SDPmain} and~\ref{thm::exprate} hold for the 
general two-group model considered in Lemma~\ref{lemma::twogroup},
upon adjusting~\eqref{eq::kilo}.
We emphasize that the covariance model under consideration in
Theorem~\ref{thm::YYaniso} is a special case of
Theorem~\ref{thm::ZHW}, cf. Lemma~\ref{lemma::twogroup},
whereas Theorem~\ref{thm::YYnorm} is a special case of 
Theorem~\ref{thm::YYaniso} for an isotropic design matrix under (A1). 
See Section~\ref{sec::reduction} for a discussion on \emph{covariance
  estimation}.

\noindent{\bf Local analysis.}
Our local analysis builds upon the global analysis; proving such results without the positivity constraints 
presents greater technical difficulties.

\begin{theorem}[\bf SDP1]
  \label{thm::exprate}
  Let $X$ be as in Lemma~\ref{lemma::twogroup}. 
  Let $\hat{Z}$ be a solution of (SDP1).
  Suppose~\eqref{eq::NKlower} holds.
Suppose (A2) holds, where the parameter $\xi$
in~\eqref{eq::Varprofilepreview} is chosen to be the same as that in 
\eqref{eq::NKlower}.
Then with probability at least $1-2 \exp(-c_1 n) - c_2/n^2$, 
$\onenorm{\hat{Z} - {u}_2 {u}_2^T}/n^2 \le \exp\big(-c_0 s^2 
w_{\min}^4\big)$,  where $s^2$ is as defined in~\eqref{eq::SNR2}. 
\end{theorem}

\noindent{\bf Signal-to-noise ratios.}
Let $\sigma_{\max}^2:= \max_{j} \twonorm{\cov(\Z_j)}$.
The notion of SNR~\eqref{eq::SNR} and the associated sample lower
bounds appear in~\cite{CHRZ07} and~\cite{Zhou06}, where $s^2
=\Omega(\log n)$ is required.
Our sample size lower bounds~\eqref{eq::NKlower} and~\eqref{eq::kilo}
indicate that once the distance between the two centers is bounded below by a
constant, namely, $\Delta= \sqrt{p \gamma} = \Omega(\sigma_{\max})$,
adding more sample points (individuals) to the clusters reduces the
number of features required for partial recovery.
This simultaneously mitigates the bias in the sense that 
{\bf (A2)} holds trivially in the large sample setting where $n > p$, 
since $\abs{V_1 - V_2} = O(p)$ by Definition~\ref{def::WH}. 
See Section \ref{sec::proofJR} for further discussion (cf. \eqref{eq::bias}). 

On the other hand, both the previous and current results show that even when sample 
size $n$ is small, by increasing the dimension $p$ such that the total sample size 
satisfies~\eqref{eq::kilo} or~\eqref{eq::NKlower}, we ensure partial
recovery of cluster structures using either the integer quadratic
program~\eqref{eq::quadratic} or its convex relaxation such as
(SDP)~\eqref{eq::sdpmain}.
We mention in passing that this definition of SNR in~\eqref{eq::SNR}
also coincides with the signal-to-noise ratio used in~\cite{GV19} for
isotropic design under {\bf (A1)}.
Moreover, under {\bf (A1)} and {\bf (A2)},
the exponentially decaying error bounds in both papers are
equivalent, despite the use of different SDP relaxations.
We next state our main result for balanced partitions.
\begin{theorem}[\bf BalancedSDP]
\label{thm::expratenew}
Let $X, \MC_j \subset [n]$ be as in Lemma~\ref{lemma::twogroup}, 
with $\abs{\MC_1} =\abs{\MC_2} =n/2$.
Let $\hat{Z}$ be a solution of~\eqref{eq::sdpball}.
Let $s^2$ be as in~\eqref{eq::SNR2}.
Suppose $s^2 > C$ for some absolute constant $C$.
Then with probability at least $1- c/n^2$, $\onenorm{\hat{Z} - {u}_2 
  {u}_2^T}/n^2 \le \exp(-c' s^2)$. 
\end{theorem}

As mentioned earlier, for balanced partitions, {\bf (A2)} is not 
required for an exponentially decaying error bound to hold, provided that one 
additional constraint $\langle E_n, Z \rangle = 0$ is added in the 
BalancedSDP~\eqref{eq::sdpball}. In this sense, the results that we obtain 
in Theorem~\ref{thm::expratenew} for balanced partitions are stronger 
and more robust than those in~\cite{GV19}.
In summary, (a) for the balanced case, when $n_1 = n_2$,
an arbitrary level of trace difference $\abs{V_1- V_2}$ is tolerated
through the new estimator BalancedSDP~\eqref{eq::sdpball}.
Hence, unlike~\cite{Royer17},~\cite{GV19}, and~\cite{nda22}, 
we do not need any explicit debiasing step for balanced partitions;
(b) for unbalanced partitions, we impose the bounded trace-difference
assumption (A2) in Theorem~\ref{thm::exprate}; cf.~\eqref{eq::Varprofilepreview}. 
When {\bf (A2)} is violated and $n_1 \neq n_2$, we may adopt ideas
from~\cite{Royer17},~\cite{GV19}, and the present work to make
adjustments to (SDP1) to further correct the bias,
but it comes with an additional computational cost nonetheless.
We defer a detailed comparison with~\cite{GV19} for anisotropic
design, as well as the distinctive SDP relaxations, to
Section~\ref{sec::SNRcompare}.
We provide a proof sketch for Theorems~\ref{thm::exprate}
and~\ref{thm::expratenew} in Section~\ref{sec::exprate}, with full proofs of these theorems and their corollaries in 
Sections~\ref{sec::proofexprate},~\ref{sec::keylemmas},~\ref{sec::biasredo}, 
and~\ref{sec::proofofmisclass}, respectively.

\subsection{Construction of an OracleSDP}
\label{sec::oracle}
A straightforward calculation leads to the expression of the 
reference matrix $R:= \E(Y) \E(Y)^T$ as in~\eqref{eq::Rtilt}, in view of \eqref{eq::EYpre}.
However, unlike the setting in~\cite{GV15}, 
$\E A \not= R$, resulting in a large bias.
This motivates the construction of OracleSDP~\eqref{eq::sdpB}:
\ben 
\label{eq::sdpB}
  &&  {\bf (OracleSDP) }  \quad \text{maximize} \quad \ip{B, Z}  \quad
  \text{subject to} \quad Z \in \M_{\opt} \text{ as in~\eqref{eq::moptintro}},\\
\label{eq::defineBintro}
&& \text{where} \; \;  B  :=  A - \E \tau I_n  \; \text{ for $A$ as   in~\eqref{eq::defineAintro} and }  \tau = \sum_{i=1}^n \ip{Y_i, Y_i}/n.
\een
Here, the adjustment term $\E \tau I_n$ plays no role in optimization,
since the extra trace term $\propto  \ip{I_n, Z} = \tr(Z) =n$ is a
constant function of $Z$ on the feasible set $\M_{\opt}$~\eqref{eq::moptintro}.
However, both $A$~\eqref{eq::defineAintro} and $B$~\eqref{eq::defineBintro} are
defined to bridge the gap between $YY^T$ and $R =
\E(Y)\E(Y)^T$~\eqref{eq::Rtilt}.
Lemma~\ref{lemma::EBRtilt}
states that the bias $\E B - R$ is 
substantially reduced for $B$ as in~\eqref{eq::defineBintro}, thanks 
to the adjustment term $\E \tau  I_n$, and is essentially negligible 
when $V_1 = V_2$. 
\begin{lemma}
\label{lemma::EBRtilt}
Let $X$ be as in Lemma~\ref{lemma::twogroup}. 
Suppose (A2)~\eqref{eq::Varprofilepreview} holds.
Suppose $\xi \ge \inv{2n} (4 \vee \inv{w_{\min}})$ and $n \ge 4$. 
Then $\twonorm{\E B - R}  \le 2 \abs{V_1 -  V_2} (w_1 \vee w_2) + p
\gamma/3 \le 2 \xi  n p \gamma/3$.
When $V_1 = V_2$, we have $\twonorm{\E B - R} \le  p \gamma/3$. 
\end{lemma}

We prove Lemma~\ref{lemma::EBRtilt} in Section~\ref{sec::biasLemmaproof}.
We emphasize that our algorithm solves SDP1~\eqref{eq::hatZintro}
rather than OracleSDP~\eqref{eq::sdpB}.
However, formulating the OracleSDP estimator enables both global and local
analyses by controlling  the operator norm of the bias term, namely, 
$\twonorm{\E B - R}$ as in  Lemma~\ref{lemma::EBRtilt}, and a related
quantity to be introduced in Lemma~\ref{lemma::optlocal}.
Given the OracleSDP formulation in \eqref{eq::sdpB}, the bias analysis and
the concentration bounds on $\twonorm{YY^T -\E YY^T}$ enable
simultaneous analyses of SDP1 and the closely related spectral
clustering method; cf. Theorem~\ref{thm::SVD}.

We show in Proposition~\ref{prop::optsol} that the optimization goals of (SDP1) 
and (SDP)~\eqref{eq::sdpmain} are equivalent, and both are equivalent to 
that of OracleSDP~\eqref{eq::sdpB}. 
Proposition~\ref{prop::optsol} is proved in Section~\ref{sec::oraclesupp}.
\begin{proposition}[\bf OracleSDP]
  \label{prop::optsol}
  Let $\tilde{A} := Y Y^T -\lambda E_n$. The optimal solutions $\hat{Z}$ as in \eqref{eq::sdpmain}
  using $A$ as in~\eqref{eq::defineAintro}  must have their  diagonals set to $I_n$.
Thus, the set of optimal solutions $\hat{Z} \in \argmax_{Z \in
  \M^{+}_{G} } \ip{A, Z}$ in~\eqref{eq::sdpmain}, where $\M^{+}_{G}=
\{Z: Z \succeq 0,  I_n \succeq \diag(Z)\}$,
coincide with those on the convex subset $\M_{\opt}$ as in \eqref{eq::moptintro},
\ben
\label{eq::Aquiv}
\label{eq::optsolAB}
\argmax_{Z \in  \M^{+}_{G} }  \ip{A , Z} = \argmax_{Z \in \M_{\opt}}
\ip{\tilde{A}, Z}
 = \argmax_{Z \in \M_{\opt}}  (\ip{A - \E \tau I_n, Z} ) \; \text{\bf (OracleSDP).}
\een 
\end{proposition}

\subsection{Proof of Theorem~\ref{thm::SDPmain}}
\label{sec::proofSDPmain}
For a matrix  $A =(a_{ij}) \in \R^{n \times n}$, we denote by $\norm{(a_{i j})}_{\infty\to1}$ its 
$\ell_{\infty} \to \ell_1$ norm, which is 
$$\norm{(a_{i j})}_{\infty\to1} = 
\max_{\norm{s}_{\infty} \le 1} \norm{A s}_1 = 
\max_{s, t \in \left\{-1, 1\right\}^n}
\ip{A, st^T}.$$
The concept of the cut norm plays a major role in the work of~\cite{FK99}
on efficient approximation algorithms for dense graph and 
matrix problems. We define the cut norm in Section~\ref{sec::cutnormsupp}; cf. Definition~\ref{def::cutnorm}. 
The cut norm is also essential for the arguments of~\cite{GV15} to
carry over to the present work. With proper adjustments, we adapt their methodology
to our setting and prove the first main result,
Theorem~\ref{thm::SDPmain}.

First, following~\cite{GV15}, we show that $\hat{Z}$ provides an almost optimal
solution to the reference problem if $B$ and $R$ are close; See Lemma~\ref{lemma::GVGD}.
Second, we establish that the excess risk  $\ip{R, Z^{*} - \hat{Z}}$
exhibits a non-trivial global curvature over the feasible set
$\M_{\opt}$~\eqref{eq::moptintro} at the maximizer $Z^{*}$ in
Lemma~\ref{lemma::onenorm}.

\begin{lemma}[Lemma 3.3~\cite{GV15}]
  \label{lemma::GVGD}
Let $\M_{\opt}$  be as in \eqref{eq::moptintro}.
Let $\hat{Z} =\argmax_{Z \in \M_{\opt}} \ip{B, Z}$ for any given $B$
and $Z^{*} := \argmax_{Z \in  \M_{\opt}} \ip{R, Z}$.
Then
\ben
\label{eq::upperZZR}
0 \le  \ip{R, Z^{*} - \hat{Z}} \le  2 K_G \norm{B -R}_{\infty \to
  1}, \; \text{where} \; K_G <\frac{\pi}{2 \ln (1 + \sqrt{2})} \le 1.783.
\een
\end{lemma}

\begin{lemma}[Excess risk lower bound]
\label{lemma::onenorm}
Let $R$ be as in~\eqref{eq::Rtilt}.
Then for every $Z \in \M_{\opt}$,
 \bens
 \label{eq::Rlower}
 \ip{R, Z^{*} - Z}
 & \ge &  p \gamma w_{\min}^2 \shnorm{Z- Z^{*}}_1,   \text{ where } \;\; Z^{*} =  u_2 u_2^T.
 \eens
\end{lemma}

Theorem~\ref{thm::reading} is instrumental in proving Theorem~\ref{thm::SDPmain} 
in conjunction with Lemmas~\ref{lemma::GVGD} and~\ref{lemma::onenorm}.
Importantly, computing the operator and cut norms for $B-R$ is also a key 
technical step unifying Theorems~\ref{thm::SDPmain}
and~\ref{thm::SVD}, as  we demonstrate in Section~\ref{sec::SVDthm}.
We prove Theorem~\ref{thm::reading} in Section~\ref{sec::proofJR}.
We prove Lemma~\ref{lemma::onenorm} in Section~\ref{sec::proofofonenorm}. 

\begin{theorem}[\bf{\textnormal{$R$} is the leading term}]
  \label{thm::reading}
  Suppose the conditions in Theorem~\ref{thm::SDPmain}
  (resp. Theorem~\ref{thm::exprate}) hold.
Then with probability at least $1-2\exp(-cn)$, for the oracle
$B$ in \eqref{eq::defineBintro} and the reference $R$ in~\eqref{eq::Rtilt},
we have $\twonorm{B - R} \le \xi n p \gamma$ and
$\infonenorm{B - R}  \le   \xi n^2 p \gamma$.
\end{theorem}

\begin{proofof}{Theorem~\ref{thm::SDPmain}}
Under the conditions of Lemmas~\ref{lemma::GVGD}
and~\ref{lemma::onenorm}, we have
\bens 
\shnorm{\hat{Z} - Z^{*}}_1/n^2 
& \le&
\frac{\ip{R, Z^{*} - \hat{Z}}}{n^2p \gamma w_{\min}^2 }  \le 
\frac{2 K_G \norm{B -R}_{\infty \to 1}}{n^2  p \gamma w_{\min}^2 }
 \le  \frac{2 K_G \xi}{w_{\min}^2 } =: \delta,
\eens 
where by Theorem~\ref{thm::reading}, $\infonenorm{B - R} \le \xi n^2 p \gamma$.
Finally, the bound on $\shnorm{Z^{*}- \hat{Z}}^2_{F}$ follows from the
observation that
$\shnorm{Z^{*}- \hat{Z}}^2_{F} \le \shnorm{Z^{*}- \hat{Z}}_{\max}
\shnorm{Z^{*}- \hat{Z}}_1$.
Since all entries of $\hat{Z}$ and $Z^{*}$ belong to $[-1, 1]$, we have $\shnorm{Z^{*}- \hat{Z}}_{\max} \le 2$, yielding $\shnorm{Z^{*}- \hat{Z}}^2_{F} \le 2 \delta n^2$.
\end{proofof}

\subsection{Spectral clustering}
\label{sec::SVDthm}
Before exploring these results in
Theorems~\ref{thm::SDPmain},~\ref{thm::exprate}, and~\ref{thm::expratenew} further, we briefly discuss the
  spectral clustering result in Corollary~\ref{coro::SVD} of
Theorem~\ref{thm::SVD} and compare it with
Corollary~\ref{coro::misexp}.
In Theorem~\ref{thm::SVD}, we prove convergence results that bound the
angle and the $\ell_2$ distance between the leading eigenvector of $R$
and that of $B$ (resp., $YY^T$).
Theorem~\ref{thm::SVD} is new to the
best of our knowledge.

\begin{theorem}[\bf Spectral clustering]
  \label{thm::SVD}
Denote by $v_1$ the leading unit-norm eigenvector of $YY^T$,
which also coincides with that of $A$ \eqref{eq::defineAintro} and
$B$~\eqref{eq::defineBintro}.
Let $\bar{v}_1$ be the leading unit-norm eigenvector of $R$ as
in~\eqref{eq::Rtilt}. Then $\ip{\bar{v}_1, \vecone_n}=0$, where
\ben
\label{eq::Rleadingv1}
\bar{v}_1 &=& [w_2 \vecone_{n_1}, -w_1 \vecone_{n_2}]/\sqrt{w_2 w_1
  n} =  [\sqrt{w_2/w_1} \vecone_{n_1}, -\sqrt{w_1/w_2} \vecone_{n_2}]/{\sqrt{n} }.
\een
Let $\theta_1 = \angle({v}_1, \bar{v}_1)$ denote
the angle between the two vectors $v_1$ and $\bar{v}_1$.
Then under the conditions in Theorem~\ref{thm::reading},
we have with probability at least $1-2 \exp(-c n)$,  
\ben
\label{eq::angSVD}
&& \sin(\theta_1) :=  \sin(\angle({v}_1, \bar{v}_1))  \le  \frac{2  \twonorm{B - R}}{w_1 w_2 n p \gamma}  \le \frac{2 \xi}{w_1 w_2}, \\
\label{eq::normSVD}
&& \min_{\alpha=\pm 1}  \twonorm{\alpha v_1 -  \bar{v}_1}^2
  \leq \left(\frac{2^{3/2} \twonorm{B - R}}{w_1 w_2 n p \gamma}
  \right)^2 \le  \delta', \; \text{ where}  \; \delta' = \frac{8  \xi^2}{(w_1^2
    w_2^2)}.
  \een
\end{theorem}

Moreover, Corollary~\ref{coro::SVD}
shows that the signs of the coefficients of the leading eigenvector
$v_1$ of $YY^T$ correctly estimate the partition of the vertices, up
to $O(n/s^2)$ misclassified vertices in view of~\eqref{eq::trend}.
As such, the misclassification error is bounded to be inversely 
proportional to the SNR parameter $s^2$ in Theorem~\ref{thm::SVD} and 
Corollary~\ref{coro::SVD}.
This should be compared with (SDP1) and BalancedSDP, 
for which we have up to $O(n \exp(-c_0 s^2 w_{\min}^4))$ misclassified 
vertices as shown in Theorems~\ref{thm::exprate},~\ref{thm::expratenew},
and  Corollary~\ref{coro::misexp}. See 
Section~\ref{sec::experiments} for numerical examples, comparing SDP1 and SVD.

\begin{corollary}
 \label{coro::SVD}
 Denote by $v_1$ the leading unit-norm eigenvector of $YY^T$.  
Under the conditions stated in Theorem~\ref{thm::SDPmain}
or~\ref{thm::exprate}, it holds with probability at least $1-2 \exp(-c
n)$ that the signs of the coefficients of $v_1$ correctly
estimate the partition of the vertices into two clusters, 
up to $O(\xi^2 n)$ misclassified vertices, where we set $\xi^2 \asymp 1/s^2$ as 
in~\eqref{eq::trend} for $\xi = \Omega(1/n)$ as in \eqref{eq::Varprofilepreview}. 
\end{corollary}

\begin{corollary}[\bf Exponential decay]
  \label{coro::misexp}
    Let $\hat{x}$ denote the eigenvector of $\hat{Z}$
  corresponding to the largest eigenvalue, with $\twonorm{\hat{x}}
  = \sqrt{n}$.
  Denote by $\theta_{\SDP} = \angle(\hat{x}, u_2)$ the angle  between $\hat{x}$ and $u_2$.
  In the settings of Theorems~\ref{thm::expratenew}  and~\ref{thm::exprate}, we have  with probability $\ge 1-2 \exp(-c n) - 2/n^2$,
\ben
 \label{eq::eigenconv}
\sin(\theta_{\SDP}) & \le & 2 \twonorm{\hat{Z} - u_2 u_2^T}/{n} \le  \exp(- c_1 s^2 w_{\min}^4) \; \;  \text{ and } \\
\nonumber
\min_{\alpha = \pm 1}  \twonorm{(\alpha \hat{x} - u_2)/\sqrt{n}}
  & \le &
  {2^{3/2} \twonorm{\hat{Z} - u_2 u_2^T}}/{n}  \le 4 \exp(-c_0 s^2 w_{\min}^4/2).
 \een 
\end{corollary}

Under regularity assumptions on the geometry and spectrum of the centers and
their Gram matrix, \cite{AFW22} obtained exponentially decaying error
bounds for a variant of spectral clustering on the hollowed Gram matrix
$\mathcal{H}(XX^T)$, where $\mathcal{H}(\cdot)$ denotes the operator that zeros out the diagonal
entries. Concurrent with the present work, \cite{ZZ24} also
established such error bounds for spectral clustering based on $XX^T$ in the
large-$n$, small-$p$ regime; we refer to their work for a comprehensive
discussion of related spectral clustering algorithms.  
We prove Theorem~\ref{thm::SVD} in Section~\ref{sec::proofofSVD}.
Corollaries~\ref{coro::misclass},~\ref{coro::SVD}, and~\ref{coro::misexp} follow
from the Davis-Kahan Perturbation Theorem combined with Theorems~\ref{thm::SDPmain},
~\ref{thm::SVD}, and~\ref{thm::exprate}, respectively (see
Section~\ref{sec::proofofmisclass} for details).  The proof of
Corollary~\ref{coro::SVD},
mirroring that of Corollary~\ref{coro::misclass}, is omitted for brevity.

\subsection{Proof of Theorem~\ref{thm::reading} and debiasing}
\label{sec::proofJR}
Our global and local analyses also show the surprising 
result that Theorems~\ref{thm::SDPmain},~\ref{thm::exprate}, and~\ref{thm::SVD}
do not depend on the clusters being balanced,  nor do they require identical 
variance or covariance profiles, provided that Assumption~(A2) holds.
In this section, we further illuminate the constructions of $A$ and $B$ and discuss bias.
Theorem~\ref{thm::reading} follows immediately from
Lemmas~\ref{lemma::TLbounds} and~\ref{lemma::EBRtilt}, and
Theorem~\ref{thm::YYaniso}.
We prove Lemmas~\ref{lemma::TLbounds} and~\ref{lemma::EBRtilt} in Section~\ref{sec::proofofEBR}.
\begin{lemma}
  \label{lemma::TLbounds}
  Let $\tau$ be as in~ \eqref{eq::defineBintro}, $\lambda$ as in~\eqref{eq::defineAintro},  and $Y$ as in \eqref{eq::defineY}. 
  Then,
\ben
 \label{eq::lambdabound}
(n-1) \abs{\lambda - \E \lambda}
& = &  \abs{\tau - \E \tau}  \le \twonorm{YY^T - \E Y Y^T}.
\een
\end{lemma}

\begin{proofof}{Theorem~\ref{thm::reading}}
  We will only show $\twonorm{B- \E B}$ while noting that
  $\infonenorm{B- \E B} \le n \twonorm{B- \E B}$ by definition.
  By Lemma~\ref{lemma::TLbounds}, 
\bens
\label{eq::2Psi}
\twonorm{B- \E B}
&  \le & \twonorm{ Y Y^T  - \E Y Y^T} +
\abs{\lambda - \E   \lambda} \twonorm{E_n-I_n}
\le 2\twonorm{ Y Y^T  - \E Y Y^T},
\eens
where we use the fact that 
$\twonorm{E_n-I_n} \le \norm{E_n-I_n}_{\infty} = (n-1)$, 
and
\bens 
(n-1) \abs{\lambda -\E \lambda} = \abs{\tau -\E \tau}
=\abs{\tr(YY^T - \E YY^T)}/n \le \twonorm{Y Y^T  - \E Y Y^T}. 
\eens
Moreover, the concentration of measure bounds on
$\twonorm{YY^T - \E  YY^T}$ imply that, up to a constant factor,  the
same bounds also hold for $\twonorm{B - \E B}$ in view of  Lemma~\ref{lemma::TLbounds}, since
\ben
A - \E A  = 
  B -\E B 
\label{eq::Bdev}
& := &
Y Y^T - \E Y Y^T  - (\lambda- \E \lambda)(E_n - I_n), \;
\text{ and} \\
\label{eq::Bdev2}
\text{moreover} \; \twonorm{B - R} & = & \twonorm{B- \E B + \E B -R} 
\le  \twonorm{B- \E B} + \twonorm{\E B -R}.
\een
Theorem~\ref{thm::reading} holds by Lemma~\ref{lemma::EBRtilt}, 
in view of~\eqref{eq::Bdev},~\eqref{eq::Bdev2},  and
Theorem~\ref{thm::YYaniso}.
\end{proofof}

\noindent\textbf{Bias and variance tradeoff.}
The upper bound on $\norm{B - R}$ crucially depends on the SNR, $n, p,
\gamma$, and (A2) as shown in 
Theorem~\ref{thm::reading}.
Specifically, the tolerance level on the RHS of \eqref{eq::Varprofilepreview}
is chosen to be of the same order as the upper bound \eqref{eq::definemuYY} we 
obtain on $\twonorm{YY^T -  \E YY^T}$ in Theorem~\ref{thm::YYaniso}.
Roughly speaking, we impose $(A2')$~\eqref{eq::bias} to ensure \eqref{eq::normequate}:
\ben
\label{eq::normequate}
&& \text{Bias and variance tradeoff:} \; \twonorm{\E B - R} \asymp \twonorm{B - \E B} \le 2 \norm{YY^T - \E 
  YY^T}, \\
\label{eq::bias}
&& (A2') \; \abs{V_1 - V_2} \le  w =
O\left(n \Delta  \vee \sqrt{n p} \right) \quad \text{ in view
  of~\eqref{eq::definemuYY}}, \; \text{  where} \; \Delta = \sqrt{p \gamma}.
  \een
Notice that $\sqrt{n p} < n \sqrt{p \gamma}$ when $n > 1/\gamma$.
In particular, under {\bf (A2)}, we do not need a separate estimator 
for the trace of $\Sigma_j = \cov(\Z_j) = \E \Z_j \Z_j^T, j \in 
[n]$; cf.~\eqref{eq::gamma} and Lemma~\ref{lemma::twogroup}. 
Clearly, holding all other parameters fixed, a larger separation 
$\Delta$, or a larger $D:= np$, or a larger SNR $s^2$ will make it 
easier to satisfy {\bf (A2)}.
For ease of discussion, we assume (A1),  where we set $C_0 \asymp 1$.
Now, by definition of the SNR~\eqref{eq::SNR},
the lower bound in~\eqref{eq::kilo} implies that $\abs{V_1 -  V_2} \le
\sqrt{s^2} (n \vee (1/{\gamma}))$ is a sufficient condition for {\bf (A2)} to hold, since
\bens
\abs{V_1 -  V_2} \le [n \vee (1/{\gamma})] 
\sqrt{s^2} \asymp ({n}/{\xi} ) \vee (1/{(\xi  \gamma)} ) \le \xi n p \gamma,
\eens
where $\xi^2 \asymp 1/s^2$.
Given a fixed average quality $\gamma$, the tolerance for $\abs{V_1 -
  V_2}$ depends on the sample size $n$ and the separation 
parameter $\Delta$ (and hence the total data size $D:= np$).

\noindent\textbf{Bias and the sample size.}
Since $\abs{V_1 - V_2} = O(p)$ by definition,
\eqref{eq::bias} holds trivially in the large-sample setting
where $n \gg p$ and $\Delta =\Omega(1)$.
To see this, consider (A1) again.
Suppose we generate two clusters according
to Lemma~\ref{lemma::twogroup} with diagonal matrices
$H_1$ and $H_2$,
where $\Sigma_j := H_j H^T_j =
\diag([\sigma_{1 j}^2, \ldots, \sigma_{p j}^2])$  for all $ j \in \{1, 2\}$.
Then for each row vector $Z_j$  in $\MC_i$, we have by
Lemma~\ref{lemma::twogroup},
\ben
\label{eq::VA1}
V_i  := \tr(H_i H_i^T) =  \E \ip{\Z_j, \Z_j} =\sum_{k=1}^p 
\sigma_{k i}^2 =O(p).
\een
Hence, given fixed $p$ and $\gamma$ (and $\Delta$), increasing the
sample size $n$ leads to a \emph{higher tolerance} for the variance
discrepancy $\abs{V_1 - V_2}$,
thus effectively widening the range of models that satisfy (A2).
This confirms that the restriction imposed by (A2) is indeed most
stringent in the small-sample, high-dimensional regime, and becomes
non-restrictive as $n$ grows, consistent with our earlier discussion
in Section~\ref{sec::theoryaniso}.

\subsection{Related work} 
\label{sec::related}
Results in~\cite{Zhou06} and~\cite{CHRZ07} were among the first such 
results towards understanding rigorously and intuitively why proposed MAXCUT-based algorithms and previous
methods~\citep{PattersonEtAl} work in low sample settings
when $p \gg n$
and $n p$ satisfies~\eqref{eq::kilo}.
However, such results were only known for balanced MAX CUT, 
and moreover, these were structural as no polynomial time algorithms 
were given for finding MAX CUT.
In the theoretical computer science literature,
earlier work focused on learning from mixtures of well-separated
Gaussians (component distributions), where one aims to 
classify each sample according to which component distribution it 
comes from.
In earlier works~\citep{DS00,AK01}, the separation requirement 
depends on the number of dimensions of each distribution; this has
subsequently been reduced to be independent of $p$, the dimensionality of 
the distribution for certain classes of
distributions~\citep{AM05,KSV05}.

While our aim is different from those results, where $n > p$ is almost
universal and we focus on cases $p > n$ (a.k.a. high dimensional
setting), we do have one common axis for comparison, the
$\ell_2$-distance between any two centers of the distributions as
stated in~\eqref{eq::miles}:
  \ben
  \label{eq::miles}
  \Delta^2=  p \gamma \ge C \max_{i} \twonorm{\cov(\Z_i)} \big(1/{\xi^2} \vee
     \sqrt{{p}/{(n \xi^2)}} \big) \quad \text{ for some} \quad 0< \xi <1/2,
  \een
which is essentially optimal; cf.~\eqref{eq::kilo} and~\eqref{eq::NKlower}.
The main contribution of the present work is: we use the 
proposed SDP~\eqref{eq::sdpmain} and the related spectral algorithms 
to find the partition, and prove quantitatively tighter bounds than those obtained
by~\cite{Zhou06} and~\cite{CHRZ07} by removing these logarithmic factors 
entirely. Recently,  these barriers have also been broken down by a sequence of 
work~\citep{Royer17,FC18, GV19}.
However, these recent results still need the SNR to be at the order of $s^2 =
\Omega(\log n)$ even for $n > p$, except for the work of~\cite{GV19}.
We compare with~\cite{GV19} in Section~\ref{sec::kmeansintro}.

Even earlier, a small simulation example~\citep{BCFZ09} shows that
even when the sample size $n$ is small, by increasing $p$ so that $np=
\Omega(1/\gamma^2)$, one can classify a mixture of two product
populations in $\{0, 1\}^p$ using spectral methods therein with success rate reaching an ``oracle''
curve, where success rate means the ratio between correctly classified
individuals and the sample size $n$. This behavior was intriguing.
To be clear, the ``oracle'' was computed assuming that distributions 
were known. Our concentration of measure bounds, namely, 
Theorems~\ref{thm::reading} and~\ref{thm::YYaniso}, enable us to 
bypass the extraneous lower bound on sample size $n$ in~\cite{BCFZ09}
as shown in Theorem~\ref{thm::SVD} and Corollary~\ref{coro::SVD}.
We reproduce these experiments in Section~\ref{sec::experiments}, 
where we show that a full range of tradeoffs between the sample 
size and the number of features are feasible so long as 
$s^2$~\eqref{eq::SNR} is lower bounded.
We also refer to~\cite{VW02},~\cite{AM05},~\cite{KMV10},~\cite{KK10},~\cite{RCY11},~\cite{LR15},~\cite{GV15},~\cite{JMR16},~\cite{abbe17},~\cite{BWY17},~\cite{IMPV17},
~\cite{zhou2019analysis},
~\cite{LLLS+20},~\cite{CCLN21},~\cite{LZZ21},~\cite{yurtsever2021scalable},~\cite{AFW22},~\cite{nda22},~\cite{DWYZ23}, ~\cite{ZZ24},
and references therein for more related works on SDP and clustering.

\section{Proof sketch of Theorem~\ref{thm::exprate} and~\ref{thm::expratenew}}
\label{sec::exprate}
The upper bound  on the excess risk $\ip{R, Z^{*} - \hat{Z}}$
in~\eqref{eq::upperZZR} holds for all $\hat{Z} \in  \M_{\opt}$, and
hence is rather crude. In this section, we first present Lemma
\ref{lemma::signalgrad}, through which we derive a tighter upper bound
on the excess risk $\ip{R, Z^{*} - \hat{Z}}$, so as to replace~\eqref{eq::upperZZR}.
In particular,
the upper bound on $\ip{R, Z^{*} - \hat{Z}}$ will depend on the $\ell_1$ distance 
between $\hat{Z}$ and $Z^{*}$, rather than the cut norm of the matrix 
$B-R$.
All constants such as $1/6, 2/3, \ldots$ are arbitrarily chosen.
\begin{lemma}[\bf Elementary inequality]
\label{lemma::signalgrad}
Let the reference matrix $R$ be constructed as in~\eqref{eq::Rtilt}.
Let $B$ be as in Theorem~\ref{thm::SVD}.
By optimality of $\hat{Z} \in \M_{\opt}$ as~\eqref{eq::moptintro}, we have
\ben
\nonumber
p \gamma w_{\min}^2 
\shnorm{\hat{Z} - Z^{*}}_1  & \le  & 
\ip{R, Z^{*} - \hat{Z}} \le
\ip{Y Y^T - \E YY^T, \hat{Z} - Z^{*} } + \ip{\E B - R,  \hat{Z} - Z^{*} } \\
&&
\label{eq::signal}
- \ip{ (\lambda - \E  \lambda ) (E_n - I_n),   \hat{Z} - Z^{*} } =:
F_1 + F_2 + F_3.
\een
\end{lemma}

With high probability, we obtain a uniform control for $F_1=\ip{Y Y^T -  \E YY^T,  \hat{Z} - Z^{*} }$ over all $\hat{Z} \in \M_{\opt}$ in
an $\ell_1$-{\em neighborhood} of $Z^{*}$ of radius $r_1$ in
Lemma~\ref{lemma::YYlocal}.
\begin{lemma}
  \label{lemma::YYlocal}
Suppose all conditions in Theorem~\ref{thm::YYaniso} hold.
Let $r_1 \le 2 q n$ for a positive integer $1 \le q < n$. 
Then on event $\MG_{1}$, where $\prob{\MG_{1}} \ge 1-{c}/{n^2}$, we
have for $\xi \le  w_{\min}^2/16$ and $\sigma_{\max} =\max_{i}
\twonorm{H_i}$,
\ben
\nonumber
\lefteqn{
\sup_{\hat{Z} \in \M_{\opt} \cap (Z^{*} + r_1 B_1^{n \times n})}
|\ip{Y Y^T - \E Y Y^T, \hat{Z} - Z^{*} }|}\\
\label{eq::signal2}
& &
\quad \quad\le \frac{5}{6} \xi p
  \gamma \shnorm{\hat{Z} - Z^{*}}_1 + 
C' \big( \sigma_{\max} n  \sqrt{p \gamma}
+ \sigma^2_{\max} \sqrt{n p} \big) \left\lceil \frac{r_1}{2n}\right\rceil \sqrt{\log (2e n/\lceil \frac{r_1}{2n}\rceil)}.
\een
\end{lemma}
Such an upper bound \eqref{eq::signal2} now depends on the $\ell_1$ radius $r_1 \le 2n(n-1)$.
This approach also gives rise to the notion of a {\em local analysis},
following~\cite{CCLN21} and~\cite{MT99}.
Therefore, Lemma~\ref{lemma::YYlocal} is one of the main contributions in this
paper, which in turn depends on the geometry of the constraint set
$\M_{\opt}$ and  the sharp concentration bounds on
$\shnorm{YY^T-\E YY^T}$ in Theorem~\ref{thm::YYaniso}.
Similar to the global analysis, the performance of SDP1 also crucially 
depends on controlling the bias effect: 
we now control $\ip{\E B - R, \hat{Z}-Z^{*}}$ uniformly over all $\hat{Z} \in \M_{\opt}$ in 
Lemma~\ref{lemma::optlocal}.  We emphasize  that (A2) is only needed for 
Lemma~\ref{lemma::optlocal}. 
\begin{lemma}
 \label{lemma::optlocal}
  Suppose all conditions in Theorem~\ref{thm::exprate} hold. 
  Then for all $\hat{Z} \in \M_{\opt}$,
\ben
\label{eq::event0}
\abs{\ip{\E B - R,  \hat{Z} - Z^{*} }} & =: & \abs{F_2} \le 
2 p \gamma \shnorm{\hat{Z} - Z^{*} }_1 \big(\xi  + 1/{[4(n-1)]}\big).
\een
\end{lemma}
\begin{lemma}
  \label{lemma::optlocal2}
Under the settings of Theorem~\ref{thm::YYaniso},
with probability at least $1-\exp(cn)$,
\bens
\forall \hat{Z} \in \M_{\opt}, \quad \abs{\ip{ (\lambda - \E  \lambda ) (E_n - I_n),   \hat{Z} -
      Z^{*} } } =:  \abs{F_3}  \; \le\;  \xi p \gamma  \shnorm{\hat{Z} - Z^{*}}_1 /3.
\eens
\end{lemma}
Lemmas~\ref{lemma::optlocal} and \ref{lemma::optlocal2} show that 
each term may take out only a small fraction of the signal on the LHS of \eqref{eq::signal} respectively.
We prove Lemmas~\ref{lemma::signalgrad},~\ref{lemma::YYlocal}, and~\ref{lemma::optlocal}
in Sections~\ref{sec::proofofsignal},~\ref{sec::proofofYYlocalmain}, 
and~\ref{sec::proofofoptlocal}, respectively. 
Lemma~\ref{lemma::optlocal2} follows from the Lemma~\ref{lemma::TLbounds}
and Theorem~\ref{thm::YYaniso} immediately.
Combining these results leads to an exponentially decaying error bound 
with respect to the SNR $s^2$ in Theorem~\ref{thm::exprate}; cf.
Section~\ref{sec::proofexprate}.

\noindent{\bf Balanced partitions.}
For BalancedSDP~\eqref{eq::sdpball}, the bias term arising from local
analysis is now 0. More precisely, we show $\forall \hat{Z} \in
  \M_{\opt}, F_2:= \ip{\E B - R,  \hat{Z} - Z^{*} } = 0$, even when
  $V_1 \not= V_2$.
  Our
  design and local analysis on semidefinite relaxations suggest that
  the original SDP1 may already have a substantially reduced bias
  $F_2$ for balanced partitions; cf. Conjecture~\ref{conj::balpar} and 
  discussions in Section~\ref{sec::kmeansintro}.
  See Section~\ref{sec::biasredo}  for details.

\section{Reduction: proof outline for Theorem~\ref{thm::YYaniso}}
\label{sec::reduction}
In this section, we present a unified framework for bounding 
$\twonorm{Y Y^T - \E Y Y^T}$.
The decomposition and projection ideas in this section are also crucial in the local
analyses for Theorems~\ref{thm::expratenew} and~\ref{thm::exprate};
cf. proof of Lemma~\ref{lemma::YYlocal} in Section~\ref{sec::proofofYYlocalmain}.
Let $c, c', C, C_2, C_3, C_4, \ldots$ be absolute constants.
First, by linearity of expectation,
we have for $Y$ as in~\eqref{eq::defineY},
with row vectors $Y_1, \ldots, Y_n \in
\R^{p}$,
\ben
\label{eq::EYpre}
\quad \E Y_i  :=
\E X_i - \E \hat \mu_n  :=
\left\{
  \begin{array}{rl} w_2 (\mu^{(1)} - \mu^{(2)})  & \text{ if }  \; i \in \MC_1; \\
    w_1 (\mu^{(2)} - \mu^{(1)}) &  \text{ if }  \; i \in \MC_2,
  \end{array}\right. \; \text{ where}  \;\hat{\mu}_n = \inv{n} \sum_{i=1}^n X_i.
\een
Next, we decompose $YY^T - \E YY^T$ in Lemma~\ref{lemma::YYdec}.
Controlling the first component, namely, $\hat{\Sigma}_Y - \Sigma_Y$
in \eqref{eq::projection} amounts to the problem of covariance 
estimation, where we assume (implicitly) that $\E X$ is given.
In particular, controlling the norm of $\hat\Sigma_Y -\Sigma_Y$~\eqref{eq::YYP1proj}
is reduced to bounding that for $\Z \Z^T - \E \Z \Z^T$ as shown in 
Lemma~\ref{lemma::YYcovcorr}.
\begin{lemma}[\bf Decomposition]
  \label{lemma::YYdec}
  Let $Y = (I-P_1)X$ be as in \eqref{eq::defineY}.
  Suppose $\Z = X- \E X$.  Let $\hat{\Sigma}_Y = (Y - \E Y)(Y - \E
  Y)^T$ and $\Sigma_Y :=\E YY^T - \E(Y) \E(Y)^T$.
  Then 
\ben 
\label{eq::projection}
  YY^T - \E Y Y^T & = &  \hat{\Sigma}_Y -
  \Sigma_Y + \E(Y)(Y-\E(Y))^T + (Y-\E(Y))(\E(Y))^T, \\
  \text{ where} \;
  \label{eq::YYP1proj}
  \hat{\Sigma}_Y - \Sigma_Y
  & = & (I-P_1) (\Z \Z^T - \E \Z\Z^T)  (I-P_1), \; \text{for } \; P_1 = E_n/n.
\een
\end{lemma}

\begin{lemma}
  \label{lemma::YYcovcorr}
Let $\sigma_{\max}^2 := \max_{j} \shnorm{\cov(\Z_j)}_2  = \max_{i}
\shnorm{H_i}_2^2$ and $C_0$ be as in  Theorem~\ref{thm::YYaniso}.
In the settings of Theorem~\ref{thm::YYaniso} (resp.~\ref{thm::ZHW})
and Lemma~\ref{lemma::YYdec}, we have with probability at least $1 -
2\exp(-c_6 n)$,
\ben 
\label{eq::ZZHop}
\twonorm{\hat\Sigma_Y - \E \hat\Sigma_Y} \le 
\label{eq::sigmax}
\twonorm{\Z \Z^T - \E  \Z \Z^T} \le  C_2 C_0^2  \sigma_{\max}^2 (\sqrt{n
  p} \vee n),
\een
where $K \le C_0 = \max_{j} \norm{W_j}_{\psi_2} \le C K$ for an absolute constant $C$.
\end{lemma}

Next, we state a reduction  principle: 
to control $\norm{\E(Y)(Y-\E(Y))^T + (Y-\E(Y))(\E(Y))^T}$,
we need to bound the projection of each mean-zero random vector $\Z_j,
\forall j \in [n]$, along the direction of $\mu
={(\mu^{(1)}-\mu^{(2)})}/{\sqrt{p \gamma}} \in \Sp^{p-1}$.
In other words, Lemma~\ref{lemma::tiltproject} shows that 
a particular direction for which we compute the 
one-dimensional marginals for $\Z_j, j \in [n]$ is the direction between the two centers, namely, $\mu^{(1)}$ and $\mu^{(2)}$ in $\R^{p}$.
Moreover, the bound in~\eqref{eq::pairwise2} is deterministic and independent of 
the covariance structure of $\Z$.
Combining Lemma~\ref{lemma::tiltproject} with the sub-gaussian tail
bounds, we have Lemma~\ref{lemma::projWH}.
\begin{lemma}[\bf Reduction: a deterministic comparison lemma]
\label{lemma::tiltproject}
Let $\Z_j, j \in [n]$ be row vectors of $X - \E X$ and $\hat{\mu}_n= \inv{n} \sum_{i=1}^n X_i$.
For $x_i \in \{-1,1\}$,
\ben
\label{eq::pairwise2}
\sum_{i=1}^n x_i \ip{Y_i -\E Y_i, \mu^{(1)} -\mu^{(2)}}
& \le &
{2(n-1)}/{n} \sum_{i=1}^n \abs{\ip{\Z_i, \mu^{(1)}  -\mu^{(2)}} }. 
\een
Then we have for $M_Y :=\E(Y)(Y-\E(Y))^T + (Y-\E(Y))(\E(Y))^T$,
\bens 
 \twonorm{M_Y}
& \le & 4 \sqrt{n}  \sqrt{w_1 w_2} \sup_{q \in S^{n-1}} |\sum_{i} q_i \ip{\Z_i, \mu^{(1)} -\mu^{(2)}}|. 
\eens 
\end{lemma}

\begin{lemma}[\bf Projection: probabilistic view]
\label{lemma::projWH}
Let $M_Y$ be as in  Lemma~\ref{lemma::tiltproject}.
Let $\mu$ be as defined in~\eqref{eq::definemu}.
Suppose all conditions in Theorem~\ref{thm::YYaniso} hold.
Then with probability at least $1 -2 \exp(-c' n)$,
$\twonorm{M_Y}  \le 
2 C_3  C_0 (\max_{i} \twonorm{H_i^T \mu}) n \sqrt{p \gamma}
\le \xi n p \gamma/12$. 
\end{lemma}

Throughout this paper, it is understood that Theorem~\ref{thm::ZHW}, 
Lemma~\ref{lemma::YYdec},~\eqref{eq::ZZHop}, and~\eqref{eq::pairwise2}
hold under the covariance model as considered in 
Definition~\ref{def::WH}
regardless of the weights or the number of 
mixture components in data matrix $X$.
Such generalization is useful for the $k$-component 
mixture problems~\eqref{eq::model} for any $k \ge 2$; 
cf. Section~\ref{sec::kmeansintro}.
Theorem~\ref{thm::YYaniso} (for $k=2$) follows 
from
Lemmas~\ref{lemma::YYdec} and~\ref{lemma::projWH}, and
\eqref{eq::ZZHop} immediately,  and the probability statements hold upon adjusting the constants.

\noindent{\bf Remarks on the diagonal covariance assumption.}
In Theorem~\ref{thm::SDPmain}, each noise 
vector $\Z_j, \forall j \in [n]$ has independent, mean-zero,
sub-gaussian coordinates with uniformly bounded $\psi_2$ norms.
Suppose now we generate two clusters according
to Lemma~\ref{lemma::twogroup} with diagonal matrices $H_1$ and $H_2$, respectively, where
$H_j H^T_j =  \diag([\sigma_{1 j}^2, \ldots, \sigma_{p j}^2])$ 
for all $ j \in \{1, 2\}$.
Here, we use $\diag(x)$ to denote the diagonal matrix whose 
main diagonal entries are the entries of $x=[x_1, \ldots, x_p]$.
 For each row vector in $\MC_i, i =1, 2$, it follows from
 Lemma~\ref{lemma::twogroup} that $\cov(\Z_j) =  H_i H_i^T$ for nodes
 $j \in  \mathcal{C}_i$ and $V_i = \tr(H_i H_i^T) = \sum_{k=1}^p \sigma_{k i}^2$.
Let  $\sigma_{\max}  :=  \max_{i} \twonorm{H_i} =
(\max_{j, k} \E (z_{j k}^2) )^{1/2}$. 
Then clearly, $\max_{j} \twonorm{\cov(\Z_j)}=\sigma_{\max}^2$.
Given the independence of the coordinates of $\Z_j$, we have
\eqref{eq::Zpsi0} holds with  $C_0 \le C K$, since
\bens 
\norm{\Z_j}_{\psi_2} & := & \sup_{h \in \Sp^{p-1}}\norm{\ip{\Z_j, 
    h}}_{\psi_2} \le C \max_{k \le p} \norm{z_{j k}}_{\psi_2}
\le  C K \sigma_{\max}, \quad \text{where} \\
\max_{k \le p} \norm{z_{j k}}_{\psi_2}
& \le &   \sigma_{\max} (\max_{j,k}\norm{w_{j k}}_{\psi_2}) 
= \sigma_{\max} K, \quad \text{ for } \quad  K := \max_{j,k} \norm{w_{jk}}_{\psi_2}.
\eens
\noindent{\bf Remarks on covariance estimation.}
More generally, we consider the  general data generative process in 
Definition~\ref{def::WH}, and denote the (sample-by-sample) covariance by 
\ben 
\label{eq::gamma}
\Gamma:= \E \Z \Z^T 
=  \diag( [\tr(\Sigma_1), \ldots, \tr(\Sigma_n)]), \; \text{where} \; 
\Sigma_j = \cov(\Z_j) = H_j H_j^T; 
\een
Thus, we estimate a diagonal matrix with $p$ dependent features.
Essentially, \eqref{eq::ZZHop} matches the optimal bounds for
covariance estimation, where the mean-zero random matrix consists of 
independent columns $\Z^j, j \in [p]$, which are isotropic, 
sub-gaussian random vectors in $\R^n$, or columns which can be 
transformed to be isotropic through a common covariance matrix;
see, for example, Theorems 4.6.1 and 4.7.1~\citep{Vers18}.
The differences between \eqref{eq::ZZHop} and such known results are:
(a) we do not assume that columns are independent; and (b)  we do not
require anisotropic row vectors to share identical covariance
matrices. See~\cite{Zhou19,Zhou24} and references therein for extensions
of the Hanson-Wright inequalities in the complex matrix-variate data
setting.
We prove Theorem~\ref{thm::YYaniso} in Section~\ref{sec::proofYYconc},
and Lemmas~\ref{lemma::YYcovcorr}  and~\ref{lemma::projWH} in 
Sections~\ref{sec::ZZHproof} and~\ref{sec::proofprojWH}, respectively.
See also Section~\ref{sec::reductionproofs} for the proofs of 
Lemmas~\ref{lemma::YYdec} and~\ref{lemma::tiltproject}.


\section{The $k$-means criterion of a partition: a comparison}
\label{sec::kmeansintro}
We now discuss the related
$k$-means criterion and its semidefinite relaxations.
Denote by $X \in \R^{n \times p}$ the data matrix with row vectors 
$X_i$ as in~\eqref{eq::SSE}.
The $k$-means criterion of a partition $\MC = \{\MC_1, \ldots,
\MC_k\}$ of sample points $\{1, \ldots, n\}$ is based on the total sum-of-squared Euclidean distances from each point
$X_i \in \R^{p}$ to its assigned cluster centroid $\vc_j$, namely,
\ben
\label{eq::SSE}
g(X, \MC, k) & := & \sum_{j=1}^k \sum_{i \in \MC_j} \twonorm{X_i - \vc_j}^2, \quad \text{
  where} \quad \vc_j  := \sum_{\ell \in \MC_{j}} X_{\ell} /\abs{\MC_j}\in \R^{p}.
\een
Getting a global solution to~\eqref{eq::SSE} through an integer
programming formulation~\citep{PW07} is NP-hard
and it is NP-hard for $k=2$~\citep{DFK+04,ADHP09}.
The partition $\CC$ can be represented by
a block-diagonal matrix
of size $n \times n$, defined as: $\forall i, j \in [k]^2, \forall
a, b \in \MC_{i} \times \MC_{j}$, 
$B_{ab}^* = 1/{\size{\MC_j}} \; \text{ if } \; i = j$ and $B_{ab}^* = 0$  otherwise.
Let $\Psi_n$ denote the linear space of real $n$ by $n$ symmetric 
matrices. The collection of such matrices can be described by
$$\mathcal{P}_k
=\{B \in \Psi_{n}:  B \ge 0, B^2 = B, \tr(B) = k,B \vecone_n = \vecone_n\}.$$
Here $B \ge 0$ means that all elements of $B$ are nonnegative.
Hence matrices in $\cpk$ are block-diagonal, symmetric, nonnegative
projection matrices with $\vecone_n$ as an eigenvector.
Minimizing
the $k$-means objective $g(X, \MC, k)$ is equivalent to
\ben 
\label{eq::relax7}
\text{maximize} \quad \ip{X X^T, Z}  \quad \text{subject to } \; \; 
Z \in \cpk; \; \text{cf.~\cite{PW07}.}
 \een
\noindent{\bf Peng-Wei relaxations.}
\label{sec::variations}
~\cite{PW07} first replace the requirement that $Z^2 =
Z$, namely, $Z$ is a projection matrix, with $I_n \succeq Z \succeq 0$.
Let $\M_k = \{Z \in \Psi_n:  Z   \ge 0, Z \succeq 0, \tr(Z) = k, 
Z \vecone_n = \vecone_n \}$.
Now,~\cite{PW07} consider the semidefinite relaxation of the $k$-means
objective~\eqref{eq::relax7},
\ben
\label{eq::relax16}
\text{maximize}
\quad \ip{ X X^T, Z}   \quad \quad \text{subject to }  \quad  Z \in 
\M_k.
\een
Note that in $\M_k$, we have $\twonorm{Z}
\le \norm{Z}_{\infty} = 1$, since $Z_{ij} \ge 0,  \forall i, j$ and row sum $\onenorm{Z_{j, \cdot}}
=1, \forall j$. 
The main issue with the $k$-means relaxation is that the solutions
tend to put sample points into groups of the same size, and moreover,
the diagonal matrix $\Gamma$ as in~\eqref{eq::gamma}
can cause a bias, especially when $\tr(\Sigma_1), \tr(\Sigma_2), \ldots$ differ from each other.
To address this bias, building upon the original Peng-Wei SDP
relaxation~\eqref{eq::relax16}, the authors
of~\cite{BGLR+20},~\cite{Royer17}, and~\cite{GV19}
propose a preliminary estimator of $\Gamma$, denoted by $\hat{\Gamma}$, and
consider
\ben
\label{eq::relaxadjust}
\text{maximize}
\quad \ip{X X^T - \hat{\Gamma}, Z} \quad \text{ subject to }  \quad  Z
\in  \M_k \; \text{ for } \M_k \quad \text{as in~\eqref{eq::relax16}. }
\een

\noindent{\bf Variation 2.}
To speed up computation, one can drop the nonnegative 
constraint on elements of $Z$ in \eqref{eq::relax16}.
In particular,~\cite{PW07} consider 
\ben
\label{eq::relax17}
&&  \text{maximize} \quad \ip{XX^T, Z} \quad \text{ subject to }
\quad Z \vecone_n = \vecone_n, I_n \succeq Z \succeq 0, \tr(Z) = k.
\een
~\cite{PW07} show that the set of feasible
solutions to~\eqref{eq::relax17} has immediate connections to the SVD
of $YY^T$.
Recall $YY^T = (I-P_1) XX^T (I-P_1)$. Let $\lambda_1 \ge  \ldots \ge \lambda_{n-1}$ be the 
largest $(n-1)$ eigenvalues of $YY^T$ in descending order. 
When $Z$ is a feasible solution to~\eqref{eq::relax17},
$\vecone_n/\sqrt{n}$ is the unit-norm leading eigenvector of $Z$ and
define $Z_1 :=  (I-P_1) Z = (I-P_1)Z (I-P_1) \succeq 0$.
Then,~\eqref{eq::relax17} is reduced to
 \ben 
 \label{eq::relax20}
 && \text{maximize} \quad \ip{YY^T, Z_1}   \text{ subject to }  \quad
 I_n \succeq Z_1 \succeq 0, \quad \tr(Z_1) = k-1.
 \een
The optimal solution to~\eqref{eq::relax20} can be achieved if and
only if $\ip{YY^T,  Z_1} = \sum_{i=1}^{k-1} \lambda_i$~\citep{OW93}.
Then the SVD-based algorithm for solving~\eqref{eq::relax20}
and~\eqref{eq::relax17} is given as follows: \\
\noindent{\bf (a)} Use the singular value decomposition method to compute the first $k-1$
largest eigenvalues of $YY^T$, and their corresponding eigenvectors
$v_1, \ldots, v_{k-1}$; \\
\noindent{\bf (b)} Set $Z_1 =\sum_{j=1}^{k-1} v_j v_j^T$  \text{and return} $Z =\vecone_n
\vecone_n^T/n + Z_1$ \text{ as a solution to \eqref{eq::relax17}}.
Now for $k=2$, we have $Z_1 = v_1 v_1^T$ and Theorem~\ref{thm::SVD}.

In summary, the key differences between \eqref{eq::relaxadjust} and
SDP~\eqref{eq::sdpmain} are as follows:
(a) In the convex set~$\M_{\opt}$~\eqref{eq::moptintro}, we do not
enforce non-negativity for all entries (i.e., $Z_{ij} \ge 0$ for all
$i, j$), nor the trace or row-sum constraints ($Z \vecone_n =
\vecone_n$). By replacing these with the $n$ diagonal linear constraints
$\{Z_{jj} = 1, \forall j \in [n]\}$, the theoretical complexity of the
semidefinite optimization problem is substantially reduced; see, for
example,~\cite{burer2003nonlinear,monteiro2003first,yurtsever2021scalable},
and references therein.
(b) To derive concentration of measure bounds that are sufficiently tight,
we perform a crucial data-processing step by centering the data
according to their column means, following \eqref{eq::defineY}. This
allows the bias to be substantially reduced, or eliminated entirely
for balanced partitions using the BalancedSDP estimator.
(c) When we do enforce $\ip{E_n, Z} = 0$ in
BalancedSDP~\eqref{eq::sdpball}, the trace-difference assumption (A2)
is eliminated entirely in Theorem~\ref{thm::expratenew} for balanced
partitions.
(d) Consequently, beyond this reduction in the size of the constraint
set, a primary merit of (SDP1) and its variants proposed in the present 
work is that one does not need a separate estimator for $\tr(\Sigma_j)$ for balanced 
partitions, or for general partitions under (A2).

We acknowledge that the results by~\cite{GV19} pertain to general $k \ge 2$,
whereas our work focuses on $k=2$.
While it is natural to consider $k=2$ within the MAXCUT framework, 
parts of our probabilistic bounds, specifically Theorem~\ref{thm::ZHW} and 
Lemma~\ref{lemma::YYcovcorr}, are developed for the most general
$k$-means setting~\eqref{eq::model}, for any $k \ge 2$.
Moreover, our deterministic bounds in Lemmas~\ref{lemma::tiltproject}
and~\ref{lemma::YYdec} can be readily adapted and applied to the
rank-$k$ model setting, as mentioned in Section~\ref{sec::reduction}.

\subsection{The $k$-means criterion}
\label{sec::SNRcompare}
Denote by $\Sigma_j = \cov(\Z_j)$ and $\sigma^2 :=C_0^2  \max_{j}\shnorm{\Sigma_j}_2$.
Note that both~\cite{GV19} and the present work require that
$\Delta = \Omega(\sigma)$.
In particular, the SNR defined in~\cite{GV19}, denoted $s_{GV}^2$, is
\ben
\label{eq::SNRGV}
\text{\bf (GV): }  \quad s_{GV}^2  = \frac{\Delta^2}{\sigma^2}
\wedge \frac{n \Delta^4} {C_0^4 \max_{j}\fnorm{\Sigma_j}^2},
\quad \text{whereas our} \text{ \bf SNR (Z): }   \quad
s_{Z}^2 = \frac{\Delta^2}{\sigma^2} \wedge   \frac{n \Delta^4}{p \sigma^4},
\een
by \eqref{eq::SNR2}, where $\Delta^2 = p \gamma$.
Theorem~1 by~\cite{GV19} states that as long as $s_{GV}^2 = \Omega(k)$, an 
exponentially decaying misclassification error with respect to the SNR 
is achieved, a result analogous to Theorem~\ref{thm::exprate} and Corollary \ref{coro::misexp}.
Under the assumptions in Definition~\ref{def::WH}, the SNR $s_{GV}^2$ in~\eqref{eq::SNRGV} is
asymptotically equivalent to $s_{Z}^2$ in~\eqref{eq::SNR2} provided
that $\Sigma_j$ is well-conditioned in the sense that
$0 < \lambda_{\min}(\Sigma_j) \le \lambda_{\max}(\Sigma_j) < \infty$.
This follows from the trace-operator norm relationship:
\bens 
\text{ for each } \; j, \quad 
p \lambda_{\min}(\Sigma_j)^2 \le 
\sum_{k=1}^p \lambda_k(\Sigma_j)^2 = \fnorm{\Sigma_j}^2 
\le p \twonorm{\Sigma_j}^2,
\eens 
which implies that $\fnorm{\Sigma_j}^2$ scales linearly with the
dimension $p$. Hence, for isotropic designs or well-conditioned covariance matrices,
the error rate in Theorem \ref{thm::exprate} is essentially equivalent
to that obtained by \cite{GV19}.

These two definitions diverge primarily when $n \ll p$, as elaborated 
below; in the large-sample setting where $n \gg p$, both reduce to 
$s_{Z}^2 = s_{GV}^2 = {\Delta^2}/{\sigma^2}$. 
Moreover, it is always true that our $s^2_Z$ in \eqref{eq::SNR2} is upper
bounded by $s_{GV}^2$ in~\eqref{eq::SNRGV}, that is, $s_Z^2 \le s_{GV}^2$.
Let $r(\Sigma_j) = \frac{\fnorm{\Sigma_j}^2}{\norm{\Sigma_j}_2^2}$
denote the stable rank. 
In summary, the error rate $\exp(-c s^2)$ in Theorem~\ref{thm::exprate}
will be slightly slower for anisotropic designs when $n < p$ but
$r(\Sigma_j) =O(p), \forall j$. On the other hand, when
$\max_j r(\Sigma_j) \ll p$, that is, when $\fnorm{\Sigma_j}^2$ is significantly smaller
than $p \norm{\Sigma_j}_2^2, \forall j$,
the ratio ${s_Z^2}/{s_{GV}^2}$ becomes vanishingly small.
We emphasize that such regimes fall outside the scope of our analysis in the present work.
More accurately, we consider the following three cases for
$\Delta^2 = p \gamma \ge \sigma^2 := C_0^2   \max_{j}\shnorm{\Sigma_j}_2$,
  \bit  
\item  
  {\bf Case 1:} Suppose $n \gamma > \sigma^2$.
  Then $s_{Z}^2 = s_{GV}^2 = {\Delta^2}/{\sigma^2}$, since
\bens 
n \Delta^2  = np \gamma >
p\sigma^2, \quad \text{ and consequently,} \quad
    \frac{n \Delta^4} {C_0^4   \max_{j}\fnorm{\Sigma_j}^2} \ge \frac{n 
    \Delta^4}{p \sigma^4} > \frac{\Delta^2}{\sigma^2}. 
  \eens  
\item {\bf Case 2:}
  Suppose $p \gamma \ge \sigma^2 \ge n \gamma$, which implies $p \sigma^2 \ge np \gamma$. Here, $np\gamma$ satisfies:
\bens  
p \sigma^2 \ge n p \gamma \ge \frac{C_0^4 \max_{j}
  \fnorm{\Sigma_j}^2}{C_0^2 \max_{j} \shnorm{\Sigma_j}_2} \quad  \text{ so that} \quad
\frac{n \Delta^4} {C_0^4 \max_{j}\fnorm{\Sigma_j}^2} \ge  
\frac{\Delta^2}{\sigma^2} \ge \frac{n \Delta^4}{p \sigma^4}.  
\eens  
Then $s^2_{GV} = {\Delta^2}/{\sigma^2}$ while $s^2_Z
= {n \Delta^4}/{(p \sigma^4)}$. Hence,
${s_Z^2}/{s_{GV}^2} = {n (p \gamma)}/{(p \sigma^2)} = {n
  \gamma}/{\sigma^2}.$
\item {\bf Case 3:}
 Suppose $n p \gamma  \le  \frac{C_0^4 \max_{j}
   \fnorm{\Sigma_j}^2}{C_0^2 \max_{j}  \shnorm{\Sigma_j}_2}$.
Then 
  \bens 
  s_{Z}^2 \asymp \frac{n \Delta^4}{p C_0^4 \max_{j}  \twonorm{\Sigma_j}^2}  =
  \frac{n \Delta^4}{p \sigma^4} \le 
  \frac{n \Delta^4} {C_0^4   \max_{j}\fnorm{\Sigma_j}^2}
\le  \frac{\Delta^2}{C_0^2 \max_{j} \shnorm{\Sigma_j}_2} := \frac{\Delta^2}{\sigma^2}. 
\eens
Thus, we have ${s_Z^2}/{s_{GV}^2} \approx \frac{\max_j \fnorm{\Sigma_j}^2}{p
  \max_j \shnorm{\Sigma_j}_2^2} \le 1$ since
$\forall j, \; \fnorm{\Sigma_j}^2 \le p \shnorm{\Sigma_j}^2_2 \le p (\max_{j} \shnorm{\Sigma_j}^2_2)$.
\eit
\noindent{\bf Discussion.}
For {\bf Case 2},
the further $n \gamma$ is below 
$\sigma^2$, the larger the performance gap becomes. When $n \gamma 
\approx \sigma^2$, the two estimators perform nearly identically.
When  $n \gamma \ll \sigma^2$, the gap widens. 
For {\bf Case 3}, we have
$n \le \frac{\max_{j} \fnorm{\Sigma_j}^2}{\max_{j}
  \norm{\Sigma_j}^2_2} \le p$.
The ratio satisfies:
\bens
    \frac{s_Z^2}{s_{GV}^2} = \frac{\max_j \fnorm{\Sigma_j}^2}{p \max_j \norm{\Sigma_j}_2^2} \le \max_j \frac{r(\Sigma_j)}{p}.
    \eens
In isotropic cases, $r(\Sigma_j) \approx p, \forall j$, making the ratio 
$\approx 1$. When $\max_j r(\Sigma_j) \ll p$, this ratio reaches its
worst-case bound in this small-$n$, large-$p$ regime.

\section{Experiments}
\label{sec::experiments}
In this section, we use simulation to illustrate the effectiveness and
convergence properties of the three estimators. We implement (i) 
{\bf SDP1} as described in~\eqref{eq::hatZintro},  (ii)
{\bf BalancedSDP} as described in~\eqref{eq::sdpball}, 
with the extra constraint $\tr(E_nZ) =0$,
and (iii) {\bf SVD}, the Peng-Wei spectral method
following~\cite{PW07}. 
For both SDP1 and BalancedSDP we classify according to the signs of
$\hat{x}$ as  prescribed by Corollary~\ref{coro::misclass}.
We list the SVD procedure in Table~\ref{tab:svd}.
The experiment set up is similar to the one used in ~\cite{BCFZ09}.  The
data matrix $X \in \{0, 1\}^{n \times p}$ is generated as a mixture of two
populations, and consists of independent Bernoulli random variables, where
the mean parameters $ \E (X_{ij}) := q_{\psi(i)}^j$ for all $i \in [n]$
and $j \in [p]$, where $\psi(i) \in \{1, 2\}$ assigns nodes $i$ to  a
group $\MC_1$ or $\MC_2$ for each $i \in [n]$. Let $\size{\MC_1} = w_1 n$
and $\size{\MC_2} = w_2 n$.  We conduct experiments for both balanced
($w_1 = w_2$) and unbalanced cases.  The entrywise expected values are
chosen according to Table~\ref{tab:parameters}(left) for the two populations,
where $f_1, f_2, f_3$ and $m_1, m_2$ are configurable.  We experiment with
two configurations of population variance profiles. (i) Config1:
$f_1=50\%, f_2=50\%, f_3=0$, where $V_1 = V_2$. Here $\gamma = 0.0016$ and
$1/\gamma = 625$.  (ii) Config2: $f_1=50\%, f_2=30\%, f_3=20\%$, and
$m_1=0.3, m_2=0.1$, where $V_1 \ne V_2$, and $\gamma = 0.00928$ and
$1/\gamma =107$.  See Table~\ref{tab:parameters}(right) for the
actual $p\gamma$ values.
\begin{table}
\begin{tabular}{p{6in}}
\hline
{\bf SVD: Spectral method for $k$-means clustering~\citep{PW07}:} \\ \hline
{\bf Input:} Centered data matrix $Y \in \R^{n \times p}$, $k=2$\\
{\bf Output:} A group assignment vector $P$ \\
  
\noindent{\bf Step 1.}
Use SVD to obtain the leading eigenvector $v_1$ of  $Y Y^T$ and let $v := v_1$;\\
\noindent{\bf Step 2.}
Let $S$ be the vector of sorted values of $v$ in descending
order. For each index $j$ in $[n]$, compute
the two means $\vc_{1}$, $\vc_2$, one for each of the two groups,
namely, $\MC_1 = S_L := \{S_1, \ldots, S_j\}$ and $\MC_2 = S_R :=
\{S_{j+1}, \ldots, S_n\}$ to the left (inclusive) and the right of this index;\\
 \noindent{\bf Step 3.}
Compute the total sum-of-squared Euclidean distances from each point within
a particular group to the respective mean, according to~\eqref{eq::SSE};
Let $t$ be the index that gives the minimum total distance, and its corresponding value
  be $S_t$; \\
\noindent{\bf Step 4.}
  Set $P_i = 1$ if $v_i \ge S_t$, and $P_i = -1$ if $v_i < S_t$.
 \\ \hline
\end{tabular}
  \caption{SVD: Spectral method for $k$-means clustering~\citep{PW07}}
  \label{tab:svd}
\end{table}

\begin{table}
\begin{tabular}{cc}
\begin{tabular}{cccc}
 percentage of features  & $f_1$ & $f_2$ & $f_3$ \\ \hline
  mean parameters (pop. 1)   &  $\frac{1 + \alpha}{2} + \frac{\e}{2}$ & $\frac{1 - \alpha}{2} + \frac{\e}{2}$ & $m_1$ \\  \hline
  mean parameters (pop. 2)   &  $\frac{1 - \alpha}{2} + \frac{\e}{2}$ & $\frac{1 + \alpha}{2} + \frac{\e}{2}$ &  $m_2$\\ \hline
\end{tabular}
  &
  \hspace{15pt}
\begin{tabular}{rrrr}
  $p$  &    $V_1$   &  $V_2$  & $\Delta^2=p\gamma$\\  \hline
  100  & 24.17 &  21.77 & 0.928 \\  \hline
  200  & 48.33 &  43.54 & 1.856 \\  \hline
  400  & 96.66 &  87.07 & 3.712 \\  \hline
  1000& 241.66 & 217.69 & 9.280 \\  \hline
\end{tabular}
    \end{tabular}
  \caption{Left: Entrywise expected values. $\alpha=0.04$, $\e = 0.1 \alpha$. 
    Right: Population profiles in Config2, where $V_1 \neq V_2$}
  \vspace{-0.5cm}
  \label{tab:parameters}
\end{table}

\begin{figure*}[!tb]
  \centering
  \vspace{-0.5cm}
\begin{tabular}{cc}
  \hspace*{-0.6cm}
    \vspace{-0.5cm}
\includegraphics[width=2.8in]{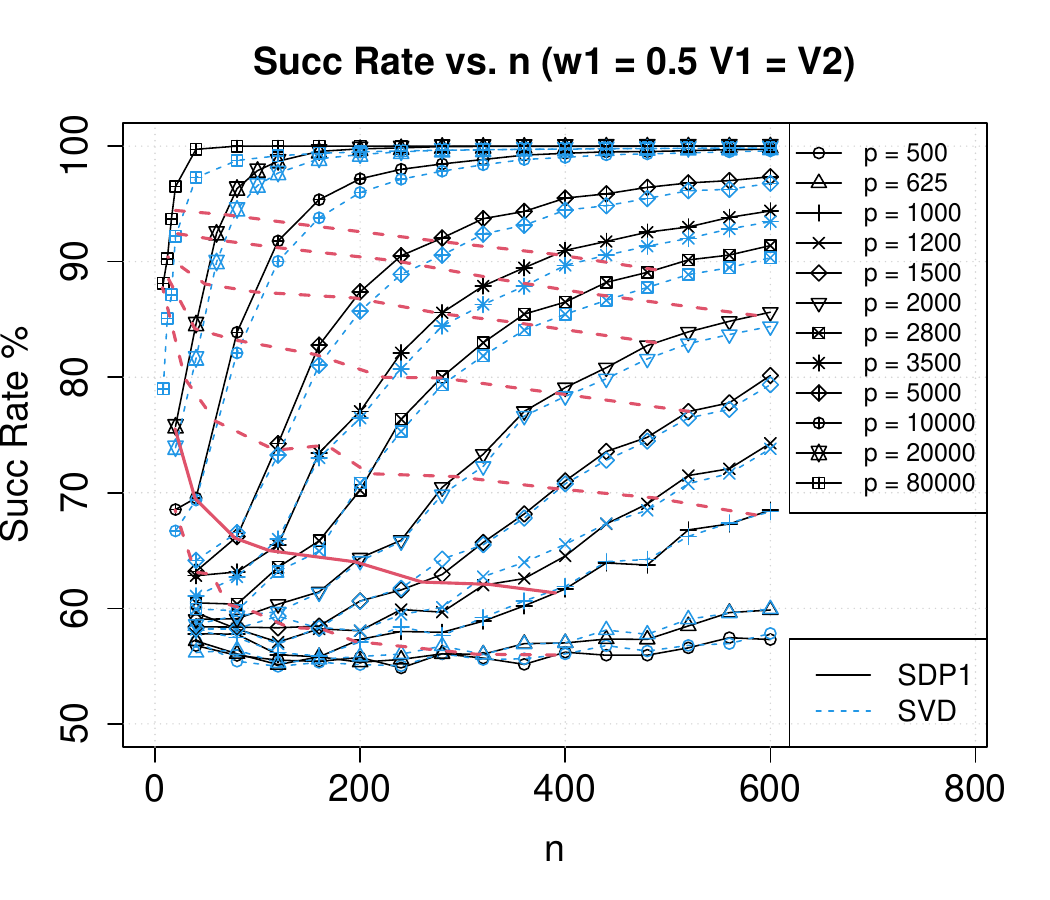} & 
\includegraphics[width=2.8in]{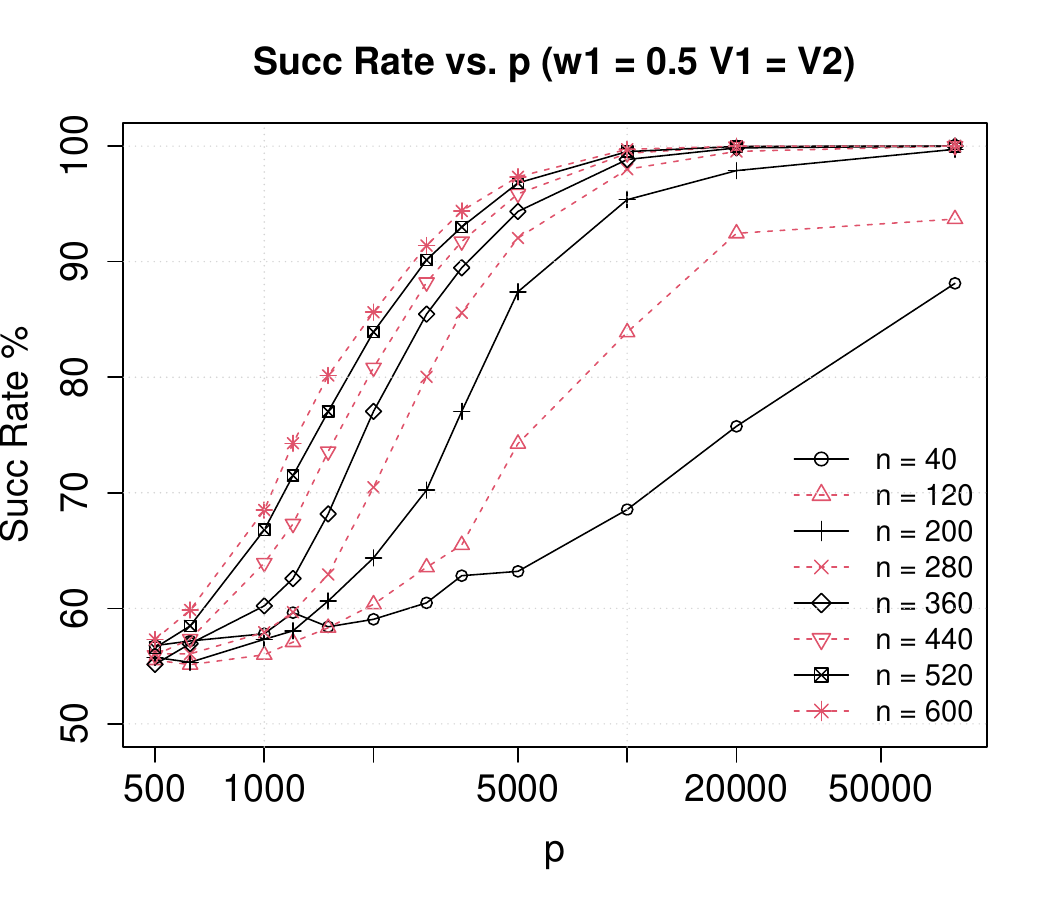} \\
  \hspace*{-0.6cm}
\vspace{-0.5cm}
\includegraphics[width=2.8in]{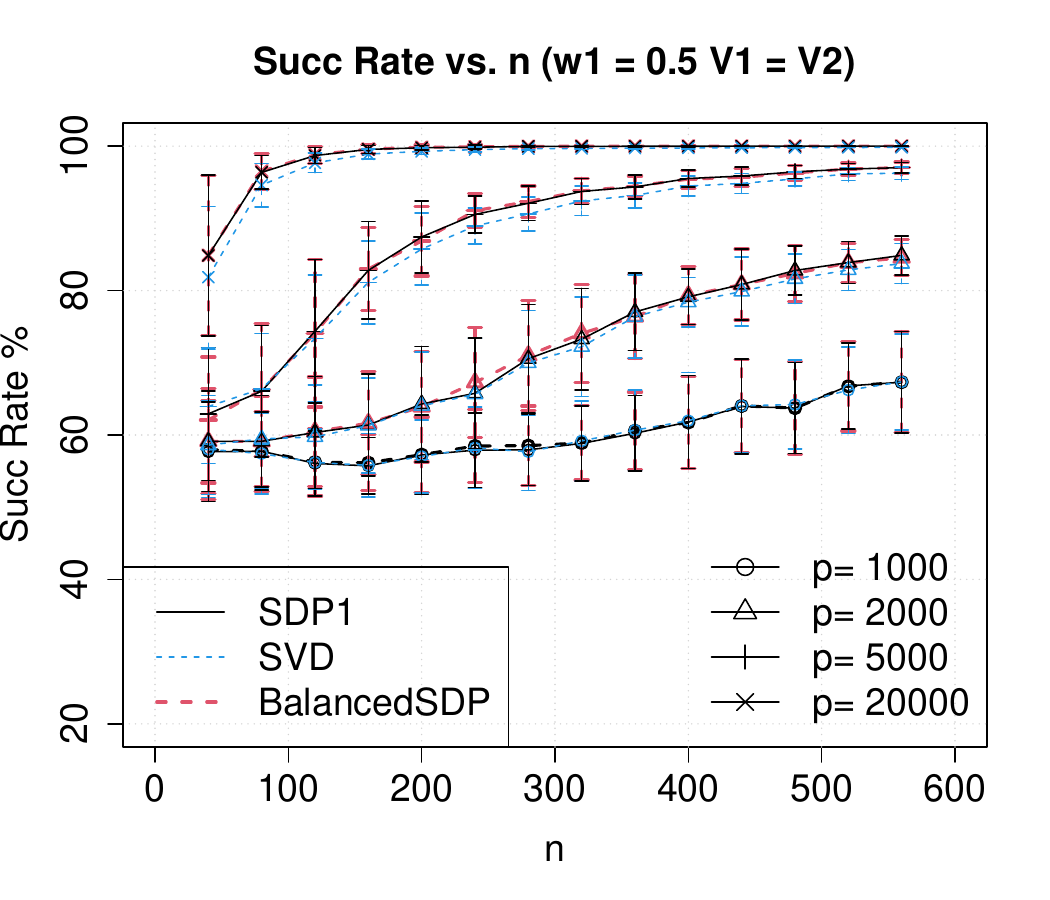}&                                                                                 
\includegraphics[width=2.8in]{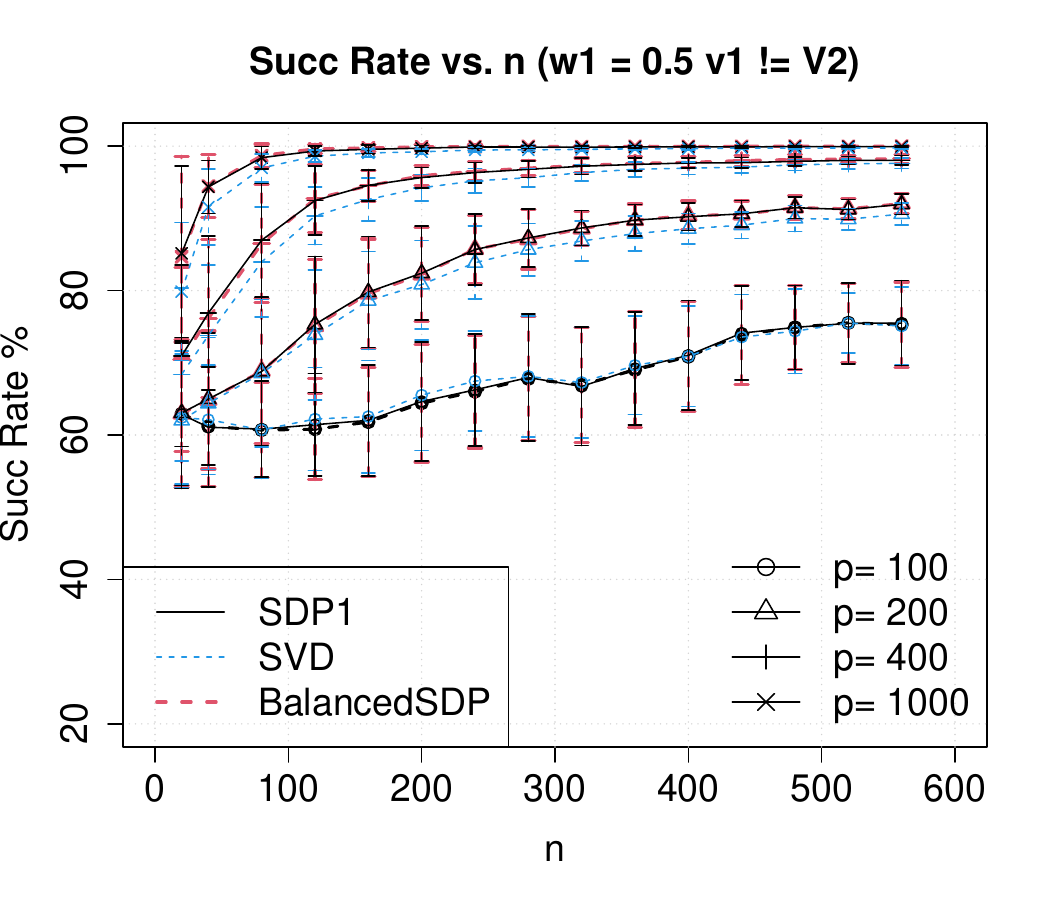}  \\
\end{tabular}
\caption{
  Top left: Success rates of SDP1 and SVD (balanced case and $V_1 = V_2$)
  for different $p$ as $n$ increases.
  Lines with the same markers are for the 2 estimators.
Red lines highlight the success rates at different levels
  of $n p \gamma^2$ ranging from  $0.5$ to $3.5$ (bottom to top with a
  step of $0.5$). The solid red line is for $n p \gamma^2 = 1$. 
Top right: Same data as the top left plot, showing SDP1's success rate
  for different $n$ as $p$ increases (x-axis in log scale).
Bottom row: Success rates of SDP1, SVD and BalancedSDP for two different variance profiles.
  Lines with the same markers are for the 3 estimators.
  Error bars represent 1 standard deviation.}
\vspace{-0.5cm}
\label{fig::succ-rate}
\end{figure*} 

\noindent{\bf Success rate and misclassification rate.}
For each experiment, we run 100 trials, and for each trial, we first generate a
data matrix $X_{n \times p}$ according to the mixture of two Bernoulli
distributions with parameters described above, and then feed
$Y$~\eqref{eq::defineY} to the three estimators for classification.  We measure
the success rate and misclassification rate based on $P$, the output assignment
vector. The success rate is computed as the number of correctly classified
individuals divided by the sample size $n$.  Hence the misclassification rate is $1
- $ success rate. Fig.~\ref{fig::succ-rate}(top left) shows the average success rates of SDP1
and SVD (over $100$ trials) for the balanced cases with $V_1 = V_2$
as $n$ increases for different values of $p$.  We observe that SDP1 has higher
average success rate for each setting of $(n, p)$ when $np\gamma^2 > 1.5$,
despite the exhaustive search in SVD. For $np\gamma^2 < 1.5$, the rates are
closer.  We also see that when $p < 1/\gamma = 625$, for example, when $p =
500$, the success rate remains flat across $n$.  Note that a success rate of
$50\%$ is equivalent to a total failure.  In contrast, when $n <1/\gamma = 625$, we
can obtain a higher success rate as $p$ increases, as shown in
Fig.~\ref{fig::succ-rate}(top right) for SDP1. In general, $np\gamma^2 > 1$ is indeed necessary
to obtain a success rate larger than $60\%$, when $p \ge 1/{\gamma}$.  When $n
< 625$, $np \gamma^2$ plays the role of the SNR, since $n p \gamma^2 < p
\gamma$. This remains the case throughout our experiments.

\noindent{\bf BalancedSDP.}
In the bottom row of Fig.~\ref{fig::succ-rate}, we compare the success rates of all three estimators
under two configurations. 
We observe the average success rates of BalancedSDP and SDP1 are essentially the
same for each setting of $(n,p)$ for both configurations.
These results are consistent with Conjecture~\ref{conj::balpar},
and confirm that the additional constraint in BalancedSDP
does not play a major role on the average success rate for the
current set of experiments.

\begin{figure*}
  \begin{centering}
      \vspace{-0.5cm}
  \begin{tabular}{ccc}
    \hspace*{-0.7cm}
          \vspace{-0.5cm}x
\includegraphics[width=1.82in]{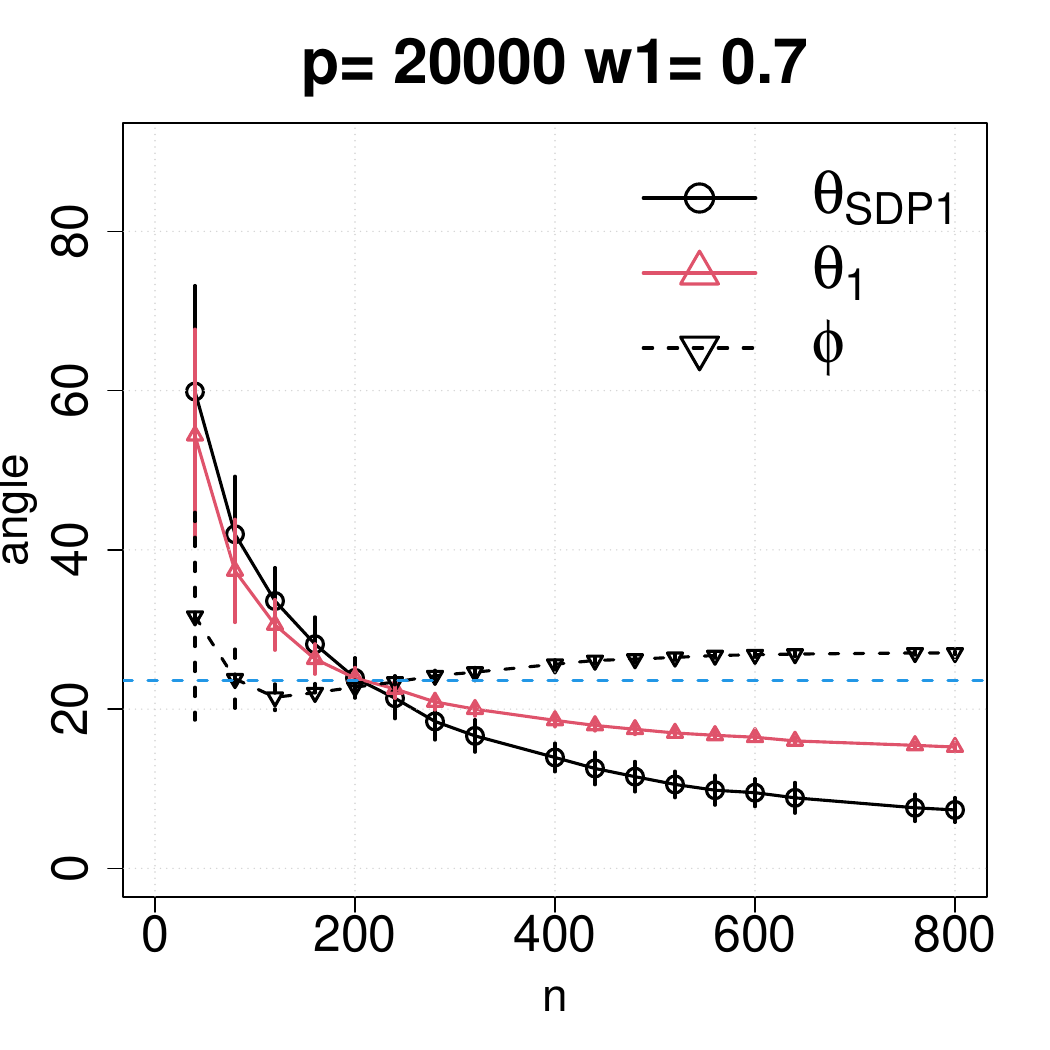} &
\includegraphics[width=1.82in]{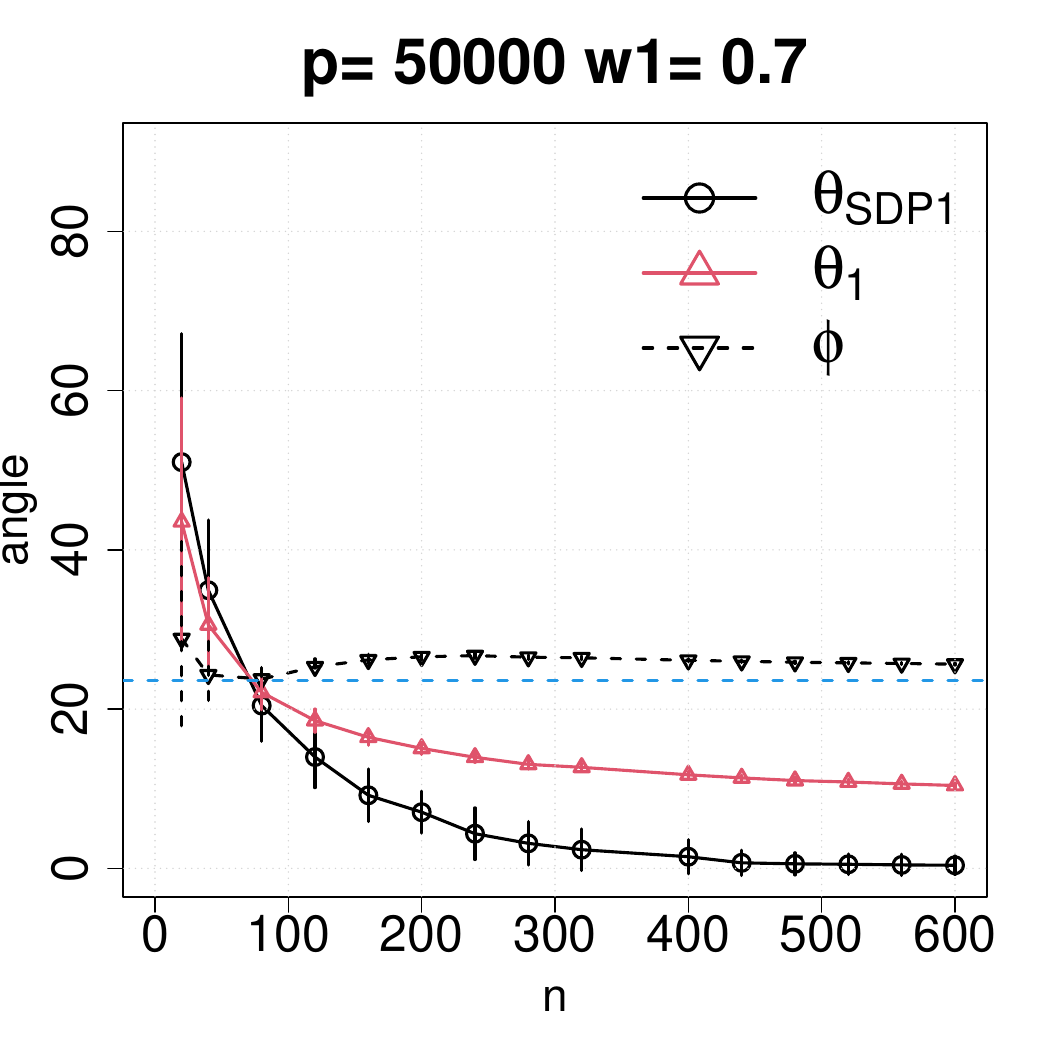} &
\includegraphics[width=1.82in]{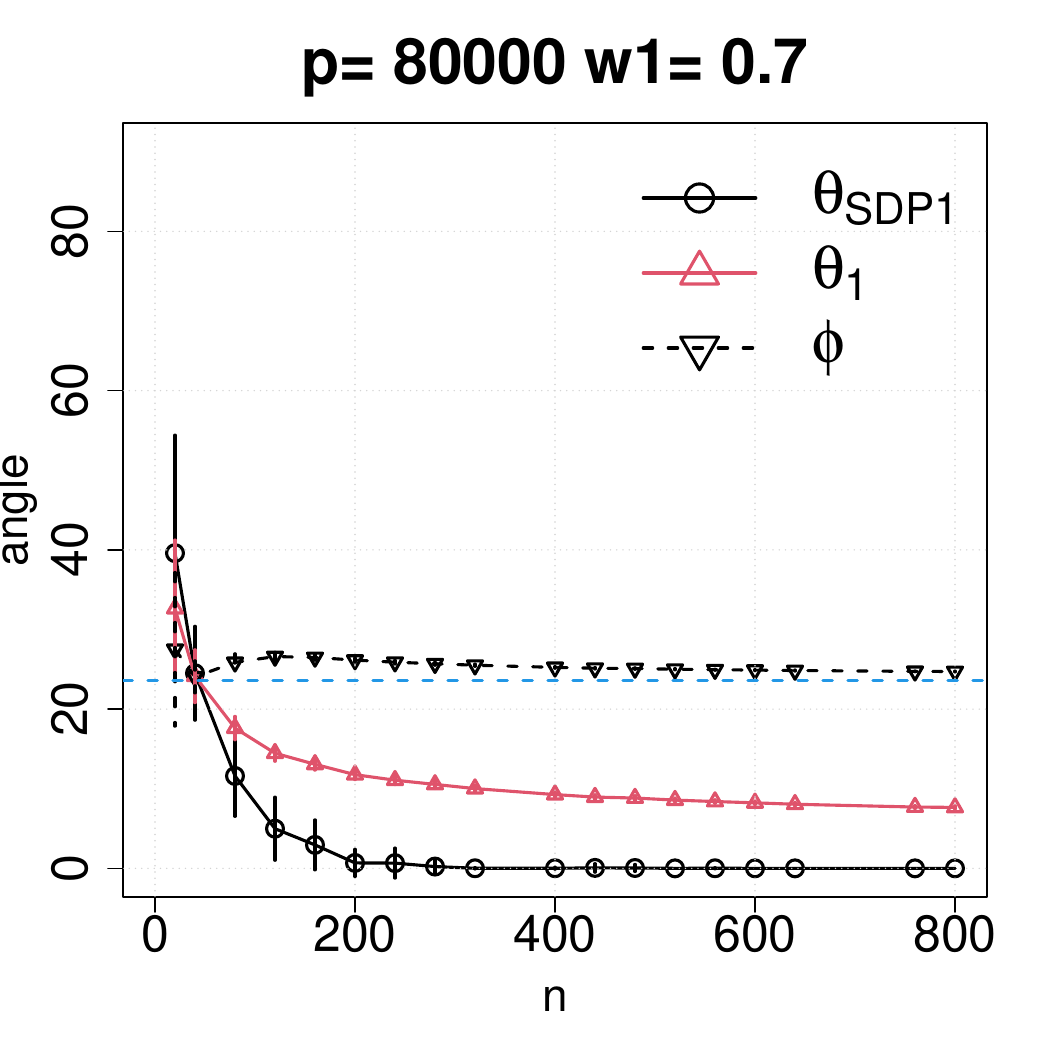} \\
      \vspace{-0.5cm}
\hspace*{-0.7cm}
\includegraphics[width=1.82in]{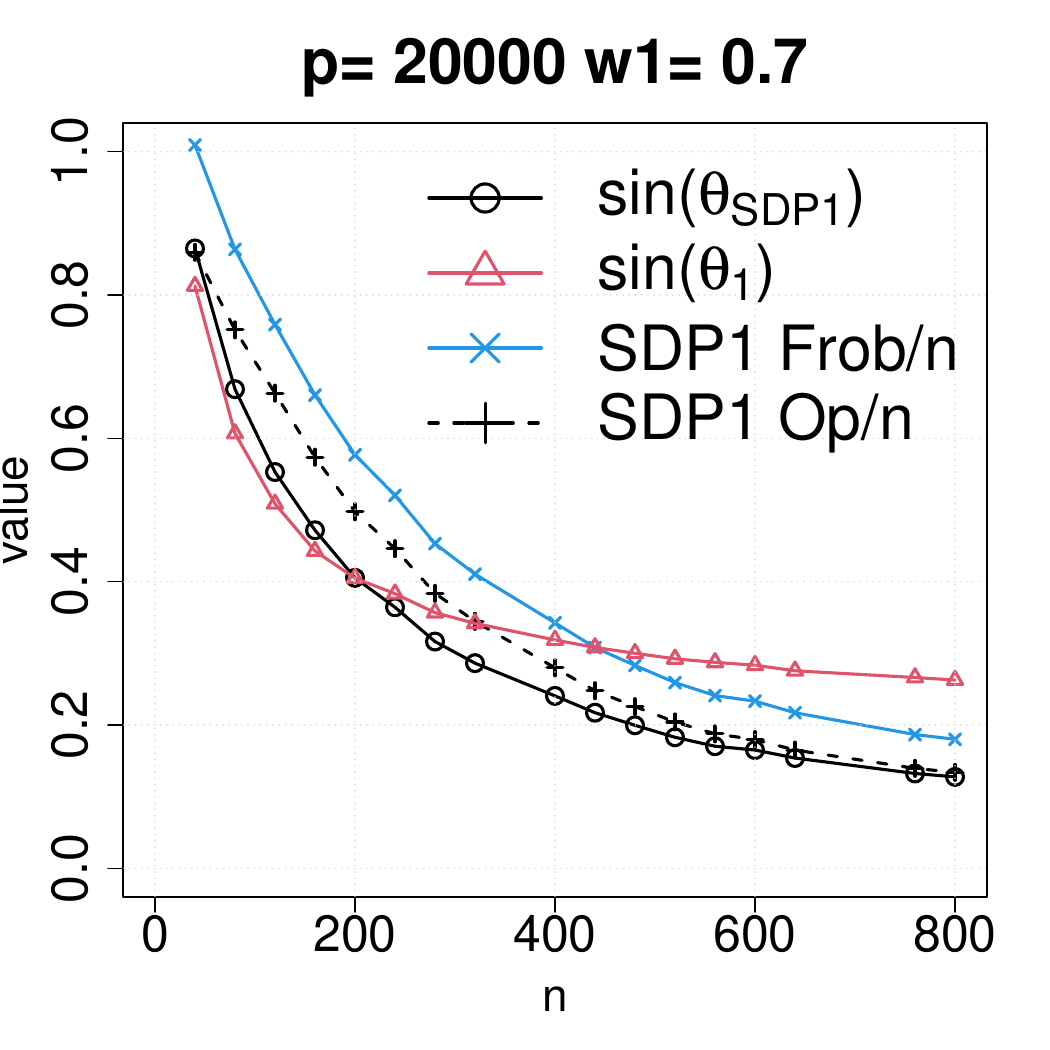} &
\includegraphics[width=1.82in]{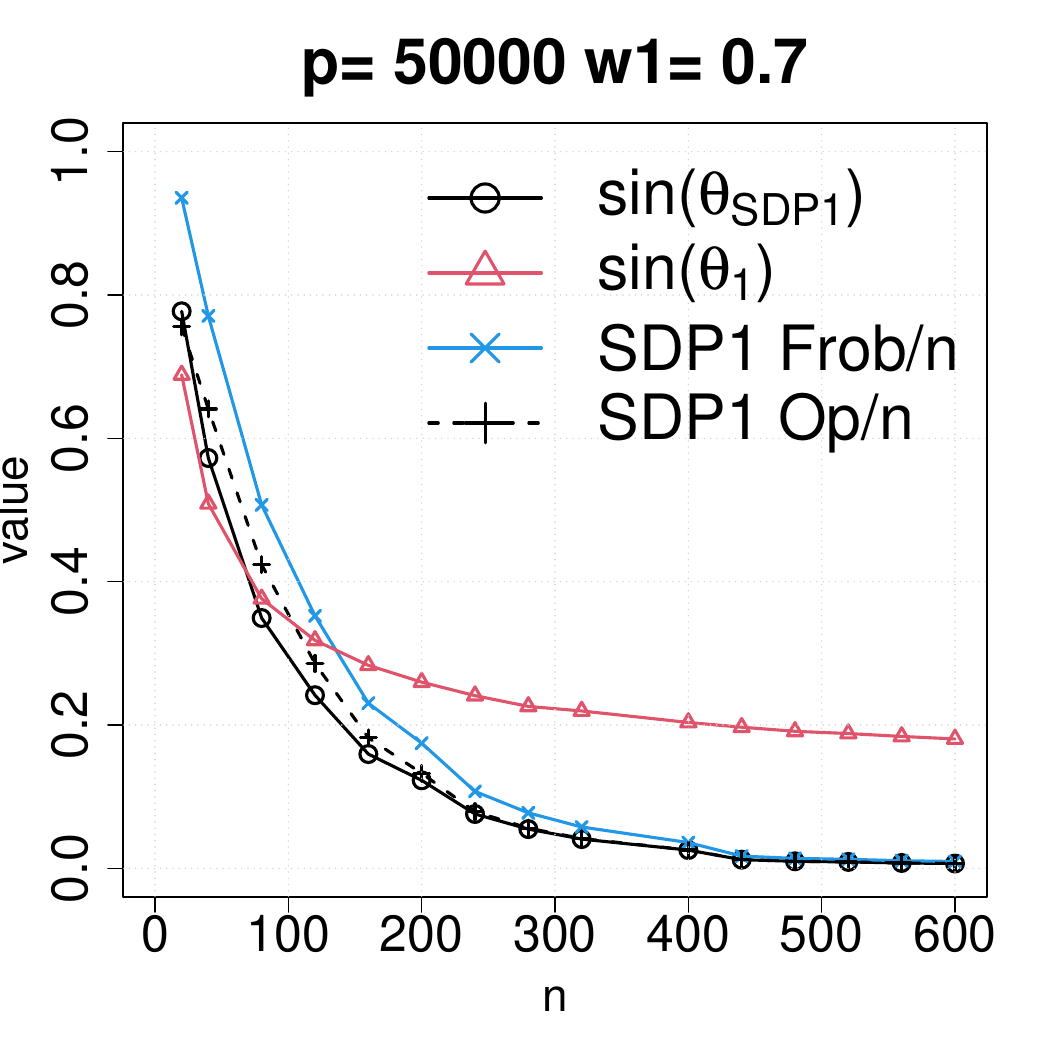} &
\includegraphics[width=1.82in]{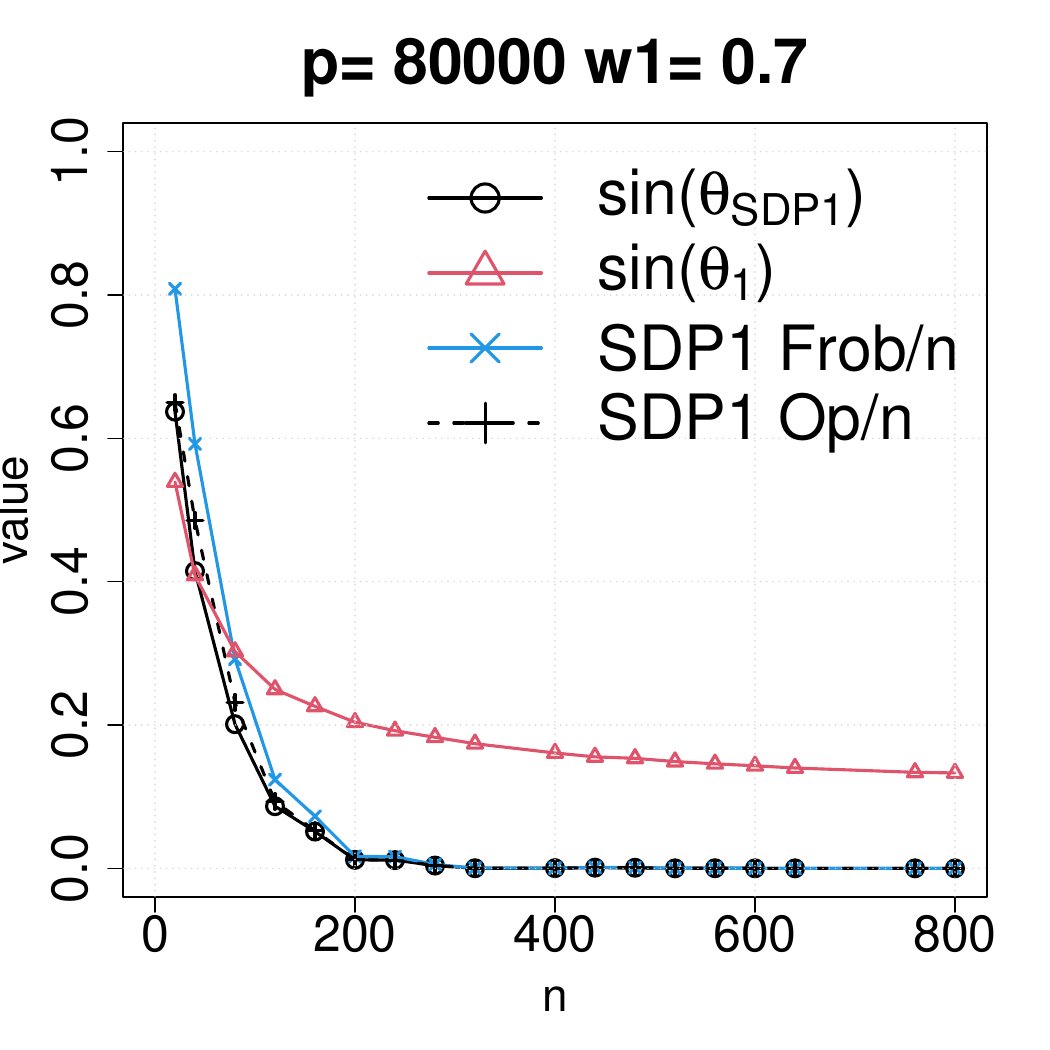} 
  \end{tabular}
\caption{Unbalanced case $w_1=0.7$,  $p \in \{20000, 50000, 80000\}$.
}
\label{fig::angle-norm-imb}
\end{centering}
\vspace{-0.5cm}
\end{figure*}

\noindent{\bf Angle and $\ell_2$ convergence.}
Here we take a closer look at the trends of $\hat{x}$ and $\hat{Z}$, the
solution to SDP1~\eqref{eq::hatZintro} as $n$ increases, and of $v_1$, the
leading eigenvector of $YY^T$.  In this experiment, we set $p \in \{20000,
50000, 80000\}$, and increase $n$.  In the top panel of
Fig.~\ref{fig::angle-norm-imb}, which is for the unbalanced case of $w_1=0.7$,
and $V_1 = V_2$, we plot $\theta_{\SDP1} := \angle(\hat{x}, u_2)$ between
$\hat{x}$ and its reference vector $u_2$ as defined in
Theorem~\ref{thm::SDPmain} and  Corollary~\ref{coro::misclass}.  For SVD,
$\theta_1 := \angle(v_1,\bar{v}_1)$ between $v_1$ and its reference
$\bar{v}_1$, where  $\bar{v}_1$ is as defined in Theorem~\ref{thm::SVD}.  In
this case, the angle $\angle(\bar{v}_1, u_2)$ between the two reference vectors
is about 22 degrees (blue horizontal dashed line).  We observe that as $n$
increases, for both algorithms, the angles $\theta_{\SDP1}$ and $\theta_1$
decrease, but $\theta_{\SDP1}$  drops much faster and decreases to $0$ when $n
> 200$ for $p = 80,000$.  We also show the angle $\phi = \angle(\hat{x}, v_1)$
between the two leading eigenvectors $\hat{x}$ and $v_1$, which largely remains
flat across all $n$.  In the bottom panel of Fig.~\ref{fig::angle-norm-imb}, we
see that for SDP1, all three metrics, namely $\sin(\theta_{\SDP1})$,
$\shnorm{Z^{*} - \hat{Z}}_2/{n}$, and $\shnorm{Z^{*} - \hat{Z}}_F/{n}$, where
$Z^{*} = u_2 u_2^T$, decrease as $n$ increases, following an exponential decay
in $n$ as predicted by our theory in Theorem~\ref{thm::exprate} and
Corollary~\ref{coro::misexp}, where in each plot, $p, \gamma$ are fixed. The
gaps between the three curves for SDP1 shrink when $p, n$ increase.  For SVD,
$\sin(\theta_1)$ also decreases as $n$ increases, but at a slower rate of
$1/n$, again as predicted by Theorem~\ref{thm::SVD}.

\section{Proof of Theorem~\ref{thm::exprate} and
  Lemma~\ref{lemma::onenorm} }
\label{sec::proofexprate}
Let $\lceil  r_1/(2n) \rceil =: q< n$ and $r_1 :=\onenorm{\hat{Z}  -
  Z^{*}}$. Recall $\sigma_{\max}^2 : = \max_{i} \twonorm{H_i}^2$.
Then, we have by Lemma~\ref{lemma::YYlocal}, on event $\MG_1$,
  \ben 
\label{eq::eventG1} 
\lefteqn{\sup_{\hat{Z} \in \M_{\opt} \cap (Z^{*} + r_1
    B_1^{n \times n})}  \abs{\ip{Y Y^T - \E(YY^T), \hat{Z} - Z^{*} }}
  \le  \frac{5}{6}\xi p \gamma \onenorm{\hat{Z} - Z^{*}}  }\\
\nonumber
&& + C' C_0 \sigma_{\max}
n \sqrt{p \gamma}   + C_1 C_0^2 \sigma_{\max}^2
\sqrt{n p }  \left\lceil \frac{r_1}{2n}\right\rceil \sqrt{\log (2e n/ \lceil    \frac{r_1}{2n}\rceil)}.
\een
Let $C,  C', C_1, c, c_1, c_2,....$ be absolute positive constants.
For the rest of the proof, we assume  $\MG_{1} \cap \MG_{2}$ holds,
where $\prob{\MG_{2}\cap \MG_1} \ge 1- \exp(cn)-c/n^2$.
By Lemmas~\ref{lemma::signalgrad},~\ref{lemma::optlocal2}, and~\ref{lemma::onenorm}, 
\eqref{eq::event0}, and \eqref{eq::eventG1}, we have  for all $\hat{Z}
\in \M_{\opt} \cap (Z^{*} + r_1 B_1^{n\times n})$,
\ben
\label{eq::Rleft}
0 & \le &  p \gamma w_{\min}^2 \onenorm{Z^{*} - \hat{Z}} \le  \ip{R, Z^{*} - \hat{Z}}  \\
& \le &
\nonumber
\abs{\ip{Y Y^T - \E(YY^T), \hat{Z} - Z^{*} }} + \abs{F_2} + \abs{F_3}
\le (10/3) \xi  p \gamma \onenorm{\hat{Z} - Z^{*}} 
\\
\nonumber
& &
+ C' (C_0 \sigma_{\max}
q \sqrt{\log (\frac{2e n}{q})} \cdot 
\big(n \sqrt{p \gamma} + C_0 \sigma_{\max} \sqrt{n p}\big). 
\een
Hence we can control $\xi \le w_{\min}^2 /16$ and the
lower bound in~\eqref{eq::NKlower} so that each 
component $F_2, F_3$ is only a fraction of the signal at the level of 
$p \gamma w_{\min}^2\onenorm{(Z^{*} -  Z)}$. By moving $(10/3)\xi  p
\gamma \onenorm{\hat{Z} - Z^{*}} < \frac{5}{24}  p \gamma  w_{\min}^2
r_1$, where $\xi \le  w_{\min}^2/16$, to the LHS of \eqref{eq::Rleft},
we have
\ben
\nonumber
\lefteqn{
  \frac{19}{24} p \gamma w_{\min}^2 r_1
\le  p \gamma w_{\min}^2 \onenorm{Z^{*} - \hat{Z}} -   (10/3) \xi  p
\gamma \onenorm{\hat{Z} - Z^{*}} }\\
\label{eq::finalcond}
& \le  & C C_0 \sigma_{\max}
\sqrt{\log (2e n/q)} \cdot 
\big(2   n q \sqrt{p \gamma}  + C_0 \sigma_{\max} q \sqrt{n p }\big).
\een
Then $q$ must satisfy one of the following two conditions to ensure~\eqref{eq::finalcond}:  for $r_1 \le 2n q$,
\bit
\item[1.]
Under the assumption that 
  $\Delta^2 w_{\min}^4 = p \gamma w_{\min}^4 = \Omega(C^2_0  \sigma_{\max}^2)$,
  suppose
  \bens
\sqrt{p \gamma} w_{\min}^2 
  & \le  &  C_1 C_0 \sigma_{\max} \sqrt{\log (2e n/q)},  
  \eens
  which implies that $\frac{c p \gamma w_{\min}^4}{C^2_0  \sigma_{\max}^2}
  \le \log (2e n/q)$  and hence
  $q \le  n \exp\big(-\frac{c_1 \Delta^2  w_{\min}^4}{C^2_0 \sigma_{\max}^2}\big).$
\item[2.]
Alternatively, under the assumption that  $n p \gamma^2
w_{\min}^4=\Omega( C_0^4)$, we require
\bens 
p \gamma w_{\min}^2 r_1 & \le &
2 qn p \gamma w_{\min}^2 \le
C_2  C^2_0 \sigma_{\max}^2 \sqrt{\log (2e n/q)} q \sqrt{n p}, 
\eens
and hence $n p \gamma^2 w_{\min}^4 \le  C_3 (C_0 \sigma_{\max})^4
\log (2e n/q)$,  which implies that
\bens
\log (q/2e n)  \le  - \frac{C' n p 
  \gamma^2 w_{\min}^4}{C^4_0 \sigma^4_{\max}},
\; \text{ resulting in }\; q \le n \exp\big(- \frac{c_2 n p \gamma^2
  w_{\min}^4}{C^4_0 \sigma^4_{\max}}\big).
 \eens
\eit
Putting things together, we have for $s^2$ as defined 
  in~\eqref{eq::SNR2},
\bens
\label{eq::localSNR}
\onenorm{Z^{*} - \hat{Z}} := r_1 \le 2 q n \le  n^2 \exp(-C s^2
w_{\min}^4), \; \text{ and } \;  q \le n \exp(- C s^2 w_{\min}^4). \quad \scriptstyle\Box
\eens

\subsection{Proof of Lemma~\ref{lemma::onenorm} }
\label{sec::proofofmain1}
\label{sec::proofofonenorm}
\begin{proofof2}
We will prove that  \eqref{eq::Rlower} holds for all $Z \in [-1,  1]^{n \times n} \supset \M_{\opt}$.
 Recall we have
\bens
\nonumber
 R= \E(Y) \E(Y)^T
 &= &
p \gamma 
\left[
\begin{array}{cc}
  w_2^2 E_{n_1} & - w_1 w_2E_{n_1 \times n_2} \\
 - w_1 w_2 E_{n_2 \times n_1} & w_1^2 E_{n_2} 
\end{array}
\right] =:                                            
p \gamma 
\left[
\begin{array}{cc} \mathbb{A} & \mathbb{B}\\
 \mathbb{B}^T & \mathbb{C}
\end{array}
\right].
\eens
Now all entries of $Z^{*}, Z$ belong to $[-1, 1]$.
Clearly, for the upper left and lower right diagonal blocks, denoted by $\D = \{\mathbb{A}, \mathbb{C}\}$, we have
$Z^{*} - Z \ge 0$, since all entries of $Z^{*}$ on these blocks are $1$s.
Similarly, for the off-diagonal blocks $\{\mathbb{B}, \mathbb{B}^T\}$,
we have $Z-Z^{*} \ge 0$.
Thus we have
\bens
\ip{R, Z^{*} - Z}/(p \gamma)
& \ge &
(w_2^2 \wedge w_1^2 \wedge w_1 w_2)
\big(
  \sum_{(i,j) \in \D} (Z^{*} - Z)_{ij} + \sum_{(i,j) \in \{\mathbb{B},
    \mathbb{B}^T\}}  (Z- Z^{*})_{ij}  \big) 
\eens
The lemma thus holds since
$\sum_{(i,j) \in \D} (Z^{*} - Z)_{ij}  + \sum_{(i,j) \in \{\mathbb{B},
  \mathbb{B}^T\}}  (Z- Z^{*})_{ij}  =  \onenorm{Z- Z^{*}}.$
\end{proofof2}

\section{Proof of Corollaries~\ref{coro::misclass} and~\ref{coro::misexp}}
\label{sec::proofofmisclass}
\begin{theorem}\textnormal{\bf{(Davis-Kahan)}}
\label{thm::DK}
For $A, M \in \R^{n \times n}$ being symmetric matrices and $E = M - A$. 
Let $\lambda_1(A) \geq \lambda_2(A) \geq \ldots \geq \lambda_n(A)$ 
be eigenvalues of $A$ with orthonormal eigenvectors $v_1, v_2, \ldots, v_n$, and 
$\lambda_1(M) \geq  \lambda_2(M) \geq \ldots \geq \lambda_n(M)$ be eigenvalues of $M$ with
orthonormal eigenvectors $w_1, w_2, \ldots, w_n$ of $M$, with $\theta_i = \angle(v_i, w_i)$. Then
\begin{gather}
\label{eq:eigen-vectors}
\theta_i \sim \sin(\theta_i) \leq 
\frac{2\twonorm{E}}{\gap(i, A)} \; \text{ where} \; \; \gap(i, A) = \min_{j \not= i}\abs{\lambda_i(A) - \lambda_j(A)}.
\end{gather}
\end{theorem}
Theorem~\ref{thm::DK} is a well-known result in perturbation theory. 
See~\cite{BCFZ09} for a proof. See also Theorem 4.5.5~\cite{Vers18}
and Corollary 3 in~\cite{YWS15}.

\subsection{Proof of Corollary~\ref{coro::misclass}}
The upper bound follows from \eqref{eq::hatZFnorm} (Theorem~\ref{thm::SDPmain}) and Corollary 3~\citep{YWS15},
while noting that  the largest eigenvalue of $u_2 u_2^T$ is $n$ while all
  others are 0, and hence the spectral gap in the sense of Theorem~\ref{thm::DK} equals $n$.
  Then, we have by Theorem~\ref{thm::SDPmain},
  \bens
\min_{\alpha = \pm 1}
\twonorm{(\alpha \hat{x} - u_2)/\sqrt{n}}^2 \le \frac{2^{3} \twonorm{\hat{Z} - u_2 u_2^T}^2}{\gap(1, u_2 u_2^T)^2} 
  \le   {2^{3} \fnorm{\hat{Z} - u_2 u_2^T}^2}/{n^2}  \le   2^{3}   {4  K_G  \xi}/{w_{\min}^2}.  \; \; \scriptstyle\Box
  \eens

\subsection{Proof of Corollary~\ref{coro::misexp}}
  The angle between $\hat{x}/\sqrt{n}$ and $u_2/\sqrt{n}$ can be expressed as
  \begin{eqnarray}
  \label{eq::angsdp}
  \cos(\theta_{\SDP}) = \cos(\angle(\hat{x}, u_2)) 
  & = & \ip{\hat{x}, u_2}/n
  \end{eqnarray}
The upper bound on $\sin(\theta_{\SDP})$ follows from 
Theorems~\ref{thm::exprate} and~\ref{thm::DK};
By Davis-Kahan Theorem, cf. Corollary 3~\citep{YWS15}, we
have with probability at least $1-2 \exp(-c n) - 2/n^2$,
\bens
\min_{\alpha = \pm 1}
\twonorm{(\alpha \hat{x} - u_2)/\sqrt{n}}
  &  \le & \frac{2^{3/2} \twonorm{\hat{Z} - u_2 u_2^T}}{\gap(1, u_2 u_2^T)} 
  \le   {2^{3/2} \fnorm{\hat{Z} - u_2 u_2^T}}/{n} \\
    &  \le &   2^{3/2}   (2 \onenorm{\hat{Z} - u_2 u_2^T})^{1/2}/{n}  \le 
   4 \exp(-c_0 s^2 w_{\min}^4/2). \; \; \scriptstyle\Box
   \eens

\section{Proof of Theorem~\ref{thm::SVD}}
\label{sec::proofofSVD}
We first state  Propositions~\ref{fact::topeigen} and~\ref{fact::fullbasis}.
  \begin{proposition}
  \label{fact::topeigen}
Let $YY^T = \sum_{j=1}^{n-1} \lambda_j v_j v_j^T$.
  Denote by $\tilde{A} = YY^T - \lambda E_n$, then
 \ben
  \label{eq::eigenA}
\tilde{A} & := & YY^T - \lambda E_n = 
\sum_{j=1}^{n-1} \lambda_j v_j v_j^T + \frac{n \tau}{n-1} \vecone
\vecone^T/n \succeq 0.
\een
The leading eigenvector of $\tilde{A}$ (resp. $A$ and $B$) will
coincide with that of $YY^T$ with
 \ben
 \label{eq::eigenYY}
 \lambda_{\max}(\tilde{A}) & = & \lambda_{\max}(YY^T)   \ge 
n \tau/(n-1),
\een
where strict inequality holds if and only if not all eigenvalues of
$YY^T$ are identical.
Thus the symmetric matrices $A = \tilde{A}  + \lambda I_n$ and $B = A
+ \E \tau I_n$ also share the same leading eigenvector $v_1$ with
$YY^T$, so long as not all eigenvalues of $YY^T$ are identical, with $\lambda_{\max}(A) \ge \tau$.
\end{proposition}

Proposition~\ref{fact::fullbasis} is also not surprising, since the 
sum of all off-diagonal entries of $A$ is 0.  We need it for the proof
of Lemma~\ref{lemma::TLbounds} and Theorem~\ref{thm::SVD}.
\begin{proposition}
\label{fact::fullbasis}
Suppose that we observe one instance of the 
gram matrix $\hat{S}_n := X X^T$. Recall
$\lambda = \inv{{n \choose 2}} \sum_{i<j} \ip{Y_i, Y_j}
\;\; \text{ and } \; \; \tau =\inv{n} \sum_{i=1}^n \ip{Y_i, Y_i}.$
Then
\ben
\label{eq::YYgram}
\ip{YY^T, E_n} = \vecone^T_n YY^T \vecone_n = 0, \;\text{ since}\;
YY^T  =   (I-P_1) XX^T  (I-P_1).
\een
Moreover, by construction, we have for $A$ as defined in~\eqref{eq::defineAintro},
\bens
\ip{A, E_n}
& = & \vecone_n^T YY^T \vecone_n - \lambda \ip{ (E_n -I_n), E_n}  = -
\lambda n (n-1) =\tr(YY^T) \\
\text{ where}  \; \; \lambda & = & \inv{n(n-1)}\sum_{i\not= j} \ip{Y_i, Y_j} 
=-\frac{\tr(YY^T)}{n (n-1) }= -\frac{\tau}{n-1}.
\eens 
In other words, we have $\ip{A, P_1} = \ip{A, \vecone_n \vecone_n^T/n}
= \tr(YY^T)/n =: \tau.$
\end{proposition}

\begin{proofof}{Theorem~\ref{thm::SVD}}
  It is known that for any real symmetric matrix, there exists a set of $n$ 
  orthonormal eigenvectors.
  Recall that $R$ is rank one with 
  $\lambda_{\max}(R) = \tr(R) = w_1 w_2 n p \gamma$ while
 $u_2 R u_2 /n =
  (4 w_1 w_2) w_1 w_2 n p \gamma \le \inv{4} n p \gamma.$
Hence $u_2/\sqrt{n}$ coincides with $\bar{v}_1$ when $w_1 = w_2 =
1/2$.
Hence for $\gap(1, R)$, as defined in Theorem~\ref{thm::DK},
$\gap(1, R) = \lambda_{\max}(R) = w_1 w_2 n p \gamma$.
We check the claim that the leading eigenvector of $B$ coincides 
with that of $YY^T$ in Proposition~\ref{fact::topeigen}.
Clearly,
\begin{eqnarray}
  \label{eq::angsvd}
  \cos(\theta_1) = \cos(\angle({v}_1, \bar{v}_1))
  & = & \ip{v_1, \bar{v}_1}
\end{eqnarray}
and hence  $\theta_1 = \arccos (\ip{v_1, \bar{v}_1})$.
  Hence we can use the first eigenvector of $YY^T$ to partition the
  two groups of points in $\R^{p}$.   To obtain an upper bound on
  $\sin(\theta_1)$, we apply the Davis-Kahan perturbation bound as
  follows.  Since $v_1, \bar{v}_1$ are the leading eigenvectors of $B$ and $R$
  respectively, \eqref{eq::angSVD} holds by
  Theorems~\ref{thm::reading} and~\ref{thm::DK}:
  \bens
\sin(\theta_1) & := & \sin(\angle({v}_1, \bar{v}_1)) \le 
\frac{2 \twonorm{B - R}}{\lambda_{\max}(R)} = 
\frac{2 \twonorm{B - R}}{w_1 w_2 n p \gamma} \le \frac{2 \xi}{w_1 w_2}.
\eens
Moreover, we have~\eqref{eq::normSVD} holds by Corollary
3~\citep{YWS15}: since
\bens
  \min_{\alpha=\pm 1}  \twonorm{\alpha v_1 -  \bar{v}_1}^2
  & \leq & \left(\frac{2^{3/2} \twonorm{B - R}}{w_1 w_2 n p \gamma}
  \right)^2  \leq  \frac{2^{3} \xi^2}{(w_1 w_2)^2} =: \delta', \text{ where} \; \; \delta' = \frac{8  \xi^2}{(w_1^2
    w_2^2)} \le \frac{c_2 \xi^2}{w_{\min}^2}.
\eens
The theorem thus holds.
\end{proofof}


\begin{proofof}{Proposition~\ref{fact::topeigen}}
  Clearly,  the additional terms involving $E_n$ and $I_n$ are either 
orthogonal to eigenvectors $v_1, \ldots, v_{n-1}$ of $YY^T$, or 
act as an identity map on the subspace spanned by   $\{v_1, \ldots, 
v_{n-1}\}$. 
Now   \eqref{eq::eigenA} holds since $\ip{v_j, \vecone_n}=0$ for all
$j$ and hence $\{v_1, \ldots, v_{n-1} , \vecone_n/\sqrt{n}\}$ forms
the set of orthonormal eigenvectors for $\tilde{A}$ (resp. $A$ and
$B$), and moreover, in view of Proposition~\ref{fact::fullbasis},
\bens
- \lambda E_n & = & \frac{n \tau}{n-1} P_1 =\frac{n 
  \tau}{n-1} \vecone_n  \vecone_n^T/n,  \; \; \text{ where } \;\; \lambda =-\frac{\tau}{n-1}. 
\eens
Since we have at most $n-1$ non-zero eigenvalues  
and they sum up to be $\tr(YY^T)$, we have
\bens
\lambda_{\max}(YY^T) & \ge &  \tr(YY^T)/(n-1) = n \tau/(n-1),
\eens
where strict inequality holds when these eigenvalues are not all  
identical. Finally,~\eqref{eq::eigenYY} holds since
$\lambda_1(\tilde{A}) := \lambda_{\max}(YY^T)$ in view of the eigen-decomposition~\eqref{eq::eigenA} and the
displayed equation immediately above.
Now for $A =  YY^T - \lambda (E_n -I_n)$, we have $\tr(A) =
\tr(YY^T)$, and hence $\lambda_{\max}(A) \ge \tau$.
The extra terms $\propto I_n$ in $A$ (resp. $B$)
will not change the order of the sequence of eigenvalues for $B$ (resp. $A$) with
respect to that established for $\tilde{A}$. Hence all symmetric matrices
$B$, $\tilde{A}$, and $A$ share the same leading eigenvector $v_1$
with $YY^T$.
\end{proofof}

\section{Proof of key lemmas for Theorem~\ref{thm::exprate}}
\label{sec::keylemmas}

\subsection{Proof of Lemma~\ref{lemma::signalgrad}}
\label{sec::proofofsignal}
\begin{proposition}
  \label{fact::trP2ZZ}
  Let $Z^{*} =   u_2 u_2^T$ be as defined in
  Lemma~\ref{lemma::ZRnormintro}.
  Then for $P_2 = Z^{*}/n$, 
  \ben
  \label{eq::refoptsol}
 \tr(P_2 (Z^{*} - \hat{Z})) = 
\inv{n} \tr(Z^{*} (Z^{*} - \hat{Z})) =
\inv{n} \shnorm{Z^{*} -
    \hat{Z}}_1 \le 2 (n-1).
  \een
\end{proposition}
\begin{proofof}{Lemma~\ref{lemma::signalgrad}}
  Denote by $\tilde{A} =Y Y^T -\lambda E_n$.
We have for $B$ as in~\eqref{eq::defineBintro},
\ben
\label{eq::redefineB}
\E B -R & := & \E (Y Y^T) -\E(Y)\E(Y)^T- \E \lambda (E_n - I_n) - \E \tau I_n,
\een
where $R = \E(Y)\E(Y)^T$, 
Moreover, we have by \eqref{eq::redefineB},
\ben
\label{eq::Wbound}
M & := &  \E(YY^T)
  -  \lambda (E_n - I_n) - \E \tau
  I_n - \E(Y)\E(Y)^T  \\
\nonumber
  & =  & \big( \E B - R\big) -  (\lambda - \E  \lambda ) (E_n - I_n). 
\een
Clearly $\hat{Z}, Z^{*} \in 
\M_{\opt}$ by definition.
By optimality of $\hat{Z} \in \M_{\opt}$, we have by~\eqref{eq::defineBintro} and \eqref{eq::sdpmain},
\bens
\ip{B, \hat{Z} - Z^{*}  } & := & \ip{Y Y^T -\lambda (E_n - I_n) - \E \tau 
  I_n,  \hat{Z} - Z^{*}}   = \ip{Y Y^T -\lambda E_n,  \hat{Z} - Z^{*}
} \ge 0, \\
\ip{R, Z^{*} - \hat{Z}}  & := & \ip{-\E(Y)\E(Y)^T, \hat{Z} - Z^{*} }\\
  & \le &
  \ip{Y Y^T  -  \lambda (E_n - I_n) - \E \tau 
    I_n,   \hat{Z} - Z^{*} } - \ip{\E(Y)\E(Y)^T, \hat{Z} - Z^{*} }\\
  & = &
  \ip{Y Y^T - \E(YY^T), \hat{Z} - Z^{*} } + \ip{M, \hat{Z} - Z^{*}
  },\;\text{ for $M$ as  in  \eqref{eq::Wbound}.}
  \eens
  Thus the RHS of
  \eqref{eq::signal} holds. Now the LHS of  \eqref{eq::signal} holds by
Lemma~\ref{lemma::onenorm}.
\end{proofof}

\subsection{Proof sketch of Lemma~\ref{lemma::YYlocal}}
\label{sec::proofofYYlocalmain}
Let $\Lambda = YY^T - \E YY^T$ and  $\Psi := \Z \Z^T - \E \Z \Z^T$. 
The local analysis on~\eqref{eq::signal2} relies on the operator norm
bound we obtained in Theorem~\ref{thm::YYaniso} as already shown in
Lemma~\ref{lemma::optlocal}.
The key distinction from the global analysis
is that we need to analyze the projection operators in the following sense.
Following the proof idea of~\cite{FC18},
we define the following projection operator:  \text{for }  $P_2 = Z^{*} /{n}$,
\bens
\cp: M \in \R^{n \times n} &\rightarrow& P_2 M + M P_2 - P_2 M P_2 \\
\cp^{\perp}: M \in \R^{n \times n} & \rightarrow &
M-\cp(M) = (I_n-P_2) M (I_n -P_2).
\eens
We use the following decomposition to bound
for all $\hat{Z} \in \M_{\opt} \cap (Z^{*} + r_1 B_1^{n \times n})$,
where $\M_{\opt} := \{Z: Z \succeq 0, \diag(Z) = I_n\} \subset [-1, 1]^{n \times n}$:
\bens
\ip{\Lambda, \hat{Z} -Z^{*}}
& = &
\ip{\Lambda, \cp(\hat{Z} -Z^{*})}  +
\ip{\Lambda, \cp^{\perp}(\hat{Z} -Z^{*})} =: S_1(\hat{Z}) +
S_2(\hat{Z}), \\
\text{where }  \quad
S_1(\hat{Z}) & = & \ip{\Lambda, \cp(\hat{Z} -Z^{*})}   =
\ip{\cp(\Lambda), \hat{Z} -Z^{*}}, \\
S_2(\hat{Z})
& = &
\ip{\Lambda, \cp^{\perp}(\hat{Z} )} \le  \twonorm{\Lambda} \shnorm{Z^{*} -  \hat{Z}}_1/n. 
\eens
First, we control $S_1(\hat{Z})$ uniformly for all $\hat{Z} \in 
\M_{\opt} \cap (Z^{*} + r_1 B_1^{n \times n})$ in
Lemma~\ref{lemma::S1}.
We then obtain an upper bound for $S_2(\hat{Z})$ in
Lemma~\ref{lemma::S2}, uniformly for all $\hat{Z} \in \M_{\opt}$.
Lemmas~\ref{lemma::S1} and~\ref{lemma::S2} show that 
the total {\it noise} contributed by the second term, namely, $S_2(\hat{Z})$ and 
a portion of that by $S_1(\hat{Z})$ is only a fraction of the signal strength on the LHS of~\eqref{eq::signal}.
We prove Lemma~\ref{lemma::S1} in Section~\ref{sec::proofofS1main}.
Lemma~\ref{lemma::YYlocal} follows
from Lemmas~\ref{lemma::S1} and~\ref{lemma::S2}.
Proof of Lemma~\ref{lemma::S2} follows from arguments in~\cite{FC18} and  Theorem~\ref{thm::YYaniso}; hence omitted.
\begin{lemma}
   \label{lemma::S1}
Suppose all conditions in Theorem~\ref{thm::YYaniso} hold.
Let $\sigma_{\max} := \max_{i} \twonorm{H_i}$ be as in~\eqref{eq::sigmax}.
Then, with  probability at least $1-{c'}/{n^2}$, for some absolute constants $c', C$,
\bens
\sup_{\hat{Z} \in \M_{\opt} \cap (Z^{*} + r_1 B_1^{n \times n})}
S_1(\hat{Z})
\le  \frac{2}{3} \xi p \gamma \shnorm{\hat{Z} -Z^{*}}_1 +C \sigma_{\max} \lceil \frac{r_1}{2n}\rceil
\sqrt{\log (2e n/\lceil \frac{r_1}{2n}\rceil)} \big(n \Delta +\sigma_{\max}   \sqrt{np} \big).
\eens
\end{lemma}

\begin{lemma}
  \label{lemma::S2}
  Suppose all conditions in Theorem~\ref{thm::YYaniso} hold.
  Let $\xi \le  w_{\min}^2/16$. Then, with probability at least $1-\exp(c n)$, for all $\hat{Z} \in \M_{\opt}$,
\bens
S_2(\hat{Z})
& \le &
\shnorm{YY^T -\E(YY^T)}_2 \shnorm{Z^{*} -  \hat{Z}}_1/n  \le  \xi p \gamma \norm{\hat{Z} -Z^{*}}_1/6.
 \eens
\end{lemma}

\subsection{Proof of Lemma~\ref{lemma::optlocal}}
\label{sec::proofofoptlocal}
To be fully transparent, we now decompose the bias term into three components.
Intuitively, $W_0, W_2$ and $\mathbb{W}$ arise due to the imbalance in 
variance profiles. Hence (A2) is needed to control this bias.
Proof of Proposition~\ref{prop::biasfinal} appears in 
Section~\ref{sec::biasproofEBR}.

\begin{proposition}{\textnormal{{\bf (Bias decomposition)}}}
  \label{prop::biasfinal}
  Denote by 
\ben
\label{eq::defineW2}              
 W_0   & =&
(V_1 -V_2) \left[\begin{array}{cc} w_2 I_{n_1} & 0 \\
0 & -w_1 I_{n_2} \end{array}    \right]\quad \text {and} \quad
 W_2  := 
\frac{V_1-V_2}{n}
\left[
\begin{array}{cc}
E_{n_1} &  0 \\
0  & - E_{n_2} 
\end{array}
\right].
\een 
Then for $W_0, W_2$ as defined in~\eqref{eq::defineW2},
\ben
\label{eq::wow}
\E B - R 
& =&  W_0 - \mathbb{W} -{\tr(R)}/{(n-1)} (I_n - {E_n}/{n}),\quad
\text{ where} \; \; \\
\label{eq::wow2}
\mathbb{W}
& := & W_2 
+
\frac{ (V_1 - V_2)(w_2 - w_1)}{n} E_n, \quad \text{and when } V_1 = V_2, W_0 = \mathbb{W}=0.
\een
\end{proposition}

\begin{proofof}{Lemma~\ref{lemma::optlocal}}
Now we have by  Proposition~\ref{prop::biasfinal}, for $P_1 := E_n/n$,
\bens
\ip{\E B - R,  \hat{Z} - Z^{*} } 
& = &
\ip{W_0 - \BW, \hat{Z} - Z^{*} }  - 
\ip{I_n -  P_1, \hat{Z} - Z^{*} } {\tr(R)}/{(n-1)},
\eens
where $W_0$ is a diagonal matrix as in~\eqref{eq::defineW2}.
For the optimal solution $\hat{Z}$ of SDP1, we have 
\ben
\label{eq::BRdecomp}
\ip{W_0, \hat{Z} - Z^{*} }  & = & \ip{W_0, \diag(\hat{Z} - Z^{*}) }= 
0, \text{ where} \quad  \diag(\hat{Z}) = \diag(Z^{*}) = I_n, \\
\label{eq::W2sum}
\abs{\ip{\BW, \hat{Z} - Z^{*} }} & \le &
\norm{\BW}_{\max} \shnorm{ \hat{Z} - Z^{*} }_1  \le 
{2\abs{V_1 - V_2}} \shnorm{(Z^{*} - Z)}_1/n.
\een
Hence, by the triangle inequality and the fact that ${\tr(R)}/{n}  =  p \gamma w_1
w_2$ and (A2),
\ben
\label{eq::twocomp}
\ip{\E B - R,  \hat{Z} - Z^{*} } 
& \le & 
\abs{\ip{\BW, \hat{Z} - Z^{*} } } + \abs{\frac{\tr(R)}{n(n-1)}  \ip{E_n, \hat{Z} - Z^{*} }} \\
\nonumber 
& \le &
p \gamma \shnorm{\hat{Z} - Z^{*} }_1 \big(\xi  +  1/(4(n-1)) \big) \le  2 \xi p \gamma \shnorm{\hat{Z} - Z^{*} }_1,
\een
where ${2\abs{V_1 - V_2}}/{n} \le \xi p \gamma$.
The lemma thus holds.
\end{proofof}

\begin{remark}{\textnormal{\bf (Balanced Partitions)}}
  \label{rm::target}
Suppose that $w_1 = w_2$, then we may hope to obtain a tighter  bound
for $\ip{\E B - R,  \hat{Z} - Z^{*} }$.
We conjecture that~\eqref{eq::W2sum} can be substantially 
tightened for the balanced case, where $\BW = W_2$. Now by symmetry of matrix $\hat{Z} - Z^{*}$, 
\ben
\label{eq::target2}
\vecone_n^T( \hat{Z} -Z^{*}) u_2
& = &
\sum_{i\in \MC_1} 
\sum_{j\in\MC_1} (\hat{Z} -Z^{*})_{ij}  - \sum_{i\in \MC_2} \sum_{j \in \MC_2} (\hat{Z}
  -Z^{*})_{ij},
\een
where the two off-diagonal block-wise sums have been cancelled due to  symmetry.
Hence for $w_1 = w_2$, we have
(a) $\inv{n}{\vecone_n^TZ^* u_2} =\vecone_n^T 
u_2=0$, and (b) for $\BW = W_2$, 
\ben
\label{eq::W2pro}
\lefteqn{
  \abs{\ip{\BW, \hat{Z} - Z^{*}} }
=  ({\abs{V_1 - V_2}}/{n})
\abs{\ip{\left[
\begin{array}{cc}
E_{n_1} &  0 \\
0  & - E_{n_2} 
\end{array}
\right],  \hat{Z} - Z^{*} }}}\\
& = &
\label{eq::target0}
({\abs{V_1 - V_2}}/{n})
\abs{\vecone_n^T( \hat{Z} -Z^{*}) u_2}   \le \abs{V_1 - V_2} \shnorm{Z^{*} - \hat{Z}}_2/2,
\een
  where recall $\diag(\hat{Z}) =\diag(Z^{*}) = I_n$ and 
  \eqref{eq::target0}   holds by symmetry and (a); cf. Lemma~\ref{lemma::targetexp}.
 Moreover, since $(Z^{*}_{ij} - \hat{Z}_{ij}) \ge 0, \forall (i, j) \in
 \MC_k$, the two block-wise sums $S_k := \sum_{i,j
   \in \MC_k}( Z^{*}_{ij} - \hat{Z}_{ij}) \ge 0$, $k=1, 2$,  are both  
nonnegative in~\eqref{eq::target2}.
Hence we expect the difference between the two block-wise sums
$\abs{S_1 - S_2}$ to be small for \eqref{eq::W2pro}.
Formally, we have Conjecture~\ref{conj::balpar}.
\end{remark}
\begin{conjecture}\textnormal{(Balanced partitions with SDP1)}
  \label{conj::balpar}
  For the balanced case where $n_1  = n_2$, the bias term \eqref{eq::twocomp} essentially becomes a
  small order term compared with the variance.
\end{conjecture}

\section{Analysis of the BalancedSDP for balanced partitions: Bias free}
\label{sec::biasredo}
Suppose $w_1 = w_2 = 0.5$.
Since $\inv{n}{\vecone_n^TZ^* u_2} =\vecone_n^T 
u_2=0$, we have by symmetry of $\hat{Z}$ and \eqref{eq::target0},
\bens
\abs{\ip{\BW, \hat{Z} - Z^{*}} }
& = &
\abs{\ip{W_2, \hat{Z}}} =[{\abs{V_1 - V_2}}/{n}] \abs{u_2^T
    \hat{Z} \vecone_n}.
\eens
By Rayleigh's Principle, Theorem~\ref{thm::mineigen} holds for the Rayleigh quotient $R(x) =\frac{x^T Z x}{x^T x}$, where $x \not=0$.
 Recall the minimum of the Rayleigh quotient is the smallest eigenvalue 
$\lambda_{\min}(Z)$.
\begin{theorem}
  \label{thm::mineigen}
Let $Z \succeq 0$ and $\ip{E_n, Z} =0$. Then $R(x)$ reaches the minimum at the $\vecone_n$, that is,
$R(\vecone_n) := \min_{x \in \R^n, x \not=0} {x^T Z x}/{x^T x} = \lambda_{\min}(Z) =0.$
\end{theorem}

 \begin{lemma}
   \label{lemma::targetexp}
Suppose $w_1 = w_2 = 0.5$.
Let $\hat{Z}$ be an optimal solution to \eqref{eq::sdpball}.
Then
\ben
\label{eq::targetupdate}
\ip{\left[
\begin{array}{cc}
E_{n_1} &  0 \\
0  & - E_{n_2} 
\end{array}
\right], Z^{*} - \hat{Z}  }
=  \vecone_n^T( \hat{Z} -Z^{*}) u_2  = \vecone_n^T\hat{Z} u_2 =0.
\een
As a consequence, we have   $\ip{W_2,  \hat{Z} - Z^{*} }  =0$ and $\ip{\E B - R,  \hat{Z} - Z^{*} }  =0$.
\end{lemma}

We rewrite (Balanced SDP) \eqref{eq::sdpball} as 
\ben 
\label{eq::BalancedZ2}
\text{maximize} && \ip{YY^T -  \lambda E_n,  Z} \quad \text{subject to} \quad  Z \in \M^+_{\opt} \; \text{ for } \\
\label{eq::moptplus}
\M^+_{\opt} & = & \{Z: Z \succeq 0,  \diag(Z) = I_n, \ip{Z, E_n} =0\}, 
\; \text{where}\; E_n =\vecone_n \vecone_n^T. 
\een 
Here the trace term $\ip{\lambda E_n, Z}$ in~\eqref{eq::BalancedZ2} 
does not play any role in optimization.
This eliminates the $F_2$ and $F_3$ terms in view
of Lemma~\ref{lemma::targetexp}.
Hence, we now have the updated
Elementary Inequality, replacing Lemma~\ref{lemma::signalgrad}.
The proof then follows that of Lemma~\ref{lemma::signalgrad}, except that
now $F_2, F_3 =0$ due to the new constraint $\ip{E_n, Z} = 0$.
The rest of the proof for Theorem~\ref{thm::expratenew} follows that 
of Theorem~\ref{thm::exprate} and hence is omitted.

\begin{lemma}{\textnormal{\bf (Elementary inequality: balanced partitions)}}
  \label{lemma::signalgradupdated}
  Suppose $w_1 = w_2 = 0.5$. Clearly $Z^{*} = u_2 u_2^T \in
  \M_{\opt}^+$. By optimality of $\hat{Z} \in
  \M^+_{\opt}$, we have $\ip{\hat{Z}, I_n} =\ip{I_n, Z^{*}} = n$ and  
$\ip{\hat{Z}, E_n} =\ip{E_n, Z^{*}} = 0$, and moreover,
\ben
\label{eq::signalnew}
p \gamma \shnorm{\hat{Z} - Z^{*}}_1/4  \le \ip{R, Z^{*} - \hat{Z}}  \le  
 \ip{Y Y^T - \E(YY^T), \hat{Z} - Z^{*} }.
\een
\end{lemma}

\begin{proofof}{Lemma~\ref{lemma::targetexp}}
Denote by $\vecone_{\MC_k} \in \{0, 1\}^n$ the group indicator vector 
for $\MC_k, k=1, 2$ and clearly $\sum_{k}\vecone_{\MC_k} =  \vecone_n$.
As a result of Theorem~\ref{thm::mineigen}, we have $\hat{Z} v_n = 0 v_n$ for $v_n :=
\vecone_n/\sqrt{n}$. Thus $\lambda_{n}(\hat{Z})
:=\lambda_{\min}(\hat{Z}) = 0$, and for all other eigenvectors $v_i,
i=1, \ldots, n-1$ of $\hat{Z}$, we have $\ip{v_i,  \vecone_n} =0$.
Now split each $v_i$ into two parts according to $\MC_k, k=1, 2$ to
obtain 
\bens
\forall j = 1, \ldots, n-1, \quad
\vecone_{\MC_1}^T v_j := \sum_{\ell \in \MC_1} v_{j, \ell}
= -\sum_{\ell \in \MC_2} v_{j, \ell}  = -\vecone_{\MC_2}^T v_j.
\eens
Thus we have for $\hat{Z} = \sum_{j=1}^n \lambda_j(\hat{Z}) v_j
v_j^T$, where $\lambda_1(\hat{Z}) \ge \ldots \ge \lambda_n(\hat{Z}) = 0$, and by symmetry,
\bens
\lefteqn{\vecone_n^T\hat{Z} u_2
= 
\sum_{i\in \MC_1} \sum_{j\in\MC_1} \hat{Z}_{ij}  - \sum_{i\in \MC_2}
\sum_{j \in \MC_2} \hat{Z}_{ij}  = 
\vecone_{\MC_1}^T \hat{Z} \vecone_{\MC_1} -
\vecone_{\MC_2}^T \hat{Z} \vecone_{\MC_2} }\\
& = & 
\vecone_{\MC_1}^T (\sum_{j=1}^{n-1} \lambda_j(\hat{Z}) v_j v_j^T) \vecone_{\MC_1} -
\vecone_{\MC_2}^T (\sum_{j=1}^{n-1} \lambda_j(\hat{Z}) v_j v_j^T)
\vecone_{\MC_2}  =  0.
\eens
Finally, we have by~\eqref{eq::twocomp}, $  \ip{\E B - R,  \hat{Z} -
  Z^{*} }  =0$, where the first component $=0$ for $\BW = W_2$, which
follows from \eqref{eq::targetupdate}, while for the second component,
we have $\ip{E_n, \hat{Z}} =0$ by definition of~\eqref{eq::sdpball} and $\ip{E_n, Z^{*} } = u_2^T E_n u_2
=0$ by definition of $Z^{*}$.
\end{proofof}

\section{Conclusion}
\label{sec::conclude}
The present work aims to illuminate the geometric and 
probabilistic features for population clustering through the MAXCUT
framework.
Centering the data matrix $X$ plays a key role in the statistical 
analysis and in understanding the roles of sample size lower bounds 
(on $n, p$ and their tradeoffs) for partial recovery of the clusters 
using the local and global analyses.
With new results on concentration of measure bounds on 
$\shnorm{YY^T -\E YY^T}$, we can simultaneously analyze SDP1 as well 
as the spectral algorithm based on the leading eigenvector of $YY^T$;
cf. Theorem~\ref{thm::SVD} and  Corollary~\ref{coro::SVD},
where we obtain error bounds similar to 
those in Theorem~\ref{thm::SDPmain}.
Importantly, because of these new concentration of measure bounds, 
our theory works for the small sample and low SNR ($n < p$ and $s^2 =
o(\log n)$) cases for both SDP1 (in its local and global analyses) and 
the spectral algorithm,  which is perhaps the more surprising case.
Using the local analysis, we discover additional 
debiasing properties of SDP1 and its variation for balanced 
partitions, and hence we remove (A2) entirely from 
Theorem~\ref{thm::expratenew}.
Importantly, we show that BalancedSDP~\eqref{eq::sdpball} has
an inherent tolerance on arbitrary variance discrepancy  between the
two component distributions in case $n_1 = n_2$,
while achieving optimal clustering for balanced partitions.
Consequently, for balanced partitions, our result in Theorem~\ref{thm::expratenew}
for BalancedSDP~\eqref{eq::sdpball} is substantially stronger than those in the literature in the sense
that we now simultaneously eliminate the bounded trace-difference
assumption (A2), as well as the need for an explicit debiasing step.
Such explicit debiasing steps typically come with an additional
computational cost; see for example~\cite{Royer17}.

\section*{Acknowledgments}  
The author would like to thank Alan Frieze and Mark Rudelson for helpful
discussions on an earlier draft of the paper, and the Simons Institute for the Theory
of Computing for the generous support and hospitality.

\appendix

\section{Proof of Theorem~\ref{thm::ZHW} and Lemma~\ref{lemma::twogroup}}
\label{sec::proofofZHW}
\label{sec::opcorrelated}
In the rest of this section, we prove Theorem~\ref{thm::ZHW}.
The proof may be of  independent interests. 
First, we prove some more general results, from which  Lemma~\ref{lemma::twogroup} follows.
 We generate $\Z$ according to Definition~\ref{def::WH}:
 \bens
 \Z_j & = & H_j W_j  \in \R^{p}, \; \text{ and hence}\;
\Z  =  \sum_{j=1}^n e_j W_j^T H_j^T = \sum_{j=1}^n \diag(e_j)  \BW  H_j^T,
\eens
where $W_1^T, \ldots, W_n^T \in \R^{m}$ are independent, mean-zero,
isotropic row vectors of $\BW = (w_{jk})$, where we assume that coordinates $w_{jk}$
are also independent with $\max_{j,k} \norm{w_{jk}}_{\psi_2} \le K$.
Hence $\Z_1, \ldots, \Z_n \in \R^{p}$ are independent, mean zero, 
sub-gaussian random vectors with covariance $\cov(\Z_j) = \E (\Z_j \Z_j^T) $ for each $j
\in [n]$ with the following properties.

First,  $\forall j \in [n]$, denote by $\Sigma_j = \cov(Z_j)$, where
\bens
\cov(\Z_j) := \E (\Z_j \Z_j^T) =   H_j \E(W_j W_j^T) H_j^T = H_j H_j^T
\eens
and $\tr(\Sigma_j) =\tr(H_j H_j^T) = \fnorm{H_j}^2$.
Here by definition of~\eqref{eq::covZ2}, we have for any $v \in \R^p$,
\ben
\label{eq::L2norm}
\norm{\ip{\Z_j, v}}_{L_2}
= (v^T \E [\Z_j  \Z_j^T] v)^{1/2} = (v^T \cov(\Z_j) v)^{1/2}
= \twonorm{H_j^T v}.
\een
Now $\forall h \in \Sp^{p-1}$, we have  by definition
of~\eqref{eq::Wpsi} and~\eqref{eq::L2norm},
\ben
\nonumber
\norm{\ip{\Z_j, h}}_{\psi_2}  & = &  
\norm{\ip{H_j W_j,  h}}_{\psi_2} = \norm{\ip{W_j,  H_j^T h}}_{\psi_2}
\\
\label{eq::C0proof1}
& \le &
\norm{W_j}_{\psi_2} \twonorm{H_j^T h}  =  \norm{W_j}_{\psi_2} \norm{\ip{\Z_j, h}}_{L_2};
\een
Thus we have shown that \eqref{eq::covZ1} and \eqref{eq::covZ2} hold for $C_0$ as
in \eqref{eq::C0define} for the independent, sub-gaussian random vectors $\Z_1, \ldots, \Z_n \in \R^p$
generated according to Definition~\ref{def::WH}, since for some universal constant $C$,
\ben 
\label{eq::C0define}
K  \le \max_{j} \norm{W_j}_{\psi_2}  \le   C K \; \forall j,  \text{
  and } \; C_0  = \max_{j} \norm{W_j}_{\psi_2},
\een 
by definition of $K = \max_{j, k} \norm{w_{jk}}_{\psi_2}  $ in~\eqref{eq::Wpsi2}. 
See Definition 3.4.1 and Lemma 3.4.2~\citep{Vers26} for the proof of \eqref{eq::C0define}. 

\begin{proofof}{Theorem~\ref{thm::ZHW}}
Let $e_1, \ldots, e_n$ be the canonical basis of $\R^n$.  By Definition~\ref{def::WH},
  \ben
  \nonumber
  \mvec{\Z} & = & \sum_{j=1}^n \mvec{\diag(e_j)  \BW H_j^T }  =  \sum_{j=1}^n H_j 
\otimes \diag(e_j) \mvec{\BW} \\
  \label{eq::LW}
  & =: &  L \mvec{\BW}  \in \R^{n p} \; \; \text{ where }\  \; L  :=  \sum_{j=1}^n H_j 
\otimes \diag(e_j) \in \R^{np \times mn}.
\een
Following Definition~\ref{def::WH}, we have 
$\cov(\mvec{\Z})  = L L^T := \sum_{i=1}^n H_i H_i^T \otimes \diag(e_i).$
On the other hand, we have for $\BW \in \R^{n \times m}$, $\Z_j = H_j W_j$ and hences 
\ben
\nonumber
\Z^T & = &  [\Z_1, \ldots, \Z_n]
= \sum_{j=1}^n \Z_j \otimes e_j^T  =  \sum_{j=1}^n H_j \BW^T
\diag(e_j), \\
\nonumber
\text{ and hence}  \; 
\mvec{\Z^T} 
& =  & \mvec{\sum_{j=1}^n H_j \BW^T \diag(e_j)}  =  \sum_{j=1}^n \mvec{H_j \BW^T \diag(e_j)} \\
\label{eq::RWT}
& = &\sum_{j=1}^n (\diag(e_j) \otimes H_j) \mvec{\BW^T} 
=: R \mvec{\BW^T}.
\een
Then there exist some permutation matrices $P, Q$ such that
\ben
\label{eq::permute}
L   & =  &   \sum_{i=1}^n H_i \otimes \diag(e_i)  =
P^T \big(\sum_{i=1}^n \diag(e_i)
\otimes H_i  \big)  Q =: P^T R Q
\een
where $H_j H_j^T$ denotes the covariance matrix for each row vector
$\Z_j$, $j \in [n]$.
We now show \eqref{eq::permute} with an
explicit construction.
There exist permutation matrices $P, Q$ such
that 
\ben
\label{eq::definePQ}
\mvec{\Z^T}   & = & P \mvec{\Z} \; \text{ and } \; 
\mvec{\BW^T}  =  Q \mvec{\BW} \\
 \nonumber
\text{ and hence by}~\eqref{eq::LW}, \;
\mvec{\Z^T} & = & P \mvec{\Z} = P L \mvec{\BW}
\een
On the other hand, we have by \eqref{eq::RWT} and \eqref{eq::definePQ}
\bens
\mvec{\Z^T} & = &
R \mvec{\BW^T}= R Q \mvec{\BW}.
\eens
This shows that for $P^T = P^{-1}$, $PL = R Q \; \; \text{ and hence} \;\; L =P^T R Q$,
and   hence~\eqref{eq::permute} holds.
See Lemma 4.3.1 and Corollary 4.3.10~\cite{HJ91}.
First we rewrite the quadratic form as follows: for any matrix $A = (a_{ij}) \in \R^n$,
   \bens
  \label{eq::quadZZ}
  \abs{\sum_{i=1}^n  \sum_{j \not=i}^n \ip{\Z_{i}, \Z_{j}}  a_{ij}}
&  = &
  \abs{\sum_{i=1}^n  \sum_{j \not=i}^n a_{ij} \sum_{k=1}^p z_{ik} z_{jk}  } \\
  &  = &
  \mvec{\Z}^T \tilde A \mvec{\Z}  = 
  \mvec{\BW}^T L^T \tilde A L  \mvec{\BW},
  \eens  
  where $\tilde A \in \R^{np \times np}$ is a block-diagonal matrix with $\diag(\tilde{A}) = 0$,
  and $p$ identical blocks $\tilde{A}^k =\offd(A), \forall k \in [p]$ of size $n 
  \times n$ along the main diagonal, where $\twonorm{\offd(A)} \le
  \twonorm{A} +   \twonorm{\diag(A)} \le 2\twonorm{A}$.
We now compute $\twonorm{L}  =  \twonorm{R} =
  \twonorm{\sum_{i=1}^n \diag(e_i)  \otimes H_i  }  \le \max_{i}
  \twonorm{H_i}$, and
\bens 
\twonorm{L^T \tilde{A} L}
  & \le &  \twonorm{\sum_{i=1}^n H_i \otimes \diag(e_i)}^2 
  \twonorm{\tilde A}   \le \max_{i} \twonorm{H_i}^2  \twonorm{\tilde
    A},\;  \text{ and} \\
  \fnorm{L^T \tilde{A} L}  & \le &
  \twonorm{L}^2 \fnorm{\tilde A}  \le  \max_{i} \twonorm{H_i}^2
  \fnorm{\tilde A}
  \le   \max_{i} \twonorm{H_i}^2 \sqrt{p}   \fnorm{A},
  \eens
  where we use the property of block-diagonal matrix for $R =
  \sum_{i=1}^n \diag(e_i)  \otimes H_i$, which is also known as
  a direct sum over $H_i, i=1, \ldots, n$.
Hence for any $t >0$, by the Hanson-Wright inequality,
cf. Theorem~1.1~\cite{RV13}, for $C^2_H :=K^2  (\max_i \twonorm{H_i}^2)$,
\bens 
\lefteqn{\prob{\abs{\mvec{\Z}^T \tilde A \mvec{\Z} } > t}  =
\prob{\abs{\mvec{\BW}^T L^T \tilde{A}  L\mvec{\BW} } > t} }\\
  & \leq & 2 \exp \big(- c\min\big({t^2}/{(C_H^4  \fnorm{\tilde{A}}^2)}, 
{t}/{(C_H^2 \twonorm{\tilde{A}})}\big) \big)   \le
2 \exp \left(- c\min\left(\frac{t^2}{C_H^4  p \fnorm{A}^2}, \frac{t}{C_H^2 \twonorm{A}}\right) \right).
\eens
Finally, for all $j \in [n]$,   we have by~\eqref{eq::L2norm} and~\eqref{eq::C0proof1}, 
\ben 
\label{eq::Zpsi2}
\norm{\Z_j}_{\psi_2}
:=  \sup_{h \in \Sp^{p-1}}\norm{\ip{\Z_j,   h}}_{\psi_2}
& \le & \norm{W_j}_{\psi_2} \twonorm{H_j} \le C K \max_{j} \twonorm{H_j}. 
\een 
This shows that~\eqref{eq::Zpsi0} also holds. The theorem thus holds. 
\end{proofof}

\subsection{Proof of Lemma~\ref{lemma::twogroup}}
\label{sec::prooftwogroup}
\begin{proofof2}
    Suppose that we allow each population to have distinct 
  covariance structures and $Z_j  = H_i W_j$ for all $j \in \MC_i, \forall i \in \{1, 2\}$. 
Then $\forall j \in \MC_i, i = 1, 2$,
\bens
\cov(\Z_j) & := & \E (\Z_j \Z_j^T) =   H_i \E(W_j W_j^T) H_i^T = H_i
H_i^T
\eens
and
$V_i  :=  \tr(\Sigma_i) =\tr(H_i H_i^T) = \ip{H_i, H_i} =
\fnorm{H_i}^2$. The rest follows from \eqref{eq::C0proof1}, \eqref{eq::Zpsi2}, and
\eqref{eq::C0define}.
\end{proofof2}

\section{Preliminary results}
\label{sec::cutnormsupp}
We first present preliminary results for the global and local analyses in this section. 
These results including Theorem~\ref{thm::YYnorm} are initially proved
in~\cite{Zhou23a}.
For completeness, we first define the matrix cut norm as in~\cite{FK99}.
\begin{definition}{\textnormal{\bf (Matrix cut norm)}~\cite{FK99}}
\label{def::cutnorm}
For a matrix  $A =(a_{ij}) \in \R^{n \times n}$, we denote by $\norm{(a_{i j})}_{\infty\to1}$ its
$\ell_{\infty} \to \ell_1$ norm, which is
\bens
\norm{(a_{i j})}_{\infty\to1} & = &
\max_{\norm{s}_{\infty} \le 1} \norm{A s}_1 = 
\max_{s, t \in \left\{-1, 1\right\}^n}
\ip{A, st^T}.
\eens
This norm is equivalent to the matrix cut norm
defined as: for $A \in \R^{n \times n}$,
\bens
\norm{A}_{\square} & = & \max_{I \subset [n], J \subset [n]}
\abs{ \sum_{i    \in I} \sum_{j \in J} a_{i, j}},  \; \;\text{ and hence } \\
\infonenorm{A}
  & = & \max_{x,y \in \{-1, 1\}^n}
  \sum_{i=1}^n \sum_{j=1}^n a_{ij}  x_i y_i \le 
  \twonorm{x}\twonorm{y} \twonorm{A} \le n \twonorm{A}.
    \eens 
\end{definition}

Recall $\sigma_{\max}^2 := \max_{j} \shnorm{\cov(\Z_j)}_2$.
Let $\mu :={(\mu^{(1)}-\mu^{(2)})}/{\sqrt{p \gamma}} \in \Sp^{p-1}$ as
in~\eqref{eq::definemu}.  Lemma~\ref{lemma::anisoproj} hold for $\Z$ as in Definition~\ref{def::WH}, where $\max_{i}
\twonorm{R_i \mu}^2$ in \eqref{eq::proanis} and \eqref{eq::proanisum}
is understood to be replaced by $\max_{j \in [n]} \twonorm{H^T_j \mu}^2$.

\begin{lemma}[\bf Projection for anisotropic sub-gaussian random  vectors]
  \label{lemma::anisoproj}
  Let $X$, $C_0$ and $K$ be as in Lemma~\ref{lemma::twogroup}.
  In particular, $K \le C_0 = \max_{j} \norm{W_j}_{\psi_2} \le C K$ for an absolute 
constant $C$, by definition of $K = \max_{j, k} \norm{w_{jk}}_{\psi_2}  $ in~\eqref{eq::Wpsi2}. 
Let $Y$ be as in~\eqref{eq::defineY}.
Suppose all conditions in Lemma~\ref{lemma::projWH} hold. Then for $R_i =H_i^T$,
\ben
\label{eq::Zcovdef}      
\norm{\ip{\Z_j, \mu}}_{\psi_2}
& \le & C_0 \norm{\ip{\Z_j, \mu}}_{L_2} =  C_0 \sqrt{\mu^T \cov(\Z_j)  \mu} \\
\nonumber
\text{where } \; \sqrt{\mu^T \cov(\Z_j)  \mu}
& = &  \sqrt{\mu^T H_i H_i^T \mu} = \twonorm{R_i \mu} \; \; \text{ for each} \; j \in \MC_i, i =1, 2,
\een
where $R_i = H_i^T$.
Thus for any $t > 0$, for some absolute constants $c, c'$,
we have for each $v \in \Sp^{n-1}$ and $u =(u_1,\ldots, u_n) \in \{-1,
1\}^n$ the following tail bounds:
\ben
\label{eq::proanis}
\prob{\abs{\sum_{i=1}^n v_i \ip{\Z_i, \mu}} \ge t} & \le &
2 \exp\left(- \frac{c t^2}{(C_0 \max_{i} \twonorm{R_i \mu})^2}\right),
\; 
\text{ and} \; \\
\prob{\abs{\sum_{i =1}^{n} u_i \ip{\Z_i, \mu }} \ge t}
\label{eq::proanisum}
& \le &
2 \exp\left(-\frac{c' t^2}{n (C_0 \max_{i} \twonorm{R_i \mu})^2}\right)
\een
\end{lemma}

\begin{proof}
  First, $\Z_j = H_i W_j$ for each $j \in  \MC_i, i =1, 2$, is a
  sub-gaussian random vector with its marginal $\psi_2$ norm bounded
  in the sense of \eqref{eq::covZ1} and \eqref{eq::covZ2}; Hence for $R_i = H_i^T$, 
\ben
\label{eq::covW1}
\norm{\ip{\Z_j,  \mu}}_{\psi_2} & = & \norm{\ip{H_i W_j, \mu}}_{\psi_2} 
\le \norm{W_j}_{\psi_2} \twonorm{R_i \mu} =
\norm{W_j}_{\psi_2} \norm{\ip{\Z_j, \mu}}_{L_2},
\een
where $\norm{\ip{\Z_j, \mu}}_{L_2}^2 =  \mu^T  \cov(\Z_j)  \mu =\mu^T
H_i H_i^T  \mu = \twonorm{R_i \mu}^2$ by Lemma~\ref{lemma::twogroup}.
Hence by \eqref{eq::covW1} and definition of $K$ as in
in~\eqref{eq::Wpsi2},
Hence, we have for $i =1, 2$ and $C_0 = \max_j \norm{W_j}_{\psi_2}$,
\bens
\forall j \in \MC_i,  \quad
\norm{\ip{\Z_j, \mu}}_{\psi_2} \le \norm{W_j}_{\psi_2} \twonorm{R_i \mu} \le
C_0 \twonorm{R_i \mu} = C_0 \norm{\ip{\Z_j, \mu}}_{L_2}.
\eens
Second, we have by independence of $\Z_j, \forall j$ and for $u =(u_1,
\ldots, u_n) \in \{-1, 1\}^n$,
\bens
\norm{\sum_{i=1}^n u_i \ip{\Z_i, \mu}}^2_{\psi_2}
& \le &
C \sum_{i=1}^n \norm{\ip{\Z_i, \mu}}^2_{\psi_2}
\le   C n (C_0 \max_{i} \twonorm{R_i \mu})^2, \text{ and } \\
\norm{\sum_{i=1}^n v_i \ip{\Z_i, \mu}}^2_{\psi_2}
& \le & C \sum_{i=1}^n v_i^2 \norm{\ip{\Z_i, \mu}}^2_{\psi_2}
\le C (C_0 \max_{i} \twonorm{R_i \mu})^2, \; \forall v \in \Sp^{n-1}.
\eens
Then \eqref{eq::proanis} and  \eqref{eq::proanisum} follow from
the sub-gaussian tail bound; See, for example, Propositions 2.6.6 and
2.7.1, and Theorem 2.7.3~\citep{Vers26}.
\end{proof}

 \begin{corollary}
   \label{coro::sumY}
Let $X$, $C_0$ and $K$ be as in Lemma~\ref{lemma::twogroup}. 
Let $Y$ be as in~\eqref{eq::defineY}.
Under the conditions in Lemma~\ref{lemma::anisoproj},
we have with probability at least $1-2\exp(-cn)$, for $\mu$ as in~\eqref{eq::definemu},
 \bens
\abs{\ip{\mu, \sum_{j \in \MC_1} (Y_j - \E(Y_j))  -
    \sum_{j \in \MC_2} (Y_j - \E(Y_j))}}
\le 2\sum_{i=1}^n \abs{\ip{\mu, \Z_i}} \le C C_0 n \max_{i} \twonorm{R_i \mu}.
\eens
\end{corollary}

\begin{proof}
Now for $t = C C_0 n \max_{i} \twonorm{R_i \mu}/2$, we have by
\eqref{eq::proanisum} and the union bound,
\bens
\prob{\max_{ u \in \{-1, 1\}^n } \sum_{j=1}^n u_j \ip{\mu, \Z_j} \ge t}
& \le &  2^n 2 \cdot \exp\left(-\frac{c' t^2}{n (C_0 \max_{i} \twonorm{R_i 
      \mu})^2}\right) \le   2 \exp(- c n).
\eens
Hence, with probability $\ge 1-2\exp(-cn)$, we have for $x =
(x_1, \ldots, x_n) \in \{-1, 1\}^n$,
 \bens
 \abs{\ip{\mu, \sum_{j \in \MC_1} (Y_j - \E(Y_j))  -
    \sum_{j \in \MC_2} (Y_j - \E(Y_j))} }& \le & 
\max_{x \in \{-1, 1\}^n} \sum_{i=1}^n x_i  \ip{\mu, Y_i - \E(Y_i)}\\
& \le & 
2 \max_{u \in \{-1, 1\}^n} \sum_{i=1}^n  u_i \ip{\mu,  \Z_i} \le 2t,
\eens
by  Lemma~\ref{lemma::tiltproject}.
\end{proof}

\section{Proof of Theorem~\ref{thm::YYaniso}, and
  Lemmas~\ref{lemma::YYcovcorr} and~\ref{lemma::projWH}}
\label{sec::proofYYconc} 
Theorem~\ref{thm::YYaniso}  follows from~\eqref{eq::projection}, 
Lemmas~\ref{lemma::YYdec},~\ref{lemma::YYcovcorr},
and~\ref{lemma::projWH}  immediately, 
and the probability statements hold upon adjusting the constants.
We prove Lemmas
~\ref{lemma::YYcovcorr}
and~\ref{lemma::projWH} in the supplementary
Sections~\ref{sec::ZZHproof} and~\ref{sec::proofprojWH} respectively.
First, we state Proposition~\ref{prop::decompose}, which holds regardless 
of the weights or the number of mixture components. 
This justifies~\eqref{eq::YYP1proj} and Lemma~\ref{lemma::YYdec}.

\begin{proposition}[\bf Covariance projection]
  \label{prop::decompose}
Let $Y = X- P_1 X$, where $P_1 =\inv{n} \vecone_n \vecone_n^T$.
  Let $\Z = X - \E  X$. We first rewrite $\hat\Sigma_Y =  (Y-\E(Y))(Y-\E(Y))^T$ as follows:
\bens
\label{eq::SigmaY}
\hat\Sigma_Y 
& := &  (I-P_1) \Z \Z^T(I-P_1)   = 
M_1 -(M_2 - M_3), \quad  \text{ where } \\
\label{eq::defineM1}
M_1 & := & \hat\Sigma_X = (X-\E(X)) (X-\E(X))^T = \Z \Z^T, \\
\label{eq::defineM2}
M_2 & = &   P_1 \Z \Z^T+ \Z \Z^T P_1 \; \text { and } M_3  = P_1 \Z \Z^T P_1.
\eens
Hence $\Sigma_Y  := \E \hat\Sigma_Y :=  (I-P_1) \E (\Z \Z^T)(I-P_1)$. Moreover, we have \eqref{eq::projection}.
\end{proposition}

\subsection{Proof of Lemma~\ref{lemma::YYcovcorr} }
\label{sec::ZZHproof}
We prove the lemma with the full generality by allowing each
vector $Z_j = H_j W_j \in \R^p$ to have its own covariance $\Sigma_j = H_j H_j^T$, 
where $\Sigma_j \in \R^{p \times p}$, as in Definition~\ref{def::WH}.
That is, we consider the  general data generative process in 
Definition~\ref{def::WH}, and denote the (sample-by-sample) covariance by 
\ben 
\label{eq::gammaproof}
\Gamma:= \E \Z \Z^T 
=  \diag( [\tr(\Sigma_1), \ldots, \tr(\Sigma_n)]), \; \text{where} \; 
\Sigma_j = \cov(\Z_j) = H_j H_j^T; 
\een
The model under consideration in Lemma~\ref{lemma::twogroup} is a
special case of the general model in \eqref{eq::gammaproof}, since we let
$H_j = H_i$ for all $j \in \MC_i$,  for $i=1, 2$.

\begin{lemma}
  \label{lemma::E0correlated}
Let  $W_{j}$ be an isotropic, sub-gaussian random vector with 
independent, mean-zero coordinates as in Definition~\ref{def::WH}
where  $K := \max_{j, k} \norm{w_{jk}}_{\psi_2}$. 
Let $\{\Z_j^T, j \in [n]\}$ be row vectors of $\Z$,  where $\Z_j = H_j
W_j$ for $j \in [n]$. Then for each $j \in [n]$, we have 
\bens
\prob{\abs{\twonorm{\Z_j}^2 - \E \twonorm{\Z_j}^2} > t}
& = &
\prob{\abs{\twonorm{H_j W_j}^2 -\fnorm{H_j}^2} > t} \\
& \le &
2 \exp \left(- c\min\left(\frac{t^2}{(K^2\twonorm{H_j} \fnorm{H_j})^2}, \frac{t}{(K\twonorm{H_j})^2} \right)\right).
\eens
\end{lemma}

\begin{proof}
Fix $j$. First, we introduce the positive semidefinite matrix $M:= H_j^T H_j
  \succeq 0 \in \R^{m \times m}$.
  Now, we bound
  for each anisotropic vector $\Z_j = H_j W_j  \in \R^{p}$, where $W_j
  \in \R^m$, and 
\ben
\forall j \in [n], \quad
\twonorm{\Z_j}^2
& = &  \ip{H_j W_j, H_j W_j} = W_j^T H_j^T H_j W_j =:  W_j^T M W_j,
\; \text{ and hence} \;\\
\label{eq::Znorm}
\E \twonorm{\Z_j}^2  & = &   \E (W_j^T (H_j^T H_j) W_j)  =\tr(H_j^T
H_j) =\fnorm{H_j}^2,
\een
where $\tr(H_j H_j^T) = \tr(M) = \fnorm{H_j}^2 \le (m \wedge p ) \twonorm{H_j}^2$.
Now clearly,
\ben
\label{eq::Mfnorm}
\fnorm{M}  = \fnorm{H_j^T H_j} \le \twonorm{H_j} \fnorm{H_j}  \quad \text{ and }
\twonorm{M} & = & \twonorm{H_j}^2;
\een
Thus we have for any $t > 0$,  by the Hanson-Wright inequality,
cf. Theorem~1.1~\cite{RV13}, for $\twonorm{H_j W_j}^2 - \E \twonorm{H_j W_j}^2 = W^T_j M W_j -\tr(M)$,
\bens 
\lefteqn{\prob{\abs{\twonorm{\Z_j}^2 - \E \twonorm{\Z_j}^2} > t}
  =  \prob{\abs{W^T_j M W_j -\tr(M)} > t } }\\
  & \le &
2 \exp \left(- c\min\left(\frac{t^2}{K^4 \fnorm{M}^2},
    \frac{t}{K^2 \twonorm{M}} \right)\right).
\eens
The lemma thus holds in view of \eqref{eq::Mfnorm}.
\end{proof}


\begin{proofof}{Lemma~\ref{lemma::YYcovcorr}}
  Recall
  $\sigma_{\max}^2 := \max_{j} \shnorm{\cov(\Z_j)}_2$ and $C_0 =
  \max_{j} \norm{W_j}_{\psi_2} \ge K$   by definition of $C_0$ and
  $K = \max_{j, k} \norm{w_{jk}}_{\psi_2}$ as in~\eqref{eq::Wpsi2}.
First, we have by Lemma~\ref{lemma::E0correlated} and for $t_{\diag}
= C_{\diag} C_0^2 (\sqrt{n} \max_{j} (\twonorm{H_j} \fnorm{H_j} )) \vee (n
\sigma_{\max}^2 )$, where $\max_{j} (\twonorm{H_j} \fnorm{H_j} ) \le \sqrt{p} \sigma_{\max}^2$,
\ben
\label{eq::E0}
\lefteqn{\prob{\max_{j \in [n]} \abs{\twonorm{\Z_j}^2 - \E \twonorm{\Z_j}^2} > t_{\diag}}
  =: \prob{\E_0}  = }\\
\nonumber
& \le & \sum_{j=1}^n 2 \exp \left(-c_1 \min\left(\frac{(
      C_{\diag}^2 C_0^4 (\max_j \twonorm{H_j} \fnorm{H_j})^2) n}{K^4 \twonorm{H_j}^2
      \fnorm{H_j}^2},    \frac{(C_{\diag} C_0^2 \max_j \twonorm{H_j}^2)n}{K^2 \twonorm{H_j}^2}
  \right)\right) \\
\nonumber
& \le & 2 n \exp(- c_1 \min(C_{\diag}^2, C_{\diag}) n)  \le  2 \exp(- c' n),
\een
where for the $p \times m$ matrix $H_j$,  we have $\fnorm{H_j} \le \sqrt{p 
  \wedge m} \twonorm{H_j}$ for all $j \in [n]$.
We use Theorem~\ref{thm::ZHW} to bound the off-diagonal part.
Hence for all $q, h \in \Sp^{n-1}$, $A(q,h) := \offd(q \otimes h)
=(a_{ij})$, $\twonorm{{A}(q, h)} \le  \fnorm{A(q,h)} \le 1$.
Let $t_{\offd} = C_1 C_0^2 \sigma_{\max}^2 ( \sqrt{p n} \vee n)
\ge C_1 K^2  \sigma_{\max}^2 ( \sqrt{p n} \vee n)$.
For a particular realization of $q, h \in \Sp^{n-1}$ and ${A}(q, h)
=(a_{ij})$ as defined above,
\bens
\lefteqn{
\prob{\abs{\sum_{i=1}^n  \sum_{j \not=i}^n \ip{\Z_{i}, \Z_{j}}
    q_{i} h_{j}} > t_{\offd}}  = 
\prob{\abs{\sum_{i=1}^n  \sum_{j \not=i}^n \ip{\Z_i, \Z_j} a_{ij} } > t_{\offd}} }\\
& \le &
2 \exp \left(- c_2\min\left(\frac{C_1^2 (C_0\sigma_{\max})^4 (\sqrt{pn}
      \vee n)^2}{(K \sigma_{\max})^4  p \fnorm{A(q,h)}^2},
    \frac{C_1 (C_0 \sigma_{\max})^2 (\sqrt{p n} \vee n)}{(K \sigma_{\max})^2 \twonorm{A(q, h)}}  \right)\right)  \\
& \le &   2 \exp \left(- c_2 n\min (C_1^2, C_1)\right) \leq 2 \exp(-c_2  n)
\eens
  for some sufficiently large constants $C_1 \ge 1$ and $c_2 > 4 \ln 9$,  by Theorem~\ref{thm::ZHW}.
Taking a union bound over all $\abs{\Net}^2 \le 9^{2n}$ pairs $q, h \in \Net$, the
$\ve$-net of $\Sp^{n-1}$, we conclude that
\ben
\label{eq::E8}
\lefteqn{\prob{\max_{q, h \in \Net}
 \abs{\sum_{i=1}^n \sum_{j \not= i}^n q_i  h_j \ip{\Z_{i}, \Z_{j}}}>
 t_{\offd}} =: \prob{\E_8}} \\
\nonumber
&\le &
\abs{\Net}^2 \cdot 2 \exp (- c_2 n \min(C_1^2, C_1) )
 \le  2\exp\left(-c_2 n + 2n \ln 9\right) = 2 \exp\left(-c_3 n \right).
\een
One can show that \eqref{eq::ZZHop} holds by a standard approximation 
argument under $\E_8^c$;  See for example Exercise
4.4.3~\cite{Vers18}.
 Thus we have on event $\E_8^c \cap \E_0^c$,
 \bens
\twonorm{\Z \Z^T- \E (\Z \Z^T) } &\le &
\norm{\diag(\Z \Z^T) -\E \diag(\Z \Z^T)}_{\max} +\twonorm{\offd(\Z \Z^T)} \\
& \le &  C_2 C_0^2  \sigma_{\max}^2 (\sqrt{ np} \vee n); \;
\;  \text{cf.~\eqref{eq::E0} and \eqref{eq::E8}}.
\eens
Moreover, we have $\prob{\E_8^c \cap \E_0^c}  \ge 1- 2 \exp(-c n)$
upon adjusting the constants.
\end{proofof}

\subsection{Proof of Lemma~\ref{lemma::projWH}}
\label{sec::proofprojWH}
  Let $c, c', C_3, C_4$ be some absolute constants.
  Let $M_Y :=\E(Y)(Y-\E(Y))^T + (Y-\E(Y))\E(Y)^T$.
  Clearly, vectors $\Z_1, \Z_2, \ldots, \Z_n \in \R^{p}$ are independent. 
  Let $\mu$ be as in~\eqref{eq::definemu}. Let $R_i = H_i^T$.

Construct an $\ve$-net $\Pi_n$ of $\Sp^{n-1}$, where $\ve  = 1/3$ and
$\size{\Pi_n} \le (1+ 2/\ve)^{n}$. For a suitably chosen constant $C_3$,  we have
by Lemma~\ref{lemma::anisoproj}, 
\bens
\prob{\E_3} & := &
  \prob{\exists q \in \Pi_n,
    \abs{\sum_{i=1}^n q_i \ip{\Z_i, \mu}} \ge \half C_3  (C_0 \max_i
    \twonorm{R_i \mu}) \sqrt{n }}  \\
\label{eq::E8enso}
& \le& 9^n \cdot 2 \exp\left(- \frac{c n (C_0 \max_i \twonorm{R_i
      \mu})^2}{C (C_0 \max_i \twonorm{R_i \mu})^2}\right)
\le  2 \exp(- c' n).
\eens
By a standard approximation argument and
Lemma~\ref{lemma::tiltproject}, on event $\E_3^c$
\bens
 \sup_{q \in \Sp^{n-1}}  \sum_{i=1}^n q_i \ip{\Z_i, \mu}
  & \le &\inv{1-\ve} \sup_{q \in \Pi_n}  \sum_{i=1}^n q_i \ip{\Z_i,
    \mu}  \le  C_3 C_0 (\max_i \twonorm{R_i \mu})\sqrt{n}, \\
\text{and}  \twonorm{M_Y} /\sqrt{p \gamma} & \le &
4\sqrt{ w_1 w_2 n} \sup_{q \in \Sp^{n-1}} \abs{\sum_{i=1}^n  q_i 
  \ip{\Z_i, \mu}}
\le  2 C_3 (C_0 \max_{i} \twonorm{R_i \mu}) n. \; \scriptstyle \Box
\eens

\section{Proof of Lemmas~\ref{lemma::EBRtilt} and ~\ref{lemma::TLbounds}}

\label{sec::proofofEBR}
We have the following facts about $R$ as defined in~\eqref{eq::Rtilt}.
Next, we state the following fact about $W_2$.
Let $M_1, M_2, M_3$, $W_0$, $W_2$, $V_1, V_2...$ be the same as in 
Propositions~\ref{prop::decompose} and~\ref{prop::biasfinal}. 
\begin{proposition}
  \label{fact::Rtrace}
Recall $R =\E (Y) \E(Y)^T$.  Clearly $\vecone^T_n R \vecone_n = 0$, and hence
  \ben
  \label{eq::negTR}
\vecone^T_n\offd( R) \vecone_n & = &\vecone^T_n R 
\vecone_n- \tr(R) = - \tr(R) = -n p \gamma w_2 w_1, \;\text{ where} \\
 \nonumber
 \tr(R)/(p \gamma)
 & = & n w_2^2 w_1 +  n w_1^2 w_2=w_1 w_2 n \quad \text{ and }  \quad \twonorm{R} = \tr(R) =w_1 w_2 n p \gamma.
\een
Thus ${\tr(R)}/{n}  =  p \gamma w_1 w_2$ and for $n \ge 4$,
\ben
\label{eq::Rop}
\twonorm{\frac{\tr(R)}{n-1} \left( I_n-
    {E_n}/{n}\right)} & \le & \frac{n}{n-1}
p \gamma w_1 w_2 \le p \gamma /3,
\een
where $\twonorm{I_n-{E_n}/{n}} = 1$ since $I_n- {E_n}/{n}$ is a projection matrix.
\end{proposition}

\begin{proposition}
  \label{fact::W2}
  Denote by
\bens 
\mathbb{W} :=
W_2 - \inv{n} \tr(W_2) I_n- \frac{\vecone_n^T \offd(W_2) \vecone_n}{n(n-1)} (E_n -I_n) 
\eens
Then $\mathbb{W}$ coincides with \eqref{eq::wow2}.
Moreover, we have for $\delta_V :=(V_2 - V_1)(w_2 - w_1)$
\ben
\nonumber
\mathbb{W} 
& =&  W_2 -\frac{ \delta_V  E_n}{n} =
\frac{V_1-V_2}{n}
\left[
\begin{array}{cc}
2 w_2   E_{n_1}  &  (w_2 - w_1) E_{n_1 \times n_2} \\
(w_2 - w_1)  E_{n_1 \times n_2}   & - 2 w_1 E_{n_2} 
\end{array}
\right], \\
\label{eq::W3cutnorm}
\twonorm{\mathbb{W}}
& \le& \abs{V_1 - V_2} (w_1 \vee w_2)  \; \text{ and } \;\;
\infonenorm{\mathbb{W}} \le  n \abs{V_1 - V_2} (w_1 \vee w_2).
\een
\end{proposition}

\begin{proof}
  By definition of $W_2$ as in~\eqref{eq::defineW2}, we have
\bens
\tr(W_2) & = &
\frac{V_2-V_1}{n}(w_2 n-w_1 n)= (V_2 - V_1)(w_2 - w_1) \;  \text{ and }  \\
\vecone_n^T W_2 \vecone_n 
& = &
\frac{V_2-V_1}{n} (w^2_2 n^2 - w_1^2 n^2)=  n(V_2 - V_1)(w_2 - w_1).
\eens
Thus
\bens 
\frac{\vecone_n^T \offd(W_2) \vecone_n}
{n(n-1)} & = & \frac{\vecone_n^T W_2 \vecone_n -\tr(W_2)}{n(n-1)} =
\frac{(V_2 - V_1)(w_2 - w_1)}{n}; \text{ and hence} \\
\mathbb{W} 
& = &
W_2 - \inv{n} \tr(W_2) I_n- \frac{\vecone_n^T \offd(W_2) 
  \vecone_n}{n(n-1)} (E_n -I_n) \\
& = &
W_2 -\frac{(V_2 - V_1)(w_2 - w_1)}{n} E_n.
\eens
Hence~\eqref{eq::wow2} holds.
Moreover, by symmetry, $\twonorm{\mathbb{W}} \le \norm{\mathbb{W}}_{\infty}   \le 
\abs{V_1 - V_2} (w_1 \vee w_2).$
\end{proof}

\subsection{Some useful propositions}
Next, we compute the mean values in Proposition~\ref{eq::M123hub}, and 
we obtain an expression on $\E B-R$ in Proposition~\ref{prop::biasfinal}.
Proposition~\ref{eq::M123hub} is proved in Section~\ref{sec::biasproofs}.
Lemma~\ref{lemma::unbalancedbias} appears in Section~\ref{sec::TLbounds}.

\begin{proposition}{\textnormal{\bf (Covariance projection: two groups)}}
  \label{eq::M123hub}
    Let $N_j = w_j n$ for $j \in \{1, 2\}$.
    W.l.o.g.,  suppose that the first $n_1$ rows in $X$ are in $\MC_1$
    and the following $n_2$ rows are in $\MC_2$.
    Let $M_1, M_2, M_3$ be defined as in Proposition~\ref{prop::decompose}.
  Let $V_1$ and $V_2$ be the same as in~\eqref{eq::Varprofilepreview}:
    \ben 
  \label{eq::defineV1}
  V_1 := \E \ip{\Z_j, \Z_j} \; \;  \forall j \in \MC_1  \;\;   \text{ and } \; \;  
 V_2 := \E \ip{\Z_j, \Z_j} \; \;  \forall j \in \MC_2.
  \een
  Let  $V_m :=  w_1 V_1 + w_2 V_2 =W_n/n$.
Let  $\hat\Sigma_Y$ be as in \eqref{eq::SigmaY}.
  Let  $W_n$ be defined as in~\eqref{eq::defineWn}:
  \ben
 \label{eq::defineWn}
  W_n & := &
     \E \sum_{i=1}^n \ip{\Z_i, \Z_i} = \sum_{i=1}^n \sum_{k=1}^p \E
     z_{ik}^2, \\
    \label{eq::EM1unb}
    \E M_1 & :=&  
\left[\begin{array}{cc} V_1 I_{n_1} & 0 \\
        0 & V_2 I_{n_2}
      \end{array}
    \right]    =: (w_1 V_1 + w_2 V_2) I_n + W_0, \\
    \label{eq::EM3unb}
 \E M_2 & =&  \frac{V_1 + V_2}{n} E_n + W_2, \; \text{and}\; \;
 \E M_3 = \frac{W_n}{n^2} E_n,\\
    \text{ where } \; \; 
W_n/n^2 & := & (\abs{\MC_1} V_1 +  \abs{\MC_2} V_2)/n^2 = (w_1 V_1 +
w_2 V_2)/n =: \frac{V_w}{n},
\een
$\tr (\E M_1) /n = V_m$, and $W_0, W_2$ are as defined in~\eqref{eq::defineW2}.
Now putting things together, we obtain the expression for covariance
of $Y$: 
\bens
\lefteqn{\Sigma_Y := \E (YY^T) - \E(Y) \E(Y)^T
 = 
\E M_1 + \E M_3- \E M_2 } \\
& = &
\left[\begin{array}{cc}
        V_1 I_{n_1} & 0 \\
0 &  V_2 I_{n_2} 
\end{array}
\right]  - \frac{w_2 V_1 + w_1 V_2 }{n} E_n - W_2 \; 
    \eens
    which simplifies to $\Sigma_Y =    V (I_n - E_n/n) \; \text{ in
      case } \; \; V_1  = V_2 = V$.
      \end{proposition}

\subsection{Proof of Lemma~\ref{lemma::EBRtilt}}
\label{sec::biasLemmaproof}
\begin{proofof2}

We have by  Proposition~\ref{prop::biasfinal},
\ben
\nonumber                                   
\norm{\E B - R}
  & =&  \norm{W_0 - \mathbb{W} -\frac{\tr(R)}{(n-1)} (I_n - \frac{E_n}{n})}
  \\
 \label{eq::normEBR}
& \le &
  \norm{W_0}                  
 +\norm{ \mathbb{W}} +\norm{\frac{\tr(R)}{(n-1)} (I_n - \frac{E_n}{n}) },
 \een
 where the $\norm{\cdot}$ is understood to be either the operator or
 the cut norm.
Clearly,  we have by 
Proposition~\ref{fact::W2},
\ben
\label{eq::W0opnorm}
\twonorm{\mathbb{W}}
\le  \abs{V_1 - V_2} (w_1 \vee w_2) \; \text{ and } \;
\twonorm{W_0}  \le \abs{V_1 -  V_2}  (w_2 \vee w_1).
\een
Combining~\eqref{eq::Rop}, \eqref{eq::normEBR}, and~\eqref{eq::W0opnorm},
we have for
$w_{\min} := w_1 \wedge w_2$ and $ n  \xi \ge \inv{2 w_{\min}}$,
\bens
\twonorm{\E B - R} & \le  &
\twonorm{W_0} + \twonorm{\mathbb{W}} + 
\twonorm{\frac{\tr(R)}{(n-1)} (I_n - \frac{E_n}{n})} \\
& \le &
\frac{2}{3} \xi n p \gamma (1- w_{\min}) + p \gamma/3 
\le \frac{2}{3} \xi n p \gamma,
\eens
where we use the fact that
$\frac{2}{3} \xi p \gamma n w_{\min} \ge  p \gamma/3$ since  $2 \xi n w_{\min} \ge  1$.
Now the bound on the cut norm follows since 
$\infonenorm{\E B - R}  \le  n \twonorm{\E B - R}   \le \frac{2}{3} \xi n^2 p \gamma$.
The lemma thus holds for the general setting; when $V_1 = V_2$, we
leave this as an exercise.
\end{proofof2}

\subsection{Proof of Lemma~\ref{lemma::TLbounds}}
\label{sec::TLbounds}

\begin{proofof}{Lemma~\ref{lemma::TLbounds}}
Now \eqref{eq::lambdabound} holds since by Proposition~\ref{fact::fullbasis},
\bens
2 {n \choose 2} \abs{\lambda - \E \lambda}
& = & \abs{n (\tau - \E \tau) } \le n \twonorm{YY^T - \E (Y Y^T)}, \\
\text{where}  \; n \abs{(\tau - \E \tau)}
& \le & n \max_{i}  \abs{\ip{Y_i, Y_i} -  \E \ip{Y_i, Y_i} } \le n \twonorm{YY^T - \E (Y Y^T)}.
\label{eq::tau2}
\eens
\end{proofof}

Corollary~\ref{coro::W2bound} follows from the proof of
Lemma~\ref{lemma::EBRtilt}, which we state to prove a bound for the
balanced cases.
The proof is given in Section~\ref{sec::W2coro}.
\begin{corollary}
 \label{coro::W2bound}
For general cases, we have by definition,
\bens
  W_2 & := &
\frac{V_1-V_2}{n}
\left[
\begin{array}{cc}
E_{n_1} &  0 E_{n_1 \times n_2} \\
0  E_{n_1 \times n_2}   & - E_{n_2} 
\end{array}
\right]
\eens
Moreover we have the following term which depends on the weights,
\bens
    \infonenorm{\frac{ (V_2 - V_1)(w_2 - w_1)}{n} E_n} & \le  &
    n \abs{(V_2 - V_1)(w_2 - w_1) } \le 
    \xi n^2 p \gamma \abs{w_2 - w_1}  \\
    \infonenorm{W_2} & := & n \abs{V_1 - V_2} (w_1^2 + w_2^2) \\
  \text{ and hence } \; \;
  \infonenorm{W_0 -W_2}&  \le & n \twonorm{W_0-W_2}
  <  n \abs{V_1 -  V_2} \le  \xi p n^2 \gamma
\eens
\end{corollary}

\begin{lemma}{\textnormal{\bf (Reductions)}}
\label{lemma::unbalancedbias}
Suppose $n \ge 4$.   Let $W_0, W_2, V_1, V_2$ be the same as in
Proposition~\ref{prop::biasfinal}.
 Recall that $R =\E(Y) \E(Y)^T$.
\noindent{\bf When $V_1 = V_2$}, we have
\bens
\E B - R  & = & - \frac{\tr(R)}{(n-1)} (I_n -    \frac{E_n}{n}) =
- p \gamma w_2 w_1 \frac{n}{n-1}(I_n -    \frac{E_n}{n}),\\
\infonenorm{ \E B - R} & = &
\infonenorm{p \gamma w_2 w_1 \frac{n}{n-1}(I_n - \frac{E_n}{n})} \le
n p \gamma/3, \; \text{ for } \; n \ge 4.
\eens
\noindent{\bf For balanced clusters ($w_1 = w_2$)},
$\E B - R  =  W_0 - W_2 -\frac{\tr(R)}{(n-1)} (I_n - \frac{E_n}{n})$ and for $n \ge 4$,
\bens
\infonenorm{ \E B - R}
& \le & n \twonorm{ \E B - R} \le 
\frac{n p \gamma}{3} + \abs{V_1 - V_2} n \le \inv{3} (1 + o(1)) \xi p n^2 \gamma.
\eens
\end{lemma}

\begin{proof}
Recall
\bens
W_0 -W_2 
& =&
(V_1 - V_2) \left[\begin{array}{cc} w_2 I_{n_1} -   E_{n_1} /n & 0 \\
 0 & - (w_1 I_{n_2}  - E_{n_2} /n) \end{array}    \right]
\eens
The case where $V_1 = V_2$ follows from  Proposition~\ref{fact::Rtrace} and~\eqref{eq::wow}.
We now show the balanced case where $w_1 = w_2$.
Under the conditions of Lemma~\ref{lemma::EBRtilt}, we have  for $\xi
= \Omega(1/n)$, by \eqref{eq::Rop} and~\eqref{eq::wow},
\bens 
\twonorm{\E B - R} & \le  &
\twonorm{W_0-W_2} + \twonorm{\frac{(V_1 - V_2)(w_1 - w_2)}{n} E_n} +
\twonorm{\frac{\tr(R)}{n-1} \left( I_n-
    {E_n}/{n}\right)} \\
& \le  &
\abs{V_1 - V_2} + p \gamma/3 \le \inv{3} \xi n p \gamma + p \gamma/3 
\eens
Similarly, the upper bound on $\infonenorm{ \E B - R}$ follows from Corollary~\ref{coro::W2bound} immediately.
\end{proof}

\subsection{Proof of Proposition~\ref{eq::M123hub}}
\label{sec::biasproofs}
\begin{proofof2}
  
  Denote by  $\tr (\E M_1) /n = (w_1 V_1 + w_2 V_2) = V_m$.
\bens
\nonumber
\E M_1 & =& 
\E  \left((X-\E(X)) (X-\E(X))^T \right)  = \E \Z \Z^T \\
& =& 
\left[\begin{array}{cc} V_1 I_{w_1 n} & 0 \\
        0 & V_2 I_{w_2 n}
      \end{array}    \right] 
\eens
Moreover, upon subtracting the component of $V_w I_n = \inv{n}\tr (\E M_1) 
I_n$ from $\E M_1$, we have $W_0$:
\bens
\E M_1   - V_w I_n  & := & \E M_1   - \inv{n}\tr (\E M_1)  I_n
=
    (V_1 -V_2) \left[\begin{array}{cc} w_2 I_{w_1 n} & 0 \\
        0 & - w_1 I_{w_2 n} 
            \end{array}    \right]        
          =:   W_0;
\eens
Next we evaluate $\E M_2$: for $\Z = X -\E X$
\bens
\nonumber
\E M_2 & = & \E \left(\left(
    \Z (\vecone{(X)} - \E \vecone{(X)} )^T +
    (\vecone{(X)}- \E \vecone{(X)})\Z^T\right)\right) \\
& = &
\label{eq::EM2unb}
\frac{1}{n}
\left[
\begin{array}{cc}
2 V_1 E_{n_1} &
  (V_1 + V_2) E_{n_1 \times n_2} \\
  (V_1 + V_2) E_{n_1 \times n_2}  &
2 V_2 E_{n_2}
\end{array}
\right]  \\
\text{and hence} \quad
\E M_2 - \frac{ (V_1 + V_2)}{n} E_n
& = &
\nonumber
\frac{V_1 - V_2}{n}
\left[
\begin{array}{cc}
E_{n_1} &
 0 E_{n_1 \times n_2} \\
 0 E_{n_1 \times n_2}  &
- E_{n_2}
\end{array}
\right]   =: W_2
\eens
For $W_n$ as defined in  \eqref{eq::defineWn}, we have
\bens
\E \ip{\hat{\mu}_n - \E \hat{\mu}_n, \hat{\mu}_n - \E
  \hat{\mu}_n} &:= & \inv{n^2}\sum_{i=1}^n \sum_k \E z_{ik}^2 =
\frac{w_1V_1 +  w_2 V_2}{n} = \frac{V_m}{n} \\
\E M_3 &:= & \E \ip{\hat{\mu}_n - \E \hat{\mu}_n, \hat{\mu}_n - \E
  \hat{\mu}_n} E_n = \frac{\abs{\MC_1} V_1 +  \abs{\MC_2} V_2}{n^2} E_n
= \frac{W_n}{n^2} E_n
\eens
Now putting things together,
\bens
\lefteqn{\E (YY^T) - \E(Y) \E(Y)^T
 = 
\E M_1 + \E M_3- \E M_2 } \\
& = &
\left[\begin{array}{cc}
        V_1 I_{n_1} & 0 \\
0 &  V_2 I_{n_2} 
\end{array}
\right] + \frac{V_m}{n} E_n - \frac{ (V_1 + V_2)}{n} E_n - W_2 \; \; \; \text{ where} \\
\frac{V_m}{n}  - \frac{ (V_1 + V_2)}{n} & = &
-\frac{V_1(1 -w_1) + V_2(1-w_2)}{n} =
-\frac{V_1 w_2+ V_2 w_1}{n} 
\eens
The proposition thus holds.
\end{proofof2}

\subsection{Proof of Proposition~\ref{prop::biasfinal}}
\label{sec::biasproofEBR}
\begin{proofof2}
First, we have by Proposition~\ref{prop::decompose}, and $R= \E(Y) \E(Y)^T$, 
\ben 
\label{eq::YYrelations3}
 \E (YY^T)  = \Sigma_Y  + R =  \E M_1  - \E M_2  + \E M_3 + R 
\een 
We have by Proposition~\ref{fact::Rtrace} and~\eqref{eq::YYrelations3}, 
\ben 
\nonumber 
\E \tau & = &
\inv{n}\sum_{i=1}^n \E \ip{Y_i, Y_i}
= \E \tr(YY^T)/n =\tr({\Sigma}_Y)/n + \tr(R)/n\\
\label{eq::Etau}
& = &
\inv{n} \left(\tr(\E M_1 + \E M_3)- \tr(\E M_2) \right)+ \tr(R)/n  \\
\nonumber 
\E \lambda & = & \inv{n(n-1)}\sum_{i\not=j}^n \E \ip{Y_i, Y_j} =
 \inv{n(n-1)} \vecone_n^T \E(\offd(YY^T)) \vecone_n \\
& = &
\label{eq::Elamb}
\inv{n(n-1)} \vecone_n^T \offd(\E M_3  -\E M_2) \vecone_n -
\frac{\tr(R)}{n(n-1)} 
\een
where in \eqref{eq::Elamb} we use the fact 
that $\offd(\E M_1) =0$ by~\eqref{eq::EM1unb} and \eqref{eq::negTR}.
Hence by definition of $B$ and $R$, we have
\ben
\E B - R & =  &
\nonumber
\E M_1 - \E M_2 + \E M_3 - \E \tau I_n  -\E \lambda 
(E_n - I_n) \\
\nonumber
& =  &
\E M_1 - \E M_2 + \E M_3 - \left(\inv{n}\tr(\E M_1 + \E M_3) -
  \inv{n} \tr(\E M_2) +  \frac{\tr(R)}{n}\right)  I_n \\
\label{eq::EBRfinal}
&  & -\left(\inv{n(n-1)} \vecone_n^T \offd(\E M_1 + \E M_3  -\E M_2)
  \vecone_n - \frac{\tr(R)}{n(n-1)}\right)(E_n -I_n)
\een
  \bit
  \item
Notice that $\E M_3 = \frac{V_w}{n} E_n$ and hence its contribution
to $\E \tau$ and $\E \lambda$ is the same;
Thus we have
\ben
\nonumber
\tr(\E M_3) & =  & \inv{(n-1)} \vecone_n^T \offd(\E M_3) 
  \vecone_n =\frac{W_n}{n} = V_m \; \text{   and
    by~\eqref{eq::EM3unb}}, \\
  \label{eq::EM30}
&& \E M_3 - \inv{n} \tr(\E M_3) I_n 
- \frac{\vecone_n^T \offd(\E 
  M_3) \vecone_n}{n(n-1)}(E_n - I_n) = 0; 
\een
\item
$\E M_1$ is a diagonal matrix and hence $\offd(\E M_1) =0$.
Now we have by~\eqref{eq::EM1unb},
\ben
\label{eq::M1fix}
&& \E M_1 -  \inv{n} \tr(\E M_1) I_n = W_0 \text{ and } \offd(\E M_1) 
= 0 
\een
\item
For $\E M_2$, we decompose it into one component proportional to 
$E_n$: $\bar{M}_2 :=\frac{V_1 + V_2}{n} E_n$ and another component 
$W_2 = \E M_2 -\bar{M}_2$. 
By Proposition~\ref{eq::M123hub}, we have 
\ben
\nonumber
W_2 &= & \E M_2 -\bar{M}_2 =\E M_2 - \frac{V_1 + V_2}{n} E_n \\
 & := &
\frac{V_1-V_2}{n}
\left[
\begin{array}{cc}
E_{n_1} &  0 E_{n_1 \times n_2} \\
0  E_{n_1 \times n_2}   & - E_{n_2} 
\end{array}
\right] \\
\label{eq::BarM2}
\text{ where by definition } &&
\bar{M}_2  - \frac{\tr(\bar{M}_2)}{n} I_n - \frac{1_n^T 
  \offd(\bar{M}_2) 1_n}{n(n-1)} (E_n- I_n) =0
\een
\eit
Thus we have by Proposition~\ref{fact::W2} and \eqref{eq::BarM2}
\ben
\nonumber
\lefteqn{
  \E M_2 - \inv{n} \tr(\E M_2) I_n- \frac{\vecone_n^T \offd(\E 
  M_2) \vecone_n}{n(n-1)} (E_n -I_n) }\\
& = &
\label{eq::M2red}
W_2 - \inv{n} \tr(W_2) I_n-
\frac{\vecone_n^T \offd(W_2) \vecone_n}{n(n-1)} (E_n -I_n)  =:
\mathbb{W}
\een
Now by \eqref{eq::Etau},~\eqref{eq::Elamb}, \eqref{eq::EBRfinal},~\eqref{eq::EM30},~\eqref{eq::M1fix},~\eqref{eq::M2red}, and Proposition~\ref{eq::M123hub}, 
\ben
\nonumber
\E B - R & =  &
\E M_1 - \E M_2 + \E M_3 - \E \tau I_n  -\E \lambda 
(E_n - I_n) \\
\nonumber
\label{eq::prelude}
& =:  &
W_0 - \mathbb{W} -  \frac{\tr(R)}{n-1} (I_n -E_n/n)
\een
where in step 2,
we simplify all terms involving $\E M_1$ and $\E M_2$, and 
eliminate all terms involving $\E M_3$.
\end{proofof2}

\subsection{Proof of Corollary~\ref{coro::W2bound} }
\label{sec::W2coro}
\begin{proofof2}
Now
\bens
W_0 - W_2   & =&
(V_1 -V_2) \left[\begin{array}{cc} w_2 I_{n_1} -E_{n_1}/n& 0 \\
        0 & -(w_1 I_{n_2} -E_{n_2} /n) \end{array} \right].
    \eens
Moreover, due to symmetry, for $w_j > 1/n$,
\bens
\twonorm{W_0-W_2}
& \le &
\norm{W_0-W_2}_{\infty} :=\max_{i} \sum_{j=1}^n \abs{W_{0,ij}
  -W_{2,ij}} \\
& \le &  \abs{V_1 - V_2}
((w_2 -1/n) + (w_1 n-1)/n) \vee( (w_1 -1/n)+ (w_2 n-1) /n)
< \abs{V_1 - V_2}
\eens
where 
\bens 
\lefteqn{((w_2 -1/n) + (w_1 n-1)/n) \vee ( (w_1 -1/n)+ (w_2 n-1) 
  /n)}\\
&  =&
((w_2 -1/n) + (w_1-1/n) \vee( (w_1 -1/n)+ (w_2 -1/n) = 1 -2/n 
\eens
Thus we have by the triangle inequality,
\bens
\infonenorm{W_0 -W_2}
&  \le&
\infonenorm{W_2} + \infonenorm{W_0} \\
& \le & \abs{V_1 - V_2}n  \big(\abs{w_2 w_1 + w_1 w_2} + (w_1^2 +
w_2^2)\big)\\
&  = & \abs{V_1 - V_2}n  \le \half \xi n^2 p \gamma 
\eens
\end{proofof2}

\subsection{Proof of Proposition~\ref{prop::optsol}}
\label{sec::oraclesupp}
The optimization goals of SDP1 
and SDP~\eqref{eq::sdpmain} are equivalent to 
that of {\bf Oracle SDP} \eqref{eq::sdpB} in view of 
Proposition~\ref{prop::optsol}.
In other words, optimizing the original SDP~\eqref{eq::sdpmain} using matrix $A$~\eqref{eq::defineAintro} 
over the larger constraint set $\M^{+}_{G}$ is equivalent to
maximizing $\ip{B, Z}$ over $Z \in \M_{\opt}$~\eqref{eq::moptintro},
using the oracle $B = A - \E \tau I_n$ as in~\eqref{eq::defineBintro}.
Denote by
\ben
\label{eq::MGplus}
\M^{+}_{G} := \{Z: Z \succeq 0,  I_n \succeq \diag(Z)\} \subset [-1, 
1]^{n \times n}.
\een

\begin{proofof}{Proposition~\ref{prop::optsol}}
Recall for $\M^{+}_{G}$ as in~\eqref{eq::MGplus},
\bens
\M_{\opt} & := & \left\{Z :  Z \succeq 0, \diag(Z) = I_n\right\}
\subset \M^{+}_{G} \subset [-1, 1]^{n \times n}; 
\eens
Notice that for the second term in \eqref{eq::defineAintro}, we have
$$\ip{(E_n -I_n), Z} = \ip{(E_n -I_n), \offd(Z)}$$
in the objective
function~\eqref{eq::sdpmain}, which does not depend on $\diag(Z)$;
Hence, to maximize
\bens
\ip{A, Z}
& = & \ip{{A}, \offd(Z)} + \ip{A, \diag(Z)} \\
& = & \ip{\offd(A), \offd(Z)} + \ip{\diag(YY^T), \diag(Z)},
\eens
over all $Z \in \M^{+}_{G}$,
one must set the diagonal $Z_{jj} \in [0, 1]$ to be 1, since
(a) $\diag(YY^T) \ge 0$ and moreover, (b) increasing $\diag(Z)$ will only make it
easier to satisfy $Z \succeq 0$ and hence to maximize $\ip{\offd(A),
  \offd(Z)}$.
Thus, the set of optimizers $\hat{Z}$ as in~\eqref{eq::sdpmain}
must satisfy $\diag(\hat{Z})  = I_n$. Thus \eqref{eq::Aquiv} holds by
definition of~$\M_{\opt}$ as above.
Moreover,~\eqref{eq::optsolAB} holds due to the
fact that $\ip{I_n, Z} = \tr(Z) = n$ for all $Z$ in the feasible set $\M_{\opt}$.
\end{proofof}

\section{Proof outline for Lemma~\ref{lemma::S1}}
\label{sec::proofofS1main}
Let $\hat\delta_Z :=\hat{Z} -Z^{*}$ and $S_1 := \ip{\cp(\Lambda), \hat\delta_Z}$.
We need to first present Lemmas~\ref{lemma::ensemble}
and~\ref{lemma::P2cov}, which are proved in Sections~\ref{sec::proofens} and~\ref{sec::proofP2cov} respectively. 
Throughout this section, we consider $\hat{Z} \in \M_{\opt} \cap 
(Z^{*} + r_1 B_1^{n \times   n})$, where $r_1 \le 2 qn$ for some positive 
integer $q \in [n]$.

Let $\sigma_{\max} :=\max_{i} \twonorm{H_i}$.
As predicted, the decomposition and reduction ideas for the global 
analysis in Section~\ref{sec::reduction} are useful for the local
analysis as well. Let 
\ben 
\label{eq::covproj}
\hat\Sigma_Y =  (Y-\E(Y))(Y-\E(Y))^T, \quad \hat{\Sigma}_Y - \Sigma_Y  =  (I-P_1) (\Z \Z^T -\E (\Z \Z^T))(I-P_1). 
\een
Then we have  by the triangle inequality,
\bens
S_1
\le    \abs{\ip{\cp((Y -\E(Y))
    \E(Y)^T), \hat\delta_Z }} + 
\abs{\ip{\cp(\E Y (Y - \E(Y))^T), \hat\delta_Z }} + \abs{\ip{\cp(\hat\Sigma_Y- \Sigma_Y), \hat\delta_Z }}.
\eens
\begin{lemma}
  \label{lemma::ensemble}
  Suppose all conditions in Theorem~\ref{thm::YYaniso} hold.
  Then   $\abs{\ip{\cp(\E(Y)(Y -\E(Y))^T), \hat{Z} -Z^{*}}}
  =\abs{\ip{\cp((Y -\E(Y)) \E(Y)^T), \hat{Z} -Z^{*}}}$, and with probability at least $1-\frac{c}{n^2}$,
    \bens
\sup_{\hat{Z} \in \M_{\opt} \cap (Z^{*} + r_1 B_1^{n \times     n})}
   \abs{\ip{\cp((Y -\E(Y)) \E(Y)^T), \hat{Z} -Z^{*}}}  \le 
 C_5 C_0 (\max_{i} \twonorm{H^T_i \mu}) \sqrt{p \gamma} r_1 \sqrt{\log (2e n/\lceil \frac{r_1}{2n}\rceil)}. 
\eens 
\end{lemma}

\begin{lemma}
  \label{lemma::P2cov}
  Under the conditions in Theorem~\ref{thm::YYaniso},
  we have with probability at least $1-\frac{c'}{n^2}$,
\bens
\lefteqn{
\sup_{\hat{Z} \in \M_{\opt} \cap (Z^{*} + r_1 B_1^{n \times  n})}
\abs{\ip{\cp(\hat\Sigma_Y- \Sigma_Y), \hat{Z} -Z^{*}} }}\\
& \le &
\frac{2}{3} \xi  p \gamma  \onenorm{\hat{Z} -Z^{*}} 
+ C_{10} (C_0 \sigma_{\max})^2
\lceil \frac{r_1}{2n}\rceil 
\left(\sqrt{n p   \log (2e n/\lceil  \frac{r_1}{2n}\rceil)} +
  \sqrt{r_1} \log(2e n/\lceil \frac{r_1}{2n}\rceil ) \right).
\eens
\end{lemma}

\begin{proofof}{Lemma~\ref{lemma::S1}}
Combining Lemmas~\ref{lemma::ensemble} and~\ref{lemma::P2cov},
we have with probability at least $1-\frac{c}{n^2}$, for $\Delta^2 =p \gamma$
and $\max_{i} \twonorm{H^T_i \mu} \le \max_{i} \twonorm{H_i} = \sigma_{\max}$,
\bens
\lefteqn{
\sup_{\hat{Z} \in \M_{\opt} \cap (Z^{*} + r_1 B_1^{n \times  n})}
S_1(\hat{Z}) \le 
\sup_{\hat{Z} \in \M_{\opt} \cap (Z^{*} + r_1 B_1^{n \times  n})}
\abs{\ip{\cp(\hat\Sigma_Y- \Sigma_Y), \hat{Z} -Z^{*}} } }\\
&& +
\sup_{\hat{Z} \in \M_{\opt} \cap (Z^{*} + r_1 B_1^{n \times  n})}
2 \abs{\ip{\cp((Y -\E(Y)) \E(Y)^T), \hat{Z} -Z^{*}} }\\
& \le & \frac{2}{3} \xi  p \gamma  \onenorm{\hat{Z} -Z^{*}} + 2 C_5
C_0(\max_{i} \twonorm{H^T_i \mu}) \lceil \frac{r_1}{2n}\rceil
\sqrt{\log (2e n/\lceil   \frac{r_1}{2n}\rceil)} n \Delta + \\ &&
C_{10} (C_0 \sigma_{\max})
\lceil \frac{r_1}{2n}\rceil
\sqrt{  \log (2e n/\lceil \frac{r_1}{2n}\rceil)} \left(C_0
  \sigma_{\max}\sqrt{r_1} \sqrt{\log(2e n/ \lceil
    \frac{r_1}{2n}\rceil)} + C_0 \sigma_{\max} \sqrt{n p } 
\right) \\
& \le &
\frac{2}{3} \xi  p \gamma  \onenorm{\hat{Z} -Z^{*}} 
+ C' C_0 \sigma_{\max}
\lceil \frac{r_1}{2n}\rceil \sqrt{\log (2e n/\lceil \frac{r_1}{2n}\rceil)}
\left( n \Delta + C_0 \sigma_{\max} \sqrt{np} \right),
\eens
where in the last  inequality, we  have by~\eqref{eq::NKlower}, for $r_1 \le 2qn$ and $q = \lceil \frac{r_1}{2n}\rceil$,
\bens
C_0 \sigma_{\max}
\sqrt{r_1} \sqrt{\log(2e n/ \lceil \frac{r_1}{2n}\rceil) } & \le &
C_0 \sigma_{\max}
\sqrt{2 n q \log (2e n/q)} \\
& \le &
C_0 \sigma_{\max}
n \sqrt{2 \log (2e)} =O(n \sqrt{p \gamma}),
\eens 
where the first inequality holds since $\max_{q \in [n]}
q \log (2e n/q)  = n \log(2e)$, given that $q \log (2e n/q)$  is
a monotonically increasing function of $q$,  for $1\le q < n$,
and the second inequality holds since $p\gamma = \Omega((C_0 \max_{i}
\twonorm{H_i})^2)$.  Lemma~\ref{lemma::S1} thus holds.
\end{proofof}

\subsection{Proof of Lemma~\ref{lemma::ensemble}}
\label{sec::proofens}
\begin{proofof2}
First, we have
\ben
\label{eq::rotation2}
\ip{(Y -\E(Y)) \E(Y)^T P_2,  \hat{Z} -Z^{*}}
&  = &
\ip{P_2 \E(Y) (Y -\E(Y))^T, \hat{Z} -Z^{*}}, \\
\label{eq::rotation}
\ip{P_2 (Y -\E(Y)) \E(Y)^T,  \hat{Z} -Z^{*}}
&  = &
\ip{\E(Y) (Y -\E(Y))^T P_2, \hat{Z} -Z^{*}}.
\een
We will bound the two terms $T_1, T_2$ in Lemmas~\ref{lemma::P2minor}
and~\ref{lemma::P2final} respectively.
\begin{lemma}
  \label{lemma::P2minor}
 Under the conditions in Theorem~\ref{thm::YYaniso},
 we have for all $\hat{Z} \in \M_{\opt}$, with probability at least $1-2\exp(-cn)$,
\bens 
 \abs{T_1} := \abs{\ip{(I-P_2) \E(Y) (Y -\E(Y))^T P_2, \hat{Z} -Z^{*}}}
\le C_4 C_0 (\max_{i} \twonorm{H^T_i \mu}) \sqrt{p \gamma} \onenorm{\hat{Z} - Z^*}. 
 \eens 
 Moreover, when $w_1 = w_2$, $T_1 =0$. 
\end{lemma} 


\begin{lemma}
 \label{lemma::P2final}
 Suppose all conditions in Theorem~\ref{thm::YYaniso} hold.
 Suppose $r_1 \le 2 qn$   for some positive integer $1\le q < n$.
 Let $\Delta = \sqrt{p \gamma}$. 
 Then, with probability at least $1-\frac{c}{n^2}$,
 \bens 
\sup_{\hat{Z} \in \M_{\opt} \cap (Z^{*} + r_1 B_1^{n \times n})}
   \ip{P_2 \E(Y) (Y -\E(Y))^T, \hat{Z} -Z^{*}}
\le C_3 C_0 (\max_{i} \twonorm{H^T_i \mu}) \Delta r_1 \sqrt{\log (2e  n/ \left\lceil \frac{r_1}{2n}\right\rceil)}.
  \eens
\end{lemma}
Then we have  by the triangle inequality,~\eqref{eq::rotation} and~\eqref{eq::rotation2},
  \ben
  \label{eq::rotation3}
  \lefteqn{
\abs{\ip{\cp((Y -\E(Y)) \E(Y)^T), \hat{Z} -Z^{*}} } =
\abs{\ip{\cp(\E(Y)(Y -\E(Y))^T), \hat{Z} -Z^{*}} } }\\
&  \le &
\nonumber
\abs{ \ip{(\E(Y) (Y -\E(Y))^T)P_2, \hat{Z} -Z^{*}}
  - \ip{P_2 \E(Y) (Y -\E(Y))^T P_2, \hat{Z} -Z^{*}}} \\
&&
\nonumber
+ \abs{\ip{P_2 \E(Y) (Y -\E(Y))^T, \hat{Z} -Z^{*}} } =: \abs{T_1}+ \abs{T_2}.
\een
Putting things together, we have 
an uniform upper bound on $\ip{\cp(\E(Y)(Y -\E(Y))^T), \hat{Z} -Z^{*}}$. 
We prove Lemma~\ref{lemma::P2minor} in 
Section~\ref{sec::P2minorproof}.
We prove Lemma~\ref{lemma::P2final} in 
Section~\ref{sec::P2finalproof}. 
\end{proofof2}

\subsection{Proof of Lemma~\ref{lemma::P2cov}}
\label{sec::proofP2cov}
Upon obtaining the bounds in Lemma~\ref{lemma::ensemble}, 
there are only two unique terms left, 
which we bound in Lemmas~\ref{lemma::finalrate}
and~\ref{lemma::P2PsiP2} respectively.
Notice that by the triangle inequality,  we have
for symmetric matrices $\hat\Sigma_Y- \Sigma_Y = (I-P_1)\Psi(I-P_1)$
and $\hat{Z} -Z^{*}$, 
\ben
\nonumber
\abs{\ip{\cp(\hat\Sigma_Y- \Sigma_Y), \hat{Z} -Z^{*}} }
& \le & 2\abs{\ip{P_2 (\hat\Sigma_Y- \Sigma_Y), \hat{Z} -Z^{*}} }+
\abs{ \ip{P_2 (\hat\Sigma_Y- \Sigma_Y) P_2, \hat{Z} -Z^{*}}  }.
\een
As the first step, we first obtain in Lemma~\ref{lemma::P2PsiP2} a
deterministic bound on $\ip{P_2(\hat\Sigma_Y- \Sigma_Y) P_2, \hat{Z} -Z^{*}}$.
Then, in combination with Theorem~\ref{thm::YYaniso},
we obtain the high probability bound as well.
We prove Lemma~\ref{lemma::P2PsiP2} in
Section~\ref{sec::proofP2ZZP2}.
Next we state Lemma~\ref{lemma::finalrate}, 
which  we prove in Section~\ref{sec::prooffinalrate}. 
\begin{lemma}
  \label{lemma::P2PsiP2}
  Denote by $\Psi := (\Z \Z^T -\E (\Z \Z^T))$.
  Then
\bens
\abs{\ip{P_2(\hat\Sigma_Y- \Sigma_Y) P_2, \hat{Z} -Z^{*}}}
& \le & (1+  \abs{w_1 -   w_2})^2 \twonorm{\Psi} \onenorm{\hat{Z} -Z^{*}} /n.
\eens
Moreover,  under the conditions in Theorem~\ref{thm::YYaniso},
we have with probability at   least $1-\exp(cn)$,
 \bens
\sup_{\hat{Z} \in 
\M_{\opt}} \abs{\ip{P_2(\hat\Sigma_Y- \Sigma_Y) P_2, \hat{Z} -Z^{*}}}
 & \le &
 \frac{1}{3} \xi  p \gamma  \onenorm{\hat{Z} -Z^{*}}. 
 \eens
\end{lemma}

\begin{lemma}
  \label{lemma::finalrate}
 Suppose $r_1 \le 2 qn$   for some positive integer $1\le q < n$.
Suppose all conditions in Theorem~\ref{thm::exprate} hold.
Then, with probability at least $1-\frac{c}{n^2}$,
\bens
\lefteqn{\sup_{\hat{Z} \in \M_{\opt} \cap (Z^{*} + r_1 B_1^{n \times  n})}
  2 \abs{\ip{P_2(\hat\Sigma_Y- \Sigma_Y), \hat{Z} -Z^{*}}}
  \le \frac{1}{3} \xi  p \gamma  \onenorm{\hat{Z} -Z^{*}} + }\\
&&
C_{10}  (C_0 \max_{i} \twonorm{H_i})^2\lceil \frac{r_1}{2n} \rceil
\left(\sqrt{np \log(2e n/\lceil \frac{r_1}{2n} \rceil) }+ \sqrt{r_1} \log(2e n/\lceil \frac{r_1}{2n} \rceil) \right).
\eens
\end{lemma}
Lemma~\ref{lemma::P2cov} follows from Lemmas~\ref{lemma::P2PsiP2}
and~\ref{lemma::finalrate} immediately.
Lemma~\ref{lemma::P2PsiP2} 
follows immediately from Lemma~\ref{lemma::P2P1items}
and Lemma~\ref{lemma::YYcovcorr}, which controls $\twonorm{\Z \Z^T -\E (\Z \Z^T)}$. 
\begin{lemma}{\bf(Deterministic bounds)}
\label{lemma::P2P1items}
Let $\Psi = \Z \Z^T -\E (\Z \Z^T)$. Then
\ben
\label{eq::P2major}
\abs{  \ip{P_2 \Psi P_1, \hat{Z} -Z^{*}}}
& \le &   \twonorm{\Psi} \onenorm{\hat{Z} -Z^{*}}/{n}, \\
\label{eq::P2P1P1}
\abs{\ip{P_2 P_1 \Psi P_1, \hat{Z} -Z^{*}}}
& \le &
\abs{w_1 - w_2} 
\twonorm{\Psi } \onenorm{\hat{Z} -Z^{*}} /n, \\
\text{ and } \; 
  \abs{\ip{P_2 P_1 \Psi P_1 P_2, \hat{Z} -Z^{*}}}
& \le &
\label{eq::P2P1P1P2}
\abs{w_1 - w_2}^2 \twonorm{\Psi} \onenorm{\hat{Z} -Z^{*}}/n.
\een
Finally, we will show the following bounds: for $\hat\delta_Z :=\hat{Z} -Z^{*}$,
   \ben
   \label{eq::P22}
&& \ip{P_2 \Psi P_2, \hat{Z} -Z^{*}}
 \le  \twonorm{\Psi} \onenorm{\hat{Z} -Z^{*}} /n, \\
 \label{eq::P212}
&& \abs{\ip{P_2 \Psi P_1 P_2, \hat\delta_{Z}}} =  \abs{\ip{P_2 P_1
    \Psi P_2, \hat\delta_{Z}} }\le  \twonorm{\Psi} \frac{\abs{w_1
    -w_2} }{n}\onenorm{\hat{Z} -Z^{*}}. 
 \een
 When $w_1 = w_2$, $P_2 P_1 =0$, and the RHS of~\eqref{eq::P2P1P1}, \eqref{eq::P2P1P1P2}
 and~\eqref{eq::P212} all become 0.
\end{lemma}

\subsection{Proof of Lemma~\ref{lemma::P2final}}
\label{sec::P2finalproof}

\begin{proofof2}
We state in Lemmas~\ref{lemma::P2major} and~\ref{lemma::P2majorstar}
two reduction steps.
For all $\hat{Z} \in  M_{\opt}
\cap (Z^* + r_1 B_1^{n \times n})$, we have $\hat{w}  \in B_{\infty}^n
\cap \lceil r_1 /(2n) \rceil B_1^n$, where
\ben
\label{eq::weightnorm}
\hat{w}_j & := & 
\inv{2n}  \onenorm{ (\hat{Z}   -Z^{*})_{\cdot j}}  \le 1 
 \text{ and } \; \sum_{j} \hat{w}_j \le 
 \lceil r_1 /(2n) \rceil,
 \een
 where $B_{\infty}^n$ and $B_1^n$  denote the unit $\ell_{\infty}$ ball and 
 $\ell_1$ ball respectively.
 Denote by $L_i :=\ip{\mu^{(1)} -\mu^{(2)}, \Z_i}$ throughout this 
section. 
 Denote by $L_1^{*} \ge L_2^{*} \ge \ldots \ge L_n^{*}$
 (resp. $\hat{w}_1^{*} \ge \hat{w}_2^{*} \ge \ldots \ge  \hat{w}_n^{*}$) 
the non-decreasing arrangement of $\abs{L_j}$ (resp. $\hat{w}_j$).
We then have the reduction as in Lemma~\ref{lemma::P2majorstar}.
  It remains to bound the sum on the RHS of \eqref{eq::L1}, namely, 
  $\sum_{j =1}^{\lceil {r_1}/{(2n)}\rceil} L_j^{*}$. 
  To do so,  we state in Proposition~\ref{prop::orderL} a high probability bound on 
  $\sum_{j=1}^q L_j^*$, which holds simultaneously for all $q \in  [n]$. 
Denote by $\hat\delta_{Z} := \hat{Z} -Z^{*}$.

\begin{lemma}{\bf (Deterministic bounds)}
\label{lemma::P2major}
Let $\hat{w}_j$ be as in~\eqref{eq::weightnorm}. Then for $\hat\delta_{Z} := \hat{Z} -Z^{*}$,
\ben
\label{eq::2L}
  \abs{\ip{P_2(\E(Y) (Y -\E(Y))^T), \hat\delta_{Z}}}
 \le 
  4 w_1 w_2 n \sum_{j =1}^{n} \abs{L_i}\hat{w}_j + 
\frac{2w_1 w_2}{n}  \abs{\sum_{i =1}^{n} L_i }\onenorm{\hat\delta_{Z} }.
\een
\end{lemma}

\begin{lemma}{\bf(Reduction to order statistics)}
\label{lemma::P2majorstar}
Under the settings of Lemma~\ref{lemma::P2major}, we have
\ben
\label{eq::L1}
\sup_{\hat{Z} \in \M_{\opt} \cap (Z^{*} +  r_1 B_1^{n \times n})} \abs{  \ip{P_2(\E(Y) (Y -\E(Y))^T), \hat{Z} -Z^{*}}}  \le 3 n \sum_{j =1}^{\lceil {r_1}/{2n}\rceil} L_j^{*}.
\een
\end{lemma}

\begin{proposition}
  \label{prop::orderL}
  Suppose all conditions in Lemma~\ref{lemma::twogroup} hold.
  Let random matrix $\Z$ satisfy the conditions as stated  therein.
  Let  $H_i$ and  $\mu = {(\mu^{(1)}-\mu^{(2)})}/\sqrt{p \gamma}$ be as 
defined in Theorem~\ref{thm::YYaniso}. 
Let $1 \le q \le n$ denote a positive integer. Then for some absolute constants $C_5, c$,
we have with probability at least $1-\frac{c}{n^2}$, simultaneously for all $q \in [n]$,
\bens 
\sum_{j=1}^q L_j^* \le C_5 C_0 (\max_{i} \twonorm{H^T_i \mu}) \sqrt{p \gamma} q \sqrt{\log (2e n/q)}.
\eens
\end{proposition}

It remains to prove Lemma~\ref{lemma::P2final}.
Indeed, we have with probability at least $1-\frac{c}{n^2}$, by
\eqref{eq::2L} and \eqref{eq::L1}, and Proposition~\ref{prop::orderL},
for $ w_1 w_2 \le 1/4$,
\bens 
\lefteqn{
  \sup_{\hat{Z} \in \M_{\opt} \cap (Z^{*} +  r_1 B_1^{n \times n})}
\abs{  \ip{P_2(\E(Y) (Y -\E(Y))^T), \hat{Z} -Z^{*}}} }\\
\nonumber
  & \le &
  3 n \sum_{j =1}^{\lceil {r_1}/{(2n)}\rceil} L_j^{*}
  \le C_3 C_0 (\max_{i} \twonorm{H^T_i \mu}) \Delta r_1 \sqrt{\log (2e
    n/\lceil \frac{r_1}{2n}\rceil)},
  \eens
  where it is understood that $q = \lceil \frac{r_1}{2n}\rceil 
  \in [n]$ for some $r_1 > 0$.
We prove Lemmas~\ref{lemma::P2major} and~\ref{lemma::P2majorstar}, in 
Sections~\ref{sec::P2majorproof}
and~\ref{sec::proofP2majorstar} respectively.
We prove Proposition~\ref{prop::orderL} in
Section~\ref{sec::prooforderL}.
\end{proofof2}

\subsection{Proof of Lemma~\ref{lemma::finalrate}}
\label{sec::prooffinalrate}
First we note that by \eqref{eq::covproj}, for $\hat\delta_Z :=\hat{Z} -Z^{*}$,
\ben
\label{eq::P2SigmaY}
\lefteqn{\ip{(\hat\Sigma_Y- \Sigma_Y) P_2, \hat\delta_{Z}} = 
\ip{P_2(\hat\Sigma_Y- \Sigma_Y),  \hat\delta_{Z}} }\\
& = &
\nonumber
\ip{P_2 \Psi, \hat\delta_{Z}} - \ip{P_2 P_1 \Psi , \hat\delta_{Z}} +
\ip{P_2 P_1 \Psi P_1, 
  \hat\delta_{Z}} -\ip{P_2 \Psi P_1, \hat\delta_{Z}}.
\een
\begin{lemma}{\textnormal{(Deterministic bounds)}}
  \label{lemma::P2SigmaY}
Denote by $\Psi=  \Z \Z^T -\E (\Z \Z^T)$ and 
$\offd(\Psi)$ its off-diagonal component, where the diagonal elements are set to be $0$. Denote by 
\ben
\label{eq::defineQ}
Q_{j} & := &
\sum_{t  \in \MC_1, t \not=j}  \Psi_{j t} - \sum_{t  \in \MC_2, t \not=j} \Psi_{j t}
= \sum_{t  \in \MC_1}  \offd(\Psi)_{jt} - \sum_{t  \in \MC_2}
\offd(\Psi)_{jt}, \\
\nonumber
S_{j} & := &
\sum_{t  \in [n], t \not=j}  \Psi_{j t} =\sum_{t  \in [n]}  \offd(\Psi)_{j t}.
\een
Let $\hat{w}_i$ be the same as in \eqref{eq::weightnorm}.
Then for $\norm{\diag(\Psi)}_{\max} := \max_{j} \abs{\Psi_{jj}}$,
\ben
\label{eq::P2ZZdiag}
\abs{\ip{P_2 \diag(\Psi), \hat{Z} -Z^{*}}}
& \le &
\norm{\diag(\Psi)}_{\max} \onenorm{\hat{Z} -Z^{*}}/n \\
\label{eq::Sdiag}
\abs{\ip{P_2 P_1 \diag(\Psi), \hat{Z} -Z^{*}}}
& \le &
2 \abs{w_1 - w_2} \norm{\diag(\Psi)}_{\max}\onenorm{\hat{Z} -Z^{*}}/n \\
\label{eq::P2ZZ}
\abs{\ip{P_2 \offd(\Psi), \hat{Z} -Z^{*}}}
  &  \le &
  2 \sum_{j \in [n]} \abs{Q_{j}}  \hat{w}_j, \quad \text{ and } \\
\label{eq::doublesum}
\abs{\ip{P_2 P_1 \offd(\Psi), \hat{Z} -Z^{*}}}
& \le &
2 \abs{w_1 - w_2} \sum_{k} \abs{S_{k} }    \cdot \hat{w}_k.
\een
\end{lemma}
We prove Lemmas~\ref{lemma::P2SigmaY} and~\ref{lemma::finalrate} in 
Sections~\ref{sec::proofofP2SigmaY} and~\ref{sec::suppfinalrate}.
Denote by $Q_1^{*} \ge Q_2^{*} \ge \ldots \ge Q_n^{*}$ (resp. $\hat{w}_1^{*} \ge \hat{w}_2^{*} \ge \ldots \ge  \hat{w}_n^{*}$) 
 the non-decreasing arrangement of $\abs{Q_i}$ (resp. $\hat{w}_j$). 
 Denote by $S_1^{*}\ge S_2^{*} \ge \ldots \ge S_n^{*}$
 the non-decreasing arrangement of $\abs{S_i}$  (resp. $\hat{w}_j$).
The RHS of  \eqref{eq::doublesum} and~\eqref{eq::P2ZZ} are bounded to be at the same order; cf. Proposition~\ref{prop::orderQ}.
Lemma~\ref{lemma::finalrate} follows from Lemma~\ref{lemma::P2SigmaY}, 
Proposition~\ref{prop::orderQ}, and Corollary~\ref{coro::orderQ}.
We prove Proposition~\ref{prop::orderQ} and Corollary~\ref{coro::orderQ} in 
Sections~\ref{sec::prooforderQ} and~\ref{sec::prooforderQcoro} respectively.
Let $C_4, C_5, c, \ldots$ be  absolute  constants. 
 \begin{proposition}
   \label{prop::orderQ}
Under the conditions in Theorem~\ref{thm::YYaniso},
we have with probability at least $1-\frac{c}{n^2}$, simultaneously
for all positive integers $q \in [n]$, 
\ben
\label{eq::Q2}
\max\big(\sum_{j =1}^{q} Q_j^{*}, \sum_{j =1}^{q} S_j^{*} \big) \le
C_4 (C_0 \max_{i} \twonorm{H_i})^2 q \big(\sqrt{np \log(2e
  n/q)} + \sqrt{nq} \log(2e n/q) \big).
\een
\end{proposition}

\begin{corollary}
\label{coro::orderQ}
Under the settings in Proposition~\ref{prop::orderQ},
we have with probability at least $1 - \frac{c}{n^2}$,
 \ben
 \label{eq::duet}
\sup_{\hat{Z} \in \M_{\opt} \cap (Z^{*} + r_1 B_1^{n \times n})}
   \abs{\ip{P_2 \offd(\Psi), \hat{Z} -Z^{*}}}
   &\le &
   4 \sum_{j =1}^{\lceil r_1 /(2n) \rceil} Q_j^{*}   \quad \text{and
   }  \\
\label{eq::trio}
   \sup_{\hat{Z} \in \M_{\opt} \cap (Z^{*} + r_1 B_1^{n \times n})}
   \abs{\ip{P_2 P_1 \offd(\Psi), \hat{Z} -Z^{*}}}  &\le &
   4 \abs{w_1 - w_2}   \sum_{j =1}^{\lceil {r_1}/{(2n)} \rceil} S_j^{*}.
   \een
   \end{corollary}

\section{Additional proofs for Theorem~\ref{thm::exprate}}
\label{sec::expdetail}
First we state
Propositions~\ref{fact::weights},~\ref{fact::weightsappend} and 
~\ref{fact::Zu2}.
\begin{proposition}
  \label{fact::weights}
  Let $Z_{\cdot j}$ represent the $j^{th}$ column of $Z$.
 Let $\hat{Z} \in \M_{\opt}$ and $\onenorm{\hat{Z} - Z^*} \le r_1 \le 2 qn$,
 where $q = \lceil \frac{r_1}{2n}\rceil < n$ is a positive integer
 and  $Z^{*}$ is as defined in \eqref{eq::refoptsol}.
Let
\ben
 \label{eq::weightsum}
   \hat{w}_j :=   \inv{2n}\sum_{i \in [n]}  \abs{(\hat{Z} -Z^{*})_{ij}} < 1
   \text{ and hence }\;    \sum_{j =1}^{n} \hat{w}_j :=  \onenorm{\hat{Z} -Z^{*}}/(2n) \le q  < n.
    \een
 Hence the weight vector $\hat{w} =(\hat{w}_1, 
   \ldots,\hat{w}_n) \in B_{\infty}^{n} \cap q B_1^{n}$, where 
   $B_{\infty}^{n} = \{y \in \R^n: \abs{y_j} \le 1\}$
   and $B_1^{n} = \{y \in \R^n: \sum_{j} \abs{y_j} \le 1\}$
 denote the unit $\ell_{\infty}$ ball and   $\ell_1$ ball respectively.
\end{proposition}

\begin{proposition}
  \label{fact::weightsappend}
 Under the settings in Proposition~\ref{fact::weights},
 we have for $\hat{w}_j$ as in~\eqref{eq::weightsum}, 
\bens
\forall j \in \MC_1, \; \; 
\hat{w}_{j} & = &
-\inv{2n}\big(\sum_{i \in \MC_1}  (\hat{Z} -Z^{*})_{ij} - \sum_{i \in \MC_2}
  (\hat{Z} -Z^{*})_{ij} \big) \quad \text{ and} \\
\forall j \in \MC_2, \; \;
\hat{w}_{j}
& = &
\inv{2n} \big(\sum_{i \in \MC_1}  (\hat{Z} -Z^{*})_{ij} - \sum_{i \in \MC_2}
  (\hat{Z} -Z^{*})_{ij} \big).
  \eens
 \end{proposition}

  \begin{proofof}{Proposition~\ref{fact::weightsappend}} 
By definition of $Z^{*}$, we have $\forall j \in \MC_1$,
\bens
\forall  i \in \MC_2, &&  (\hat{Z} -Z^{*})_{ij}  \ge 0 \quad \text{ and } \quad
\forall i \in \MC_1, \; \; (\hat{Z} -Z^{*})_{ij}  \le 0,
\eens
and hence
\bens
\hat{w}_{j} & = &
\inv{2n}\big(\sum_{i \in \MC_2} (\hat{Z} -Z^{*})_{ij} - \sum_{i \in
  \MC_1}  (\hat{Z} -Z^{*})_{ij} \big) =\onenorm{(\hat{Z} -Z^{*})_{\cdot j}}/(2n) < 1.
\eens
Similarly, we have $\forall j \in \MC_2$,
\bens
\forall  i \in \MC_1, && (\hat{Z} -Z^{*})_{ij}  \ge 0
\text{ and } \quad \forall  i \in \MC_2,  (\hat{Z} -Z^{*})_{ij}  \le
0,
\eens
and  hence
\bens
\hat{w}_{j}
& = &
\inv{2n} \big(\sum_{i \in \MC_1}  (\hat{Z} -Z^{*})_{ij} - \sum_{i \in \MC_2}
  (\hat{Z} -Z^{*})_{ij} \big)= \onenorm{(\hat{Z} -Z^{*})_{\cdot j}}/(2n) < 1.
\eens
\end{proofof}

Proposition~\ref{fact::Zu2} follows from 
Propositions~\ref{fact::weights} and~\ref{fact::weightsappend} immediately. 

\begin{proposition}
 \label{fact::Zu2}
We have by symmetry of $\hat{Z} -Z^{*}$, under the settings in Proposition~\ref{fact::weights},
\bens
  \inv{2n} \abs{\vecone_n^T( \hat{Z} -Z^{*}) u_2}
& = &
\abs{\inv{2n} \sum_{i=1}^n 
  \big(\sum_{j\in\MC_1} (\hat{Z} -Z^{*})_{ij} -\sum_{j \in \MC_2} (\hat{Z}
  -Z^{*})_{ij}\big)} \\
& \le &
\sum_{i=1}^n \abs{\hat{w}_i}=\inv{2n}  \onenorm{ (\hat{Z} -Z^{*})}
= \inv{2n} \sum_{j} \onenorm{(\hat{Z} -Z^{*})_{j\cdot} }.
\eens
\end{proposition}

\subsection{Proof outline of Lemma~\ref{lemma::P2minor}}
\label{sec::P2minorproof}
 First, we show that \eqref{eq::rotation2} and \eqref{eq::rotation} hold.
To see this, we have
\ben
\nonumber
     \ip{(Y -\E(Y)) \E(Y)^T P_2,  \hat{Z} -Z^{*}}
     &  = & \tr((\hat{Z} -Z^{*}) (Y -\E(Y)) \E(Y)^T P_2)\\
   &  = &
   \ip{P_2 \E(Y) (Y -\E(Y))^T, \hat{Z}-Z^{*}}.\\
\nonumber
     \ip{P_2 (Y -\E(Y)) \E(Y)^T,  \hat{Z} -Z^{*}}
 &  = &   \tr((\hat{Z} -Z^{*}) P_2 (Y -\E(Y)) \E(Y)^T) \\
   \label{eq::roationlocal}
   &  = &
   \ip{\E(Y) (Y -\E(Y))^T P_2, \hat{Z}-Z^{*}}.
   \een
 Denote by $V = (\E(Y) (Y -\E(Y))^T) u_2 =(V_1, \ldots, V_n)$ such that 
\bens
\forall i \in \MC_1, \quad 
V_i &  = & - w_2
\ip{\mu^{(2)} - \mu^{(1)}, \sum_{j \in C_1} (Y_j - \E(Y_j))  -
  \sum_{j \in C_2} (Y_j - \E(Y_j))},  \\
\forall i \in \MC_2, \quad 
V_i &  = &  w_1 \ip{\mu^{(2)} - \mu^{(1)}, \sum_{j \in C_1} (Y_j - \E(Y_j)) 
  -  \sum_{j \in C_2} (Y_j - \E(Y_j))},
\eens
where $Y_j \in \R^{p}$ represents the $j^{th}$ row vector of matrix $Y$.
 Similarly, we have by \eqref{eq::roationlocal},
 \bens
\nonumber
    \lefteqn{\ip{P_2 (Y -\E(Y)) \E(Y)^T,  \hat{Z} -Z^{*}}    = 
      \tr((\hat{Z}-Z^{*}) \E(Y) (Y -\E(Y))^T P_2) } \\
\nonumber
\nonumber
&  = &
\inv{n} \sum_{k}  V_k \big(\sum_{i \in \MC_1}
(\hat{Z} -Z^{*})_{i k} - \sum_{i \in \MC_2}
(\hat{Z} -Z^{*})_{i k} \big) =: S.
\eens
Hence by~\eqref{eq::weightsum} and Proposition~\ref{fact::weights}, with
some algebra, one obtains
\ben 
\label{eq::finalYY}
S & = &
2 C_S
\cdot \ip{\mu^{(2)} - \mu^{(1)}, \sum_{j \in C_1} (Y_j - \E(Y_j))  -
  \sum_{j \in C_2} (Y_j - \E(Y_j))},  \\
\nonumber
\text{ where } && 
C_S := w_2 \sum_{k \in \MC_1} \hat{w}_k + w_1 \sum_{k \in \MC_2}
\hat{w}_k \le  \sum_{k \in [n]} \hat{w}_k = \onenorm{\hat{Z} -Z^{*}}/(2n).
\een
Then
\ben
\nonumber
\lefteqn{\ip{P_2 \E(Y) (Y -\E(Y))^T P_2, \hat{Z} -Z^{*}} =
  \tr(P_2 (\hat{Z} -Z^{*})  P_2 \E(Y) (Y -\E(Y))^T) }\\
&  = &
\label{eq::P2YEYP2}
\inv{2n} u_2^T (\hat{Z} -Z^{*}) u_2 \big(2u^T_2 \E(Y) (Y -\E(Y))^T u_2 /n\big)
=: -W \sum_{k} \hat{w}_k, \; \text{where}\\
\label{eq::P2sum}
W & := & 2 u_2^T \E(Y) (Y-\E(Y))^T u_2 /n = 2 u_2^T V/n =
\frac{2}{n}(\sum_{i \in \MC_1} V_i -\sum_{i \in \MC_2} V_i ) \\
& = &
\nonumber
\ip{\mu^{(1)}-\mu^{(2)}, 
  \sum_{j \in \MC_1} (Y_j - \E Y_j) -   \sum_{j \in \MC_2} (Y_j - \E
  Y_j)} 4 w_1 w_2.
\een
where \eqref{eq::P2YEYP2} holds
by \eqref{eq::P2sum} and Proposition~\ref{fact::trP2ZZ},
since for $P_2 = {u_2 u_2^T}/{n} = Z^*/n$,
$\frac{u_2^T}{2n} (\hat{Z} -Z^{*}) u_2 = 
\tr(P_2 (\hat{Z} -Z^{*}))/2 = -\onenorm{\hat{Z} -
  Z^*}/(2n) = -\sum_{k} \hat{w}_k.$
Putting things together, we have by \eqref{eq::weightsum}, \eqref{eq::finalYY} and \eqref{eq::P2YEYP2},
 \bens
 \lefteqn{T_1 :=
\ip{\E(Y) (Y -\E(Y))^T P_2, \hat{Z} -Z^{*}}
- \ip{P_2 \E(Y) (Y   -\E(Y))^T P_2, \hat{Z} -Z^{*}}} \\
& = &
S +W \sum_{k} \hat{w}_k = 
 2\ip{\mu^{(2)} - \mu^{(1)}, \sum_{j \in C_1} (Y_j - \E(Y_j))  -
   \sum_{j \in C_2} (Y_j - \E(Y_j))} \cdot \\
 &&
\big(w_2\sum_{k \in   \MC_1} \hat{w}_k 
 + w_1 \sum_{k \in \MC_2} \hat{w}_k - 2 w_1 w_2 \sum_{k \in [n]}
 \hat{w}_k \big), \\
 \text{ where}
&& \abs{w_2\sum_{k \in   \MC_1} \hat{w}_k 
 + w_1 \sum_{k \in \MC_2} \hat{w}_k - 2 w_1 w_2 \sum_{k \in [n]}
 \hat{w}_k}
\le \inv{2n}\onenorm{\hat{Z} -Z^{*}}\le q < n.
\eens
Now, for $w_1= w_2 = 1/2$, the above sum is 0 and $T_1 =0$.
Let $r_1 \le 2 qn$.
Then, by Corollary~\ref{coro::sumY}, we have with probability 
$\ge 1-2\exp(-cn)$, for all $\hat{Z} \in \M_{\opt} \cap (Z^{*} + r_1 B_1^{n \times   n})$,
\bens
\lefteqn{
  \abs{T_1} \le \abs{\ip{\mu^{(2)} - \mu^{(1)}, \sum_{j \in C_1} (Y_j - \E(Y_j))  -
     \sum_{j \in C_2} (Y_j - \E(Y_j))}} \onenorm{\hat{Z} -Z^{*}}/n }\\
 &  \le & C_4 C_0 (\max_{i} \twonorm{R_i \mu}) \sqrt{p \gamma} \onenorm{\hat{Z} - Z^*}
 \le  C_4 C_0 (\max_{i} \twonorm{R_i \mu}) r_1 \sqrt{p \gamma}.\quad \scriptstyle\Box
 \eens

\subsection{Proof of Lemma~\ref{lemma::P2PsiP2}}
\label{sec::proofP2ZZP2}

\begin{proofof2}
The  deterministic statement in Lemma~\ref{lemma::P2PsiP2} holds in
view of
\eqref{eq::P2P1P1P2},~\eqref{eq::P22}, and~\eqref{eq::P212}, since
\bens
\label{eq::P2PsiP2}
&& \ip{P_2(\hat\Sigma_Y- \Sigma_Y) P_2, \hat{Z} -Z^{*}}
 = 
\ip{P_2 (I-P_1) \Psi (I-P_1) P_2, \hat{Z} -Z^{*}},  \; \text{ and hence}  \\
&& \lefteqn{\abs{\ip{P_2(\hat\Sigma_Y- \Sigma_Y) P_2, \hat{Z} -Z^{*}}}
\le 
\abs{\ip{P_2  \Psi P_2, \hat{Z} -Z^{*}}  }+
\abs{\ip{P_2 P_1 \Psi P_1 P_2, \hat{Z} -Z^{*}} } + }\\
\nonumber
&&
2 \abs{\ip{P_2 P_1 \Psi P_2, \hat{Z} -Z^{*}}}
\le \inv{n}\onenorm{\hat{Z} -Z^{*}  } \twonorm{\Psi} (1 + \abs{w_1 -w_2})^2. 
\eens
Thus, we have with probability at least $1-\exp(c' n)$, for all
$\hat{Z} \in \M_{\opt}$, 
\bens
\abs{\ip{P_2(\hat\Sigma_Y- \Sigma_Y) P_2, \hat{Z} -Z^{*}}}
& \le &
\label{eq::lastfac}
8 \twonorm{\Psi} \onenorm{\hat{Z} -Z^{*}} /(2n) \le
\frac{1}{3} \xi  p \gamma   \onenorm{\hat{Z} -Z^{*}},
\eens
 by Lemma~\ref{lemma::YYcovcorr}. The lemma is thus proved.
\end{proofof2}

\subsection{Proof of Lemma~\ref{lemma::finalrate}}
\label{sec::suppfinalrate}
  Let $\hat{Z} \in \M_{\opt} \cap ( Z^{*} + r_1 B_1^{n \times n})$, 
  where $r_1 \le 2 q n$  for some positive integer $1\le q \le n$. 
By  Lemma~\ref{lemma::P2P1items} (cf.  \eqref{eq::P2major} and
  \eqref{eq::P2P1P1}), we have for all $\hat{Z} \in \M_{\opt}$,
\bens
W_1 := \abs{  \ip{P_2 \Psi P_1, \hat{Z} -Z^{*}}} + 
\abs{\ip{P_2 P_1 \Psi P_1, \hat{Z} -Z^{*}}}  \le
(1+\abs{w_1 - w_2} ) \twonorm{\Psi} \inv{n} \onenorm{\hat{Z} -Z^{*}}.
\eens
Thus we have
by the triangle inequality, by~\eqref{eq::P2SigmaY},
\bens
\abs{\ip{P_2(\hat\Sigma_Y- \Sigma_Y), \hat{Z} -Z^{*}}}
  \le  W_1 + W_2,\;\text{ where}\;
W_2 :=   \abs{\ip{P_2 \Psi, \hat{Z} -Z^{*}} } +  \abs{\ip{P_2 P_1 \Psi ,
    \hat{Z} -Z^{*}} },
  \eens
  where we decompose for $\hat\delta_Z :=\hat{Z} -Z^{*}$,
  \bens
  W_2
& \le & \abs{\ip{P_2 \diag(\Psi), \hat\delta_Z }} +
\abs{\ip{P_2 P_1\diag(\Psi) , \hat\delta_Z}} \quad (W_2(\diag)) \\
&&
+ \abs{\ip{P_2 \offd(\Psi), \hat\delta_Z}}+
\abs{\ip{P_2 P_1 \offd(\Psi) , \hat\delta_Z}}
\quad (W_2(\offd)).
\eens
Then, we have for  all $\hat{Z} \in Z^{*} + r_1 B_1^{n \times n}$,
by \eqref{eq::P2P1P1}, \eqref{eq::P2major}, \eqref{eq::P2ZZdiag},
and \eqref{eq::Sdiag},
\ben
 \nonumber
 \lefteqn{
   W_2(\diag) + W_1 =  \abs{\ip{P_2 \diag(\Psi), \hat{Z} -Z^{*}}} + 
\abs{\ip{P_2 P_1 \diag(\Psi), \hat{Z} -Z^{*}}} + W_1} \\
\label{eq::Psibounds}
& \le & (1+\abs{w_1 - w_2}) \max_{k} (\abs{\Psi_{k k}} +\twonorm{\Psi})
\onenorm{\hat{Z} -Z^{*}}/n 
 \le  4 \twonorm{\Psi} \onenorm{\hat{Z} -Z^{*}}/n,
\een
and by~\eqref{eq::duet} and~\eqref{eq::trio}, for $ W_2(\offd) =\abs{\ip{P_2 \offd(\Psi), \hat{Z} -Z^{*}} } +
\abs{\ip{P_2 P_1 \offd(\Psi) , \hat{Z} -Z^{*}} }$,
\ben
\sup_{\hat{Z} \in \M_{\opt} \cap (Z^{*} + r_1 B_1^{n \times n})}  
  W_2(\offd) 
& \le &
\label{eq::W2offd}
4 \sup_{\hat{w} \in B_{\infty}^{n} \cap \lceil \frac{r_1}{2n} \rceil B_1^{n}}
\big(\sum_{j =1}^{\lceil \frac{r_1}{2n} \rceil} Q_j^{*} +\abs{w_1 - w_2}\sum_{j =1}^{\lceil \frac{r_1}{2n} \rceil} S_j^{*}\big).
\een
Let $\Psi = \Z \Z^T -\E(\Z \Z^T)$ and $q =\lceil r_1 /(2n) \rceil$,
The lemma thus holds by Proposition~\ref{prop::orderQ} and
Corollary~\ref{coro::orderQ},~\eqref{eq::Psibounds} and
\eqref{eq::W2offd}, since
\bens
&& \sup_{\hat{Z} \in \M_{\opt} \cap (Z^{*} + r_1 B_1^{n \times n})}
\abs{\ip{P_2(\hat\Sigma_Y- \Sigma_Y), \hat{Z} -Z^{*}}}
 \le   \sup_{\hat{Z} \in \M_{\opt} \cap (Z^{*} + r_1 B_1^{n \times n})} 
    4 \twonorm{\Psi} \onenorm{\hat{Z} -Z^{*}}/n\\
  &  & +
\quad 4 \sup_{\hat{w} \in B_{\infty}^{n} \cap \lceil r_1 /(2n) \rceil B_1^{n}}
\big(\sum_{j =1}^{\lceil r_1 /(2n) \rceil} Q_j^{*} +\abs{w_1 -
  w_2}\sum_{j =1}^{\lceil {r_1}/{(2n)} \rceil} S_j^{*}\big). \quad \scriptstyle\Box
\eens

\subsection{Proof of Lemma~\ref{lemma::P2P1items}}
\label{sec::proofofLemmaP2P1}

\begin{proofof2}
  Throughout this proof, denote by $\Psi= \Z \Z^T -\E (\Z \Z^T)$.
  Let $u_2$ be   the group membership vector as in~\eqref{eq::u2}.
  Let $\vecone_n =(1, \ldots, 1)$ and $\vecone_n/\sqrt{n} \in
  \Sp^{n-1}$.  Hence we have $u_2^T \vecone_n = \abs{C_1} - \abs{C_2}$.   
First, we have by Proposition~\ref{fact::Zu2}, \eqref{eq::P2major} holds since
\bens
\abs{  \ip{P_2 \Psi P_1, \hat{Z} -Z^{*}}}
& \le &
\abs{\frac{2 u_2^T \Psi \vecone_n}{n} }
\abs{\frac{\vecone_n^T (\hat{Z} -Z^{*}) u_2}{2n}}
\le  \twonorm{\Psi} \onenorm{\hat{Z} -Z^{*}}/{n}.
\eens
Second,~\eqref{eq::P2P1P1} holds
since ${\vecone_n^T  \Psi \vecone_n}/{n} \le \twonorm{\Psi}$ and
\bens
\ip{P_2 P_1\Psi P_1, \hat{Z} -Z^{*}}
& = &
2 \frac{\abs{C_1} - \abs{C_2}}{n}\frac{\vecone_n^T \Psi \vecone_n}{n}
\inv{2n} \vecone_n^T( \hat{Z} -Z^{*}) u_2.
\eens
Hence by Proposition~\ref{fact::Zu2},
$\abs{\ip{P_2 P_1 \Psi P_1, \hat{Z} -Z^{*}}} \le
\abs{w_1 - w_2} \twonorm{\Psi} 
\onenorm{\hat{Z} -Z^{*}}/{n}$.
To bound \eqref{eq::P2P1P1P2}, we have by Proposition~\ref{fact::trP2ZZ},
\bens
\ip{P_2 P_1\Psi P_1 P_2, \hat{Z} -Z^{*}}
& = &
(w_1 -w_2)^2\frac{\vecone_n^T \Psi \vecone_n}{n} \ip{\frac{u_2
    u_2^T}{n}, \hat{Z} -Z^{*}}, \; \text{ and hence}  \\
\abs{\ip{P_2 P_1\Psi P_1 P_2, \hat{Z} -Z^{*}}}
& \le  & (w_1 -w_2)^2 \twonorm{\Psi} \onenorm{\hat{Z} -Z^{*}}/n.
\eens
Finally, it remains to show ~\eqref{eq::P22} and~\eqref{eq::P212}.
First, we rewrite the inner product as follows:
 \bens
\ip{\hat{Z} -Z^{*},  P_2 \Psi P_2}
&  = &
\frac{u_2^T (\hat{Z} -Z^{*})^T u_2}{n}
\frac{u_2^T \Psi  u_2}{n} =
- \inv{n}\onenorm{\hat{Z} -Z^{*}  } \frac{u_2^T \Psi  u_2}{n};
\eens
Then~\eqref{eq::P22} holds by Proposition~\ref{fact::trP2ZZ}.
Finally,~\eqref{eq::P212} holds since
\bens
\ip{P_2 P_1 \Psi P_2, \hat{Z} -Z^{*}}
&  = & \tr(\frac{u_2^T}{n} (\hat{Z} -Z^{*}) u_2) \frac{u_2^T \vecone_n }{n}
\frac{\vecone_n^T \Psi  u_2}{n},\\
\ip{P_2 \Psi P_1 P_2, \hat{Z} -Z^{*}}
&  = &
\tr(P_2 (\hat{Z} -Z^{*}) )
\frac{u_2^T  \Psi  \vecone_n}{n}
\frac{\vecone_n^T u_2}{n}; \; \text{ and hence} \\
\abs{  \ip{P_2 P_1 \Psi P_2, \hat{Z} -Z^{*}}}  &  = &
\abs{\ip{P_2 \Psi P_1 P_2, \hat{Z} -Z^{*}}}  \le
\abs{w_1 - w_2} \inv{n} \onenorm{\hat{Z} -Z^{*}} \twonorm{ \Psi}.
\eens

\end{proofof2}

\subsection{Proof of Lemma~\ref{lemma::P2major}}
\label{sec::P2majorproof}
  The proof of  Lemma~\ref{lemma::P2major} is entirely deterministic.
    Recall by~\eqref{eq::EYpre},
    we have
\bens
\inv{n} \E(Y)^T u_2 = \inv{n}(\sum_{i\in \MC_1} \E(Y_i) -\sum_{i\in \MC_2} \E(Y_i) )
= 2 w_1 w_2 (\mu^{(1)} -  \mu^{(2)}).
\eens
  Throughout this proof, we denote by
  $\mu := {(\mu^{(1)}-\mu^{(2)})}/\sqrt{p \gamma}$.
  Let $Z^{*}$ be as defined in \eqref{eq::refoptsol}.
Thus we have by symmetry of $\hat{Z} -Z^{*}$ and definition of
$\hat{w}_j$ in \eqref{eq::weightsum} and $P_2 := u_2 u_2^T/n = Z^{*}/n$,
  \bens
  \lefteqn{  \ip{P_2(\E(Y) (Y -\E(Y))^T), \hat{Z} -Z^{*}} = 
  \inv{n} u_2^T (\E(Y) (Y -\E(Y))^T)( \hat{Z} -Z^{*}) u_2 }\\
   &  = &
 4 n w_1 w_2 \sqrt{p \gamma}
\big(\sum_{j \in \MC_2}  \hat{w}_j \ip{\mu, Y_j - \E (Y_j)} 
  -   \sum_{j \in \MC_1} \hat{w}_j \ip{\mu, Y_j - \E (Y_j)} \big),
  \eens
Moreover, we have for each $j \in [n]$, $n \ip{\mu, Y_j - \E (Y_j)} = n \ip{\mu, \Z_j} - \sum_{i =1}^{n}
\ip{\mu, \Z_i}$.
Hence we have by  \eqref{eq::weightsum}, and $L_j  = \sqrt{p  \gamma}\ip{\mu, Z_j}$,
\bens
\ip{P_2(\E(Y) (Y -\E(Y))^T), \hat{Z} -Z^{*}}
\le
4 w_1 w_2 n \sum_{j \in [n]} \abs{L_i} \hat{w}_j 
+   \frac{4 w_1 w_2}{2n} \abs{\sum_{i =1}^{n}  L_i}  \onenorm{(\hat{Z}
  -Z^{*})}. \quad \quad \scriptstyle\Box
\eens

\section{Proof of Lemma~\ref{lemma::P2majorstar}}
  \label{sec::proofP2majorstar}
    Let $1 \le q < n$ denote a positive integer.
    By Proposition~\ref{fact::weights}, we have 
$\hat{w} =(\hat{w}_1, \ldots,\hat{w}_n) \in B_{\infty}^{n} \cap \lceil
\frac{r_1}{2n}\rceil  B_1^{n}$, for all $\hat{Z}  \in \M_{\opt} \cap
(Z^{*} +  r_1 B_1^{n \times n})$. Then
for the first component on the RHS of~\eqref{eq::2L} in
Lemma~\ref{lemma::P2major}, we have
\ben
  \nonumber 
4 w_1 w_2\sum_{j \in [n]} \abs{L_j} \hat{w}_j
& \le &
\sum_{j \in [n]} L_j^{*}  \hat{w}^{*}_j =
\sum_{j =1}^{q} L_j^{*}  \hat{w}^{*}_j +
\sum_{j =q+1}^{n} L_j^{*}  \hat{w}^{*}_j, \; \text{ and }\\
\sup_{\hat{w} \in B_{\infty}^{n} \cap q B_1^{n}}  \sum_{j \in [n]} \abs{L_j} \hat{w}_j 
& \le &
\nonumber 
\sup_{\hat{w} \in B_{\infty}^{n} }
\sum_{j =1}^{q} L_j^{*}  \hat{w}^{*}_j +
  \sup_{\hat{w} \in q B_1^{n}}
  \sum_{j =q+1}^{n} L_j^{*}  \hat{w}^{*}_j \\
  & \le &
  \label{eq::L5}
\sum_{j =1}^{q} L_j^{*}  + q  L_{q+1}^{*}
\le 2 \sum_{j =1}^{q} L_j^{*}.
\een
Moreover, for the second component on the RHS of~\eqref{eq::2L},
$W_L:= \abs{\sum_{j =1}^{n}  L_j}
 \onenorm{(\hat{Z}  -Z^{*})}/(2n)   \le \sum_{j =1}^{n}  \abs{L_j}
 \abs{\sum_{j} \hat{w}_j} =: U_L$,
where by~\eqref{eq::weightsum}, $\onenorm{(\hat{Z}  -Z^{*})}/(2n) =
\abs{\sum_{j} \hat{w}_j }$, and hence
\ben 
\label{eq::L2}
 \sup_{\hat{Z} \in \M_{\opt} \cap Z^{*} + r_1 B_1^{n \times n}} W_L/n \le
  \sup_{\hat{w} \in B_{\infty}^{n} \cap \lceil \frac{r_1}{2n}\rceil B_1^{n}}
  U_L/{n}
   \le  \frac{\lceil {r_1}/{2n}\rceil}{n}\sum_{j=1}^n   L^*_j 
  \le \sum_{j =1}^{\lceil {r_1}/{(2n)}\rceil} L_j^{*}. 
\een
where \eqref{eq::L2} holds since in the second sum $\sum_{j =1}^{q} L_j^{*}$,
where $q = \lceil \frac{r_1}{2n}\rceil$,
we give a unit weight to each of the $q$ largest components $L_j^{*},
j \in [q]$, while in $\frac{q}{n} \sum_{i=1}^n L^*_j$, we give the
same weight $q/n \le 1$ to each of $n$ components $L_j^{*}, j \in [n]$.
Putting things together, we have by~\eqref{eq::2L},~\eqref{eq::L2}, and~\eqref{eq::L5},
where we set $q = \lceil \frac{r_1}{2n}\rceil \in [n]$,
  \bens
  \nonumber
\lefteqn{
  \sup_{\hat{Z} \in \M_{\opt} \cap (Z^{*} +  r_1 B_1^{n \times n})} \abs{  \ip{P_2(\E(Y) (Y -\E(Y))^T), \hat{Z} -Z^{*}}}} \\
  \label{eq::L1proof}
& \le &
\sup_{\hat{w} \in B_{\infty}^{n} \cap \lceil \frac{r_1}{2n}\rceil B_1^{n}}  n \sum_{j \in [n]} \abs{L_j} \hat{w}_j + 
 \sup_{\hat{w} \in B_{\infty}^{n} \cap \lceil \frac{r_1}{2n}\rceil B_1^{n}}
\abs{\sum_{j =1}^{n}  L_j } \sum_{j} \hat{w}_j \le  3 \sum_{j
  =1}^{\lceil \frac{r_1}{2n}\rceil} L_j^{*}. \;\quad \scriptstyle\Box 
\eens

\section{Proof of Proposition~\ref{prop::orderL}}
\label{sec::prooforderL}
  First, we bound for any integer $q \in [n]$, and $\tau>0$,
  \ben
\label{eq::L4}
  \lefteqn{
\prob{\sum_{j=1}^q L_j^* > \tau}
\le \prob{\exists J \in [n], \abs{J} = q, \sum_{j \in J} \abs{L_j} > \tau}}\\
& \le &
\nonumber 
\sum_{J: \abs{J} = q} \sum_{u \in \{-1, 1\}^q} \prob{\sum_{j \in J}   u_j L_j/\Delta > \tau/\Delta}  \le  {n \choose q} 2^{q}
\exp\left(-\frac{(\tau/\Delta)^2}{C_1 C^2_0 (\max_{j} \twonorm{R_j \mu}^2) q} \right),
\een
where \eqref{eq::L4} holds by the union bound and the sub-gaussian tail
bound; To see this, notice that for each fixed index set $J \in [n]$
and vector $u \in \{-1, 1\}^q$, and $\Delta = \sqrt{p  \gamma}$,
\bens
\label{eq::qsum}
\norm{\sum_{j \in J, \abs{J} =q}   \frac{u_j L_j}{\Delta}}_{\psi_2}^2
\le
\sum_{j \in J, \abs{J} =q}  \norm{L_j/ \Delta}_{\psi_2}^2
=: \sum_{j \in J, \abs{J} =q}  \norm{\ip{\mu, \Z_j}}_{\psi_2}^2
\le C_1 q  (C_0 \max_{j} \twonorm{R_j \mu})^2,
\eens
following the proof of Lemma~\ref{lemma::anisoproj};
Hence by Lemma~\ref{lemma::anisoproj}, upon replacing $n$ with an integer $1\le
q < n$, we have for each fixed index set $J \subset [n]$,  $\mu \in \Sp^{p-1}$, 
\ben
\label{eq::qtail}
\prob{\sum_{j \in J: \abs{J} =q}   u_j L_j/ \Delta > \tau/ \Delta}
\le \exp\left(-\frac{( \tau/ \Delta)^2}{q C_1 (C_0 \max_{j} \twonorm{R_j \mu})^2} \right).
\een
Set 
$\tau_q = C_3 C_0 (\max_{j} \twonorm{R_j \mu})\Delta q \sqrt{\log (2e
  n/q)}$ for some  absolute constant $C_3 > c \sqrt{C_1}$ for $c \ge 2$.
Then by~\eqref{eq::L4} and \eqref{eq::qtail}, we have
\bens
\nonumber 
\prob{\sum_{j=1}^q L_j^* > \tau_q}
& \le &
\nonumber
{n \choose q} 2^{q} \exp\left(-\frac{C^2_3 (C_0 \max_{j} \twonorm{R_j \mu})^2
    q^2 \log (2e n/q)}{C_1 (C_0 \max_{j} \twonorm{R_j \mu})^2 q} \right) \\
& \le &
\nonumber 
e^{q \log (2 en/q)} e^{- c q \log (2e n/q)}
\le e^{- c' q \log (2e n/q)},
\eens
where ${n \choose q} 2^{q}   \le ( en/q)^q 2^q = (2 en/q)^q = 
\exp(q \log (2en/q))$.
Following the calculation as done in~\cite{GV19}, and the inequality immediately above, we have
  \bens 
  \sum_{q=1}^n  {n \choose q} 2^{q} e^{-\tau_q^2/(C_1 C_0^2 q \Delta^2)}
\le \sum_{q=1}^n e^{- c' q \log (2e n/q)}  \le \frac{c}{n^2}. \quad
\quad \scriptstyle\Box
\eens

\section{Proof of Lemma~\ref{lemma::P2SigmaY}}
\label{sec::proofofP2SigmaY}
Denote by $\hat\delta_Z :=\hat{Z} -Z^{*}$.
  First, we have
$\ip{P_2 P_1 \Psi, \hat\delta_{Z}}
 =  \tr(\hat\delta_{Z} P_2 P_1 \Psi) = (w_1 - w_2) \vecone_n^T \Psi
 \hat\delta_{Z} \frac{u_2}{n}$, where $\frac{\hat\delta_{Z} u_2}{n}  = (-\hat{w}_1, \ldots, -\hat{w}_{n_1}, 
 \hat{w}_{n_1 +1}, \ldots, \hat{w}_{n})$ for $u_2$ as in~\eqref{eq::u2}.
Now 
\bens
\abs{\ip{P_2 P_1 \offd(\Psi), \hat\delta_{Z}}}
& = &
\abs{w_1 - w_2} \abs{\vecone_n^T \offd(\Psi) \hat\delta_{Z} \frac{u_2}{n}}  \le 2 \abs{w_1 - w_2} \sum_{k} \abs{S_{k}} \hat{w}_k; \\
\abs{\ip{P_2 P_1 \diag(\Psi), \hat\delta_{Z}}}
&= &
\abs{S_{\diag} }
\le \abs{w_1 - w_2} \sum_{k \in [n]} \abs{\Psi_{kk}} \hat{w}_k \le
\abs{w_1 - w_2} \max_{k} \abs{\Psi_{kk}} \onenorm{\hat\delta_{Z}}/n;
\eens
Hence~\eqref{eq::doublesum} and~\eqref{eq::Sdiag} hold.
Now \eqref{eq::P2ZZdiag} and~\eqref{eq::P2ZZ} hold since
\bens
\lefteqn{\ip{P_2 \Psi, \hat{Z} -Z^{*}}   = 
    \tr((\hat{Z} -Z^{*}) P_2 \Psi)    =
    \tr(P_2 \Psi ( \hat{Z} -Z^{*})) }\\
  & = &
  2 \big(\sum_{k \in \MC_1} \Psi_{k k} (-\hat{w}_k) 
-\sum_{t  \in \MC_2} \Psi_{tt} \hat{w}_t \big) + 2 \big(\sum_{j \in \MC_1} Q_{j} (-\hat{w}_j) 
+\sum_{j  \in \MC_2} Q_{j} \hat{w}_j \big), \; \text{ where} \\
&& \abs{\ip{P_2 \diag(\Psi), \hat{Z} -Z^{*}}  }
=2 \abs{\sum_{k \in C_1} \Psi_{kk} (-\hat{w}_k) +\sum_{k \in C_2} (-
  \Psi_{kk}) \hat{w}_k } \le
2 \sum_{k \in [n]} \abs{\Psi_{kk}} \hat{w}_k  \; \text{ and} \\
&& \abs{\ip{P_2 \offd(\Psi), \hat{Z} -Z^{*}}}
= 2\abs{\sum_{j \in \MC_1} Q_{j} (-\hat{w} _j) 
  +\sum_{j  \in \MC_2} Q_{j} \hat{w}_j}
\le 2 \sum_{j \in [n]} \abs{Q_{j}}{\hat{w} _j}. \quad \scriptstyle\Box
\eens

\section{Proof of  Proposition~\ref{prop::orderQ} }
\label{sec::prooforderQ}
\begin{proofof2}
  Denote by $\Psi =  \Z \Z^T -\E (\Z \Z^T)$.
 The two bounds follow identical steps, and hence we will only 
 prove~\eqref{eq::Q2}. 
First, we bound for any positive integer $1\le q < n$ and any $\tau >0$,
  \ben
   \nonumber
\prob{\sum_{j=1}^q Q_j^* > \tau}
& = &
\prob{\exists J \in [n], \abs{J} = q, \sum_{j \in J} \abs{Q_j} > \tau} \\
\label{eq::q0}
& \le &
\sum_{J: \abs{J} = q} \sum_{y \in \{-1, 1\}^q}
\prob{\sum_{j \in J}   y_j Q_j > \tau}  = q_0(\tau).
\een
We now bound $q_0(\tau)$.
Denote by $w \in \{0, -1, 1\}^n$ the $0$-extended vector of $y \in \{-1, 1\}^{\abs{J}}$, that is,
for the fixed index set $J \subset [n]$, we have $w_j =y_j, \forall j \in J$ and
$w_j= 0$ otherwise. Now we can write for $A =(a_{tj}) = w \otimes
u_2$, where $u_2$ is the group  membership vector:
$\forall t \in \MC_1, j \in J, \quad a_{jt} = u_{2,t} y_j = y_j$ and
$\forall t \in \MC_2,  j \in J, \quad  a_{jt}  =  u_{2,t} y_j = -y_j$;
otherwise, $a_{jt} = 0$. In other words, we have $n q$ number of non-zero
entries in $A$ and each entry has absolute value 1.
Thus $\twonorm{A} \le  \fnorm{A} = \sqrt{q n}$ and $Q_j$ as in~\eqref{eq::defineQ},
\bens
\sum_{j \in J}   y_j Q_j =
\sum_{j \in J}   y_j \left(\sum_{t  \in \MC_1,  t\not=j}  \ip{\Z_t, \Z_j} -
  \sum_{t  \in \MC_2,  t\not=j}  \ip{\Z_t, \Z_j}\right)
=  \sum_{j \in [n]}   \sum_{t  \not=j}   a_{jt}\ip{\Z_t, \Z_j},
\eens
since each row of $\offd(A)$ in $J$ has $n -1$ nonzero entries and
$\abs{J} = q$.
Now for each fixed index set $J \in [n]$ with $\abs{J} =q$,
and a fixed vector $y \in \{-1, 1\}^q$,
\bens
\prob{\sum_{j \in J}   y_j Q_j > \tau_q}
   \nonumber
& = &   \prob{\sum_{j} \sum_{t\not=j}  a_{jt}  \ip{\Z_t, \Z_j} > \tau_q}.
\eens
By Theorem~\ref{thm::ZHW}, we have for $q \in [n]$, $A = (a_{ij}) \in
\R^{n \times n}$, $\sigma_{\max}:= \max_{i} \twonorm{H_i}$,
and $\tau_q = C_4 (C_0 \sigma_{\max})^2 q(\sqrt{np \log(2e n/q)} +
\sqrt{nq} \log(2e n/q) )$, for $C_0 \asymp K$ by definition of $K$ in~\ref{eq::Wpsi2},
\ben
\nonumber
  \prob{\abs{\sum_{i=1}^n  \sum_{j \not=i}^n \ip{\Z_{i}, \Z_{j}} a_{ij}} >  \tau_q}
  & \le &
2 \exp \left(- c\min\left(\frac{\tau_q^2}{(K \sigma_{\max})^4  p
      \fnorm{A}^2},\frac{\tau_q}{(K \sigma_{\max})^2   \twonorm{A}} \right)\right)
\\
& \le &
\label{eq::q1}
2 \exp \left(- c (C_4^2 \wedge C_4)  q \log(2e n/q)\right), 
\een 
where $\max_{i} \norm{Z_i}_{\psi_2} \le C K \sigma_{\max}$  in the
sense of \eqref{eq::Zpsi2}.
Now by~\eqref{eq::q0} and~\eqref{eq::q1},
\bens
q_0(\tau_q)
& \le &
 {n \choose q} 2^{q}   \exp\left(- c (C_4^2 \wedge C_4) q \log(2e 
  n/q)\right)   \le
2 \exp \left(- c' (C_4^2 \wedge C_4) q \log(2e n/q)\right),
\eens
where ${n \choose q} 2^{q}   \le ( en/q)^q 2^q = (2 en/q)^q = 
\exp(q \log (2en/q))$. Hence following the calculation as done in~\cite{GV19},  we have
\bens 
\sum_{q=1}^n q_0(\tau_q) \le
\sum_{q=1}^n  {n \choose q} 2^{q} \exp\left(- c (C_4^2 \wedge C_4) q
  \log(2e   n/q)\right)
\le \sum_{q=1}^n e^{- c' q \log (2e n/q)}  \le \frac{c'}{n^2}.
\eens
The proposition thus holds.
\end{proofof2}

\section{Proof of Corollary~\ref{coro::orderQ}}
\label{sec::prooforderQcoro}
\begin{proofof2}
For any positive integer $q \in [n]$ and $\hat{w}$ as in~\eqref{eq::weightnorm},
\ben
  \nonumber
  \sup_{\hat{w} \in B_{\infty}^{n} \cap q B_1^{n}}
  \sum_{j \in [n]} \abs{Q_j} \hat{w}_j &\le &
  \sup_{\hat{w} \in B_{\infty}^{n} \cap q B_1^{n}}
  \sum_{j \in [n]} Q^*_j \hat{w}^*_j 
 =   \sup_{\hat{w} \in B_{\infty}^{n} \cap q B_1^{n}}
\big(\sum_{j =1}^{q} Q_j^{*}  \hat{w}^{*}_j +
\sum_{j =q+1}^{n} Q_j^{*}  \hat{w}^{*}_j \big) \\
 \label{eq::Qsum}
&\le &
\sum_{j =1}^{q} Q_j^{*}  + q Q_{q+1}^{*}  \le 2 \sum_{j =1}^{q}
Q_j^{*}.
\een
 Following the same arguments in Lemma~\ref{lemma::P2final}, we have
  by~\eqref{eq::P2ZZ} and \eqref{eq::weightnorm}, 
\ben
\nonumber
\sup_{\hat{Z} \in  M_{\opt}
   \cap Z^* + r_1 B_1^{n \times n}} \abs{\ip{P_2 \offd(\Psi), \hat{Z} - Z^{*}}}
  &  \le &   2  \sup_{\hat{w} \in B_{\infty}^{n} \cap \lceil r_1 /(2n) \rceil B_1^{n}}
  \sum_{j \in [n]} \abs{Q_{j}} \hat{w}_j
  \een
Now  \eqref{eq::duet} follows from~\eqref{eq::P2ZZ}, 
~\eqref{eq::Qsum}, and~\eqref{eq::Q2}, where we set  $q =\lceil r_1 
/(2n) \rceil$.
Notice that for any positive integer $q  \in [n]$,  following the same
argument, we have
 \ben
 \label{eq::Ssum}
 \sup_{\hat{w} \in B_{\infty}^{n} \cap q B_1^{n}}
\sum_{j \in [n]} \abs{S_j} \hat{w}_j 
& \le & \sup_{\hat{w} \in B_{\infty}^{n} \cap q B_1^{n}}
\sum_{j \in [n]} S^*_j \hat{w}^*_j  \le 2 \sum_{j =1}^{q} S_j^{*}.
\een
Hence \eqref{eq::trio}  follows from~\eqref{eq::doublesum}, 
   \eqref{eq::Ssum},~\eqref{eq::Q2},  and   Proposition~\ref{prop::orderQ}. 
\end{proofof2}

\section{Additional proofs for Section~\ref{sec::reduction}}

\label{sec::reductionproofs}

\subsection{Proof of Lemma~\ref{lemma::YYdec}}
\label{sec::SigmaYYproj}
\begin{proofof2}
Reduction in the operator norm holds since by Proposition~\ref{prop::decompose},
\bens
\nonumber
\hat\Sigma_Y -\Sigma_Y
& :=& (Y - \E Y)(Y - \E Y)^T - \E ((Y - \E Y)(Y - \E Y)^T) \\
\nonumber
& = & (I-P_1) (\Z \Z^T - \E(\Z \Z^T)) (I-P_1)
\eens
and clearly,
 \ben
   \twonorm{\hat\Sigma_Y -\Sigma_Y}
   \label{eq::ZZop}
& \le & \twonorm{I-P_1} \twonorm{\Z \Z^T - \E(\Z \Z^T)} \twonorm{I-P_1} \\
   \nonumber
& \le & \twonorm{\Z \Z^T - \E(\Z \Z^T)}. 
\een
\end{proofof2}

\subsection{Proofs for Lemma~\ref{lemma::tiltproject} }
\label{sec::tiltproj}
First, we verify \eqref{eq::EYpre}:
\ben
\nonumber
\forall i \in \MC_1\; \; \E Y_i
&= & \E X_i - (w_1 \mu^{(1)} + w_2 \mu^{(2)}) \\
\label{eq::EY1}
& = &
\mu^{(1)}(1 - w_1) - w_2 \mu^{(2)}
= 
w_2 (\mu^{(1)} -  \mu^{(2)}) \\
\nonumber
\forall i \in \MC_2 \; \;  \E Y_i
& = & \E X_i - (w_1 \mu^{(1)} + w_2 \mu^{(2)}) \\
\label{eq::EY2}
&= &
\mu^{(2)} (1- w_2) - w_1 \mu^{(1)}  =  w_1 (\mu^{(2)} -\mu^{(1)}).
\een

\begin{lemma}
  \label{lemma::pairwise4}
Suppose all conditions in Lemma~\ref{lemma::tiltproject} hold. Then
\ben
\nonumber
\sup_{q \in \Sp^{n-1}}\le \sum_{i=1}^n q_i \ip{Y_i - \E (Y_i),\mu^{(1)} -\mu^{(2)}}
\label{eq::pairwise4}
& \le & 2 \sup_{q \in \Sp^{n-1}}  \sum_{i=1}^n q_i \ip{\Z_i, \mu^{(1)} -\mu^{(2)}}.
\een
\end{lemma}

\subsection{Lemma~\ref{lemma::tiltproject} restated}
Besides the operator norm bound as stated in Lemma~\ref{lemma::tiltproject},
We will also prove the following bound directly:
\ben
\label{eq::cutnormred}
\infonenorm{M_Y}
& \le & 8 w_1 w_2 (n-1) \sum_{i=1}^n \abs{\ip{\Z_i, \mu^{(1)    }-\mu^{(2) } }}
      \een
Due to the symmetry, we need to compute only
\bens
 \norm{(Y-\E(Y))\E(Y)^T}
& = & \norm{(\Z -(\vecone{(X)} - \E \vecone{(X)} )) (\E(X) - \E \vecone{(X)}
  )^T} 
\eens
First, we show that \eqref{eq::pairwise2} holds. 
Now
\bens
Y_i - \E Y_i  & =&
(X_i - \E X_i) - ((\hat{\mu}_n - \E \hat{\mu}_n) =
\Z_i- \left(\inv{n} \sum_{i=1}^n (X_i - \E X_i) \right)\\
& = & \frac{n-1}{n} \Z_i - \inv{n}\sum_{j\not=i}^n \Z_j  =
\inv{n} \sum_{j\not=i}^n (\Z_i - \Z_j)
\eens
and for $x_i \in \{-1,1\}$,
\bens
\lefteqn{
\sum_{i=1}^n x_i \ip{Y_i -\E Y_i, \mu^{(1)} -\mu^{(2)}} =
 \inv{n}  \sum_{i=1}^n x_i \sum_{j\not=i}^n \ip{(\Z_i - \Z_j), \mu^{(1)}
  -\mu^{(2)}}}  \\
\nonumber 
& \le &
\inv{n}  \sum_{i=1}^n \sum_{j\not=i}^n \abs{\ip{(\Z_i - \Z_j), \mu^{(1)}
    -\mu^{(2)}} } \le  \frac{2(n-1)}{n} \sum_{i=1}^n \abs{\ip{\Z_i, \mu^{(1)}  -\mu^{(2)}} }.
\eens
Then we have by definition of the cut norm,
\eqref{eq::EY1}, \eqref{eq::EY2}, and \eqref{eq::pairwise2},
\bens
\lefteqn{
\infonenorm{(Y-\E (Y)) \E (Y)^T} = \sup_{x, y \in \{-1,1\}^{n}}
\sum_{i=1}^n x_i \cdot}\\
& &
\big(\sum_{j \in \MC_1} y_j \ip{Y_i - \E (Y_i),  w_2 (\mu^{(1)} -\mu^{(2)})} +
  \sum_{j \in \MC_2} y_j  \ip{Y_i - \E (Y_i), w_1 (\mu^{(2)} -\mu^{(1)})}\big)\\
& \le &
\sup_{x, y \in \{-1,1\}^{n}}
  \sum_{i=1}^n x_i \big(\ip{ Y_i - \E (Y_i), \mu^{(1)}
    -\mu^{(2)}} (\sum_{j \in \MC_1} w_2 y_j - \sum_{j \in \MC_2} 
w_1 y_j  ) \big)\\
& \le &
(w_2 \abs{\MC_1} + w_1 \abs{\MC_2}) \sum_{i=1}^n \abs{\ip{Y_i - \E 
    (Y_i), \mu^{(1)}  -\mu^{(2)}}} \\
& = &
2 w_2 w_1 n \max_{x \in \{-1, 1\}^n}\sum_{i=1}^n x_i \ip{Y_i - \E 
  (Y_i), \mu^{(1)}  -\mu^{(2)}}  \\
& \le &
4 w_1 w_2(n-1) \sum_{i=1}^n \abs{\ip{\Z_i, \mu^{(1)}  -\mu^{(2)}} }.
\eens
Thus  \eqref{eq::cutnormred} holds. It remains to show the operator norm bound.

\subsection{Proof of Lemma~\ref{lemma::tiltproject} }
\begin{proofof}{Lemma~\ref{lemma::tiltproject} }
Similarly, we have by \eqref{eq::EY1} and \eqref{eq::EY2},
\ben
\nonumber
\lefteqn{
  \twonorm{(Y-\E (Y)) \E (Y)^T} = \sup_{q, h \in \Sp^{n-1}}
\sum_{i=1}^n q_i \cdot}\\
& &
\nonumber
\quad \quad \big(\sum_{j \in \MC_1} h_j \ip{Y_i - \E (Y_i),  w_2 (\mu^{(1)} -\mu^{(2)})}
+  \sum_{j \in \MC_2} h_j  \ip{Y_i - \E (Y_i), w_1 (\mu^{(2)}
  -\mu^{(1)})} \big)\\
&  &
\label{eq::defQ}
\quad \quad \quad \le \sup_{q, h \in \Sp^{n-1}} 
\big(\sum_{i=1}^n q_i \ip{ Y_i - \E (Y_i),\mu^{(1)} -\mu^{(2)}}
  (\sum_{j \in \MC_1} w_2 h_j - \sum_{j \in \MC_2} w_1 h_j  ) \big)=: Q,
\een
where by \eqref{eq::defQ} and \eqref{eq::pairwise4}, and $w_1 w_2 \le 1/4$,
\bens
Q & \le &
\sup_{q \in \Sp^{n-1}} 
\abs{\sum_{i=1}^n q_i \ip{ Y_i - \E (Y_i),\mu^{(1)}
      -\mu^{(2)}} }\cdot  \sup_{h \in \Sp^{n-1}} 
\abs{\sum_{j \in \MC_1} w_2 h_j - \sum_{j \in \MC_2} w_1 h_j  } \\
& \le &
2 \sup_{q\in \Sp^{n-1}} \abs{\sum_{i=1}^n q_i \ip{\Z_i, \mu^{(1)}
    -\mu^{(2)}} } \cdot \sqrt{n w_1 w_2}  \le  \sqrt{n} \sup_{q \in \Sp^{n-1}} \abs{\ip{\sum_{i} q_i \Z_i,
    \mu^{(1)} -\mu^{(2)}}}.
\eens
where for $\twonorm{h_{\MC_i}} =  \sqrt{\sum_{j \in \MC_i} h^2_j}, i
  =1, 2$ and $h \in \Sp^{n-1}$, 
  \bens
\abs{\sum_{j \in \MC_1} w_2 h_j - \sum_{j \in 
    \MC_2} w_1 h_j }
&  \le & 
w_2 \sum_{j \in \MC_1} \abs{h_j} + w_1 \sum_{j \in 
  \MC_2} \abs{h_j}   \\
& =: &  w_2  \onenorm{h_{\MC_1}} + w_1 \onenorm{h_{\MC_2}} \\
&  \le & 
w_2 \sqrt{\abs{\MC_1}} \twonorm{h_{\MC_1}} + w_1 \sqrt{\abs{\MC_2}}
\twonorm{h_{\MC_2}} \\
&  \le & 
\sqrt{w_1 w_2 n}
\big(\sqrt{w_2} \twonorm{h_{\MC_1}} + \sqrt{w_1}  \twonorm{h_{\MC_2}} \big)
\le \sqrt{w_1 w_2 n},
\eens
where $1= w_1 + w_2 \ge 2 \sqrt{w_1 w_2}$ and by
Cauchy-Schwarz, we have for $\bar{w_0} = (\sqrt{w_2}, \sqrt{w_1})$ such
that $\twonorm{\bar{w_0}} =\sqrt{w_1 + w_2} =1$ and $z =( \twonorm{h_{\MC_1}},
\twonorm{h_{\MC_2}})$ such that $\twonorm{z} = 1$,
$$\ip{\bar{w_0}, z} = \big(\sqrt{w_2} \twonorm{h_{\MC_1}} + \sqrt{w_1}
  \twonorm{h_{\MC_2}} \big) \le \twonorm{w_0} \twonorm{z} =1.$$
\end{proofof}

\begin{proofof}{Lemma~\ref{lemma::pairwise4}}
\bens
\lefteqn{\sum_{i=1}^n q_i \ip{Y_i - \E (Y_i),\mu^{(1)} -\mu^{(2)}} = 
\inv{n} \sum_{i=1}^n q_i \ip{\sum_{j\not=i} (\Z_i - \Z_j),\mu^{(1)}
  -\mu^{(2)}} }\\
&  &
= \frac{n-1}{n} \left(\sum_{i=1}^n q_i \ip{\Z_i, \mu^{(1)} -\mu^{(2)}}
\right) + 
\inv{n} \sum_{i=1}^n q_i \ip{\Z_i -\sum_{j=1}^n \Z_j,\mu^{(1)}
  -\mu^{(2)}} \\
& = &
\sum_{i=1}^n q_i \ip{\Z_i, \mu^{(1)} -\mu^{(2)}}  -
\inv{n} \sum_{i=1}^n q_i \sum_{j=1}^n \ip{\Z_j,\mu^{(1)}  -\mu^{(2)} },
\eens
where 
\bens
\abs{\inv{n} \sum_{i=1}^n q_i \sum_{j=1}^n \ip{\Z_j,\mu^{(1)}
    -\mu^{(2)} }} 
&\le & \sup_{q \in \Sp^{n-1}}  \inv{n} \onenorm{q} \abs{\sum_{j=1}^n
  \ip{\Z_j,\mu^{(1)} -\mu^{(2)}}} \\
& \le & \abs{ \sum_{j=1}^n \inv{\sqrt{n}} \ip{\Z_j,\mu^{(1)}
    -\mu^{(2)}}} \\
& \le & 
\label{eq::pairwise5}
\sup_{q \in \Sp^{n-1}}  \sum_{i=1}^n q_i \ip{\Z_i, \mu^{(1)} -\mu^{(2)}}.
\eens
Thus \eqref{eq::pairwise4} holds and the lemma is proved.
\end{proofof}

\bibliography{final,subgaussian,clustering}

\end{document}